\documentclass[12pt]{article}

\usepackage[english]{babel}

\usepackage{amsmath,amsthm,amssymb, graphicx, multicol, array, bm, bbm}
\usepackage{graphicx}
\usepackage[colorlinks=true, allcolors=blue]{hyperref}
\usepackage{caption}
\usepackage{subcaption}
\usepackage{algorithm}
\usepackage[utf8]{inputenc}
\usepackage[margin=1in]{geometry}
\usepackage{booktabs}
\usepackage{array}
\usepackage{amsmath, amsfonts, amssymb}
\usepackage{color}
\usepackage{hyperref}
\usepackage{bbm}
\usepackage[noend]{algpseudocode}
\usepackage{subcaption}
\usepackage{graphbox}
\usepackage{accents}
\usepackage{rotating}
\usepackage{pifont}
\usepackage{pdflscape}
\usepackage{lscape}
\usepackage{subcaption}
\usepackage{float}
\usepackage{multirow}
\usepackage[table]{xcolor}
\usepackage{authblk}
\usepackage{setspace}
\usepackage{xr}
\usepackage{pdfpages}

\usepackage[style=numeric-comp,sorting=none,backend=biber,maxnames=3,minnames=3,giveninits=true,doi=false,isbn=false,url=false]{biblatex}
\DeclareFieldFormat*{title}{#1} 
\DeclareFieldFormat*{journal}{\mkbibemph{#1}} 
\renewbibmacro{in:}{} 
\definecolor{darkgreen}{rgb}{0.0, 0.5, 0.0}
\newcommand{\greenup}{\textcolor{darkgreen}{\boldmath$\uparrow$}}
\newcommand{\reddown}{\textcolor{red}{\boldmath$\downarrow$}}
\newcommand{\bluebf}[1]{\textbf{\textcolor{blue}{#1}}}
\newcommand{\redbf}[1]{\textbf{\textcolor{red}{#1}}}

\newcommand{\cmark}{\ding{51}}%
\newcommand{\xmark}{\ding{55}}%

\title{Generating Synthetic Electronic Health Record Data: a Methodological Scoping Review with Benchmarking on Phenotype Data and Open-Source Software}

\begin{document}

\author[1]{Xingran Chen}
\author[1]{Zhenke Wu \thanks{Corresponding Author.}}
\author[1]{Xu Shi}
\author[2]{Hyunghoon Cho}
\author[3]{Bhramar Mukherjee \thanks{Senior Author.}}

\affil[1]{Department of Biostatistics, University of Michigan}
\affil[2]{Department of Biomedical Informatics and Data Science, Yale University}
\affil[3]{Department of Biostatistics, Yale University}
\affil[ ]{\texttt{\{chenxran,zhenkewu,shixu\}@umich.edu}}
\affil[ ]{\texttt{\{hoon.cho,bhramar.mukherjee\}@yale.edu}}

\maketitle

\newpage
\doublespacing

\begin{abstract}

\noindent \textbf{Objectives:} To conduct a scoping review of existing approaches for synthetic Electronic Health Records (EHR) data generation, to benchmark major methods, and to provide an open-source software and offer recommendations for practitioners.

\noindent \textbf{Materials and Methods:} We search three academic databases for our scoping review. Methods are benchmarked on open-source EHR datasets, Medical Information Mart for Intensive Care III and IV (MIMIC-III/IV). Seven existing methods covering major categories and two baseline methods are implemented and compared. Evaluation metrics concern data fidelity, downstream utility, privacy protection, and computational cost.

\noindent \textbf{Results:} \textcolor{black}{48} studies are identified and classified into five categories. Seven open-source methods covering all categories are selected,  trained on MIMIC-III, and evaluated on MIMIC-III or MIMIC-IV for transportability considerations. Among them, Generative Adversarial Network (GAN)-based methods demonstrate competitive performance in fidelity and utility on MIMIC-III; rule-based methods excel in privacy protection. Similar findings are observed on MIMIC-IV, except that GAN-based methods further outperform the baseline methods in preserving fidelity.  

\noindent \textbf{Discussion:} Method choice is governed by the relative importance of the evaluation metrics in downstream use cases. We provide a decision tree to guide the choice among the benchmarked methods. An extensible Python package, “SynthEHRella”, is provided to facilitate streamlined evaluations.

\noindent \textbf{Conclusion:} GAN-based methods excel when distributional shifts exist between the training and testing populations. Otherwise, CorGAN and MedGAN are most suitable for association modeling and predictive modeling, respectively. Future research should prioritize enhancing fidelity of the synthetic data while controlling privacy exposure, and comprehensive benchmarking of longitudinal or conditional generation methods.
 
\end{abstract}

\newpage

\section{INTRODUCTION}

Electronic Health Records (EHRs) store patients' health information collected via encounters with health systems in a digital format. 
Such rich repositories that integrate disease phenotypes, multi-omics markers, medical images, medications, and laboratory results have become a critical resource for research in biomedicine\cite{fodeh2016mining, yadav2018mining} and public health\cite{birkhead2015uses, friedman2013electronic, kruse2018use}. For researchers in quantitative biomedical areas, EHR-linked biobanks\cite{sudlow2015uk, all2019all, zhou2022global} are valuable for empirical evaluations of analytical methods using multi-modal data\cite{tarczy2013survey, linder2021role, yadav2018mining, beesley2020emerging}. However, few real-world EHR data that are publicly available exist due to patient consent requirements and patient privacy considerations.

Synthetic EHR data generation \textcolor{black}{offers} a promising solution to unlock enormous research and educational potential of real-world healthcare data while safeguarding confidentiality. First, high-quality synthetic EHR data enable the development and evaluation of analytical tools with less privacy concerns\cite{rankin2020reliability, murcia2024automating, shi2022generating, perets2023subpopulation, muller2022synthesising}. \textcolor{black}{Second, it is ethically more defensible\cite{zhang2020ensuring, biswal2021eva,bing2022conditional} for minority subpopulation with higher re-identification risk.} Third, synthetic EHR generation facilitates experimenting with the size of the real training and testing data over multiple replicates for accuracy and uncertainty assessment of methods\cite{tall2020generating, chen2022simulation}. Finally, synthetic data provide ideal \textcolor{black}{risk-free} environments for training healthcare professionals, data scientists and other researchers in EHR usage, clinical workflows, and data analysis\cite{yale2020generation, laderas2017teaching}.

\textcolor{black}{Despite} the \textcolor{black}{growing} literature on synthetic EHR generation (Figure~\ref{fig:num-of-pub}) \textcolor{black}{along with numerous review papers}\cite{mendelevitch2021fidelity, hernandez2022synthetic, ghosheh2022review, budu2024evaluation,achterberg2024evaluation}, few works have conducted comprehensive benchmarking of existing methodologies. The latest benchmarking paper\cite{yan2022multifaceted}, while making valuable contributions to benchmarking synthetic EHR generation methods, has two major limitations. First, it evaluated exclusively the methods built around Generative Adversarial Network (GAN), \textcolor{black}{overlooking other categories of approaches} we identified in Section \ref{sec:summary-of-literature-review}. Second, \textcolor{black}{it relied on} closed-source data, limiting the reproducibility \textcolor{black}{and future evaluation} by other researchers. \textcolor{black}{W}ith an explosion in latest generative artificial intelligence (GenAI) methods, including large language models (LLMs) and diffusion models, a more comprehensive and updated review, evaluation, and benchmarking is needed.  

\textcolor{black}{This paper makes four key contributions to address this knowledge gap. First, we conduct a methodological scoping review to summarize existing works for methodological development in synthetic EHR generation. Second, we propose a benchmarking on fidelity, utility, and privacy evaluation of diverse categories of methods that were not included in previous reviews and benchmarking studies\cite{yan2022multifaceted, budu2024evaluation, goncalves2020generation}. Third, we use open-source EHR phenotype datasets, MIMIC-III and the recently released MIMIC-IV datasets, as our benchmarks and release SynthEHRella, an open-source, extensible, on-premises benchmarking toolkit not available in prior studies. Finally, we evaluate the transportability capability of existing methods by training them on MIMIC-III and evaluate on MIMIC-IV.} Two comparison tables regarding the methods and evaluation metrics included in our study and the latest benchmarking work\cite{yan2022multifaceted}, Table~\ref{tab:metrics-comparison} and Table~\ref{tab:methods-comparison}, are presented in the Supplementary Materials. The comprehensive scoping review in this paper and the software will be relevant for EHR researchers and the medical informatics community moving forward.


\section{METHOD}
\label{sec:method}
\subsection{\textcolor{black}{Scoping} review}
\label{sec:literature-review}

\textcolor{black}{In the spirit of a scoping review, we conducted a search on May 5th, 2024 to identify studies that propose synthetic EHR generation methods. The full search strings are provided in Section~\ref{secsupp:method} of the Supplementary Materials. Searches were performed on Google Scholar, PubMed, and Semantic Scholar. For PubMed, we used the built-in search engine, while for the others we employed the Publish or Perish software\cite{harzing2007pop} to retrieve papers. Non-English papers, patents, and non-research outputs were excluded via manual and automated filtering. Two additional verification searches were conducted on August 30th, 2024 and March 30th, 2025, separately to identify any updated literature before August 30th, 2024.}

\textcolor{black}{All search results underwent a rigorous screening by the authorship team to exclude papers unrelated to synthetic EHR pipeline development. For the inclusion-eligible literature, the following information was extracted: algorithm name, model type, year of publication, support for longitudinal and conditional generation, evaluation metrics used, codebase accessibility, whether evaluation was conducted on a MIMIC dataset, and, if so, which data modalities were included. We refer the readers to the PRISMA-ScR checklist\cite{tricco2018prisma} in the Supplementary Materials for details.}

\textcolor{black}{The results of the scoping review are concluded in Section~\ref{sec:summary-of-literature-review} and Table~\ref{tab:literature}. In terms of the benchmarking evaluation}, we selected seven distinct methods based on the specific guiding principles. We refer the readers to Section~\ref{sec:evaluation} and~\ref{sec:method-selected-for-benchmarking} for details.

\subsection{Benchmarking datasets}
\label{sec:datasets}

We introduce our benchmarking datasets, with detailed description of the datasets, coding system, and descriptive analysis in Section~\ref{secsupp:method}.

\paragraph{MIMIC-III\cite{johnson2016mimic}} MIMIC-III is a large-scale EHR dataset comprising 58,976 unique ICU admissions at Beth Israel Deaconess Medical Center (BIDMC) in Boston, Massachusetts between 2001 and 2012. The dataset consists of mainly adult patients, \textcolor{black}{and} 7,874 neonates admitted to the Neonatal Intensive Care Unit (NICU). All patient data have been de-identified \textcolor{black}{for} patient privacy. The dataset uses International Classification of Disease codes (ICD-9) to record diagnostic events.

\paragraph{MIMIC-IV\cite{johnson2023mimic}} MIMIC-IV is a recently released, updated version of MIMIC series datasets that features a larger patient cohort, using both version 9 and 10 of ICD as coding system. MIMIC-IV includes 431,231 hospital admissions of patients that were admitted into either the emergency departments (ED) or ICUs between 2008 and 2019 at BIDMC. Due to the overlap of data collection periods (2008-2012), a subset of patients in MIMIC-IV also appear in MIMIC-III, although it is difficult to link profiles across datasets because of the de-identification procedures.

\subsection{\textcolor{black}{Benchmarking} evaluation}
\label{sec:evaluation}

\textit{Statement of Evaluation Task}

ICU-based (and ED-based) EHR datasets, such as MIMIC-III and MIMIC-IV, have a longitudinal nature, where each individual patient may have multiple encounters. To this end, we convert all longitudinal records into a cross-sectional format by aggregating each records into a representation. Formally, consider an EHR dataset with cross-sectional format $\mathcal{D} = \{\mathbf{x}_{1}, \ldots, \mathbf{x}_{N}\}$, where each entry $\mathbf{x}_{i} = (x_{i}^{(1)}, \ldots, x_{i}^{(K)})^\top$ is a $K$-dimensional binary vector representing the presence/absence of diseases coded by a specific coding system (e.g., ICD-9). Each dimension $x_{i}^{(k)} = 1$ if the $k$-th disease is diagnosed at least once in all visits of $i$-th patient, $k=1, \cdots, K$, $i=1, \cdots, N$. We represent the datasets by two $N \times K$ and $M \times K$ binary matrices denoted as $\mathbf{X}_{real}$ and $\mathbf{X}_{syn}$, respectively.

\vspace{0.5cm}

\noindent
\textit{Evaluation Metrics}

We introduce key evaluation metrics. Full descriptions, including the evaluation of computational cost and numerical studies, are presented in Section~\ref{secsupp:method}.

\paragraph{Fidelity}

We first evaluate the \textbf{dimension-wise distributional discrepancy} between the real and the synthetic datasets. We compute the Maximum Mean Discrepancy (MMD) to evaluate the absolute difference in prevalence. We use Root Mean Squared Percentage Error (RMSPE) and Mean Absolute Percentage Error (MAPE) for assessing the relative difference in prevalence. They are computed as follows:

$$ \hat{\mu}^{(k)}_{syn} = \frac{1}{M} \sum^M_{i=1} x^{(k)}_{i, syn},~\quad \hat{\mu}^{(k)}_{real} = \frac{1}{N} \sum^N_{i=1} x^{(k)}_{i, real}, \quad k = 1, \dots, K;$$

$$ \texttt{MMD} = \max_{1 \leq k \leq K} \left | \hat{\mu}^{(k)}_{syn} - \hat{\mu}^{(k)}_{real} \right |;$$

$$ \texttt{RMSPE} = 100 \times \sqrt{\frac{1}{K} \sum_{i=1}^{K} \left( \frac{\hat{\mu}^{(i)}_{syn} - \hat{\mu}^{(i)}_{real}}{\hat{\mu}^{(i)}_{real}} \right)^2},~ \texttt{MAPE} = \frac{100}{K} \sum_{i=1}^{K} \left| \frac{\hat{\mu}^{(i)}_{syn} - \hat{\mu}^{(i)}_{real}}{\hat{\mu}^{(i)}_{real}} \right|. $$

All the RMSPE values are divided by a factor of 100 for ease of presentation.

To evaluate whether generative models preserve pair-wise correlations between the phecodes, we calculate the Pearson correlation matrix following\cite{yan2022multifaceted, yuan2023ehrdiff} for both real and synthetic datasets and compute the \textbf{correlation Frobenius distance} (CFD):

$$\hat{\Sigma}_{real} = {\rm Corr}(\mathbf{X}_{real}), \quad \hat{\Sigma}_{syn} = {\rm Corr}(\mathbf{X}_{syn}),$$
$$\texttt{CFD} = \left \lVert \hat{\Sigma}_{real} - \hat{\Sigma}_{syn} \right \rVert_F = \sqrt{\sum^K_{k=1} \sum^K_{k'=1} (a^{real}_{kk'} - a^{syn}_{kk'})^2},$$
where $a^{real}_{kk'}$ and $a^{syn}_{kk'}$ represent the ($k,k'$)-th element in correlation matrices $\hat{\Sigma}_{real}$ and $\hat{\Sigma}_{syn}$, respectively.

Finally, we assess \textcolor{black}{whether synthetic data can be distinguished from the real data used for training, a task} referred to as \textbf{discriminative prediction}. \textcolor{black}{We train a logistic regression model on both real and synthetic data to predict whether a given sample is real or synthetic.} \textcolor{black}{We do not fine-tune pre-trained checkpoints for discriminators, which was applied in existing work\cite{zhang2022keeping},  due to fair comparison consideration. See Section~\ref{secsupp:method} for detailed explanations.} We validate the model via 5-fold cross-validation. We report the Area Under the Curve (AUC) and Accuracy (ACC), where lower values indicate \textcolor{black}{greater similarity between synthetic and real data.}

\paragraph{Utility}

We evaluate the downstream utility of the synthetic data from two aspects. Firstly, \textbf{analytical utility} assesses \textcolor{black}{how well synthetic data approximates inferential results derived from real data}. \textcolor{black}{Specifically, we estimate associations between binary outcome and explanatory phenotypes}, denoted by $x_i^{(k)}$ and $x_i^{(k')}$, by fitting a logistic regression model. The point estimates and the 95\% confidence intervals of regression coefficients (i.e., log-odds ratios) of the predictor phenotype are reported. \textcolor{black}{Greater} overlap in 95\% Confidence Intervals (CIs) \textcolor{black}{ from real and synthetic data indicates higher analytical utility.}

Secondly, \textbf{predictive utility} measures \textcolor{black}{how effectively synthetic data trains predictive machine learning (ML) models for downstream tasks.} To evaluate existing methods' predictive utility, \textcolor{black}{we construct a binary classification task, treating the} $k$-th phenotype $x^{(k)}_i$ as the outcome, and the remaining phenotypes in $\mathbf{x}_i$ as predictors. In this work, we list three scenarios commonly used in the literature\cite{yan2022multifaceted, li2023generating} that concern different combinations of the sources of data used for training and testing a ML method: a) Train on Synthetic, Test on Real (TSTR); b) Train on Synthetic + Real, Test on Real (TSRTR); c)  Train on Real and Test on Real (TRTR). \textcolor{black}{All the testings are based on a same subset of real data for fair comparison}. All the three scenarios are visualized in Figure~\ref{fig:pred-utility}, with detailed description in Section~\ref{secsupp:method}.

For each of the above three scenarios, we calculate the AUC and ACC of the classifiers. \textcolor{black}{Since TRTR is real-data-based,} we use it as a baseline and report the differences: ${\rm AUC}_{TSTR}-{\rm AUC}_{TRTR}$ and ${\rm AUC}_{TSRTR}-{\rm AUC}_{TRTR}$, and similarly for ACC. Higher values indicate better utility of the synthetic data.

\paragraph{Privacy}

First, we \textcolor{black}{evaluate} \textbf{membership inference risk} (MIR)\cite{shokri2017membership} \textcolor{black}{which evaluates the likelihood that} an attacker \textcolor{black}{with} access to a patient's complete medical record, could determine whether this patient \textcolor{black}{was} included in the training dataset. We compute the minimum Euclidean distance between each real medical record and the synthetic EHR dataset: 

$$d_i = \texttt{min}_{\mathbf{x}\in \mathcal{D}_{syn}}\texttt{dist}(\mathbf{x}_{i, real}, \mathbf{x}),$$

where $\mathbf{x}_{i, real}$ is a record from the real dataset. \textcolor{black}{To quantify dataset-level membership inference risk, we report the mean and median of these minimum distances:}

$$ \texttt{MIR}_{mean} = \frac{1}{N} \sum^N_{i=1} d_i,~ \texttt{MIR}_{median} = \texttt{median}(d_1, \dots, d_N),$$

where higher values indicate lower membership inference \textcolor{black}{risk}. \textcolor{black}{In addition, we present Figure~\ref{fig:mir-hist},~\ref{fig:mir-cdf} and Table~\ref{tab:exact_match} for a more nuanced evaluation of the distributional differences of $d_i$ across methods along with \% of real data being perfectly matched with the synthetic samples. Detailed explanation of this evaluation can be found in Sections~\ref{secsupp:method} and~\ref{secsupp:results}.}

Additionally, we assess \textbf{attribute inference risk} (AIR)\cite{ganju2018property} which occurs when attackers who have access to partial information about a patient's real medical records could infer the missing attributes based on the synthetic data. We apply \textcolor{black}{a} 1-Nearest-Neighbor approach to match each real record with \textcolor{black}{its closest synthetic counterpart} based on the assumed known attributes. \textcolor{black}{The missing attributes in the real record are then predicted using the matched synthetic record, and the F1 score is computed.} A lower value indicates a lower attribute inference \textcolor{black}{risk}.

\section{RESULTS}
\label{sec:results}

\subsection{Summary of \textcolor{black}{scoping} review}
\label{sec:summary-of-literature-review}

\textcolor{black}{Our literature search retrieved 304 papers from PubMed, 488 from Google Scholar, and 68 from Semantic Scholar. After deduplication, 761 unique publications remained. An abstract and full-text screening is conducted to determine whether these papers proposed new methodological developments for synthetic EHR generation, resulting in the identification of 42 relevant papers. In addition, we identify 1 paper as a related work mentioned in these 42 papers, and 4 papers as existing methods mentioned in review papers retrieved. We also include Plasmode\cite{franklin2014plasmode} in our screening results due to its widespread use among practitioners, although it is not captured in the initial search. This gives a total of 48 papers as a result of the scoping review. Based on their core ideas or backbone architectures, these methods are grouped into five primary categories:}

\begin{landscape}
\begin{table}[]
\caption{A summary of existing methods for synthesizing EHR data, outlining the functionalities included, aspects of evaluation included, data source used, the availability of code, and the event modality covered in each method. Abbreviation: DWS: dimension-wise similarity; CMS: correlation matrix similarity; DP: discriminative prediction; Human: Human Evaluation; TSTR: Train on Synthetic, Test on Real ; TSRTR: Train on Synthetic + Real, Test on Real; MIR: Membership Inference Risk; AIR: Attribute Inference Risk. \dag: this method only evaluated on MIMIC-IV. \cmark\; denotes that the respective method includes the listed functionality, evaluation metric, data source, or event modality; \xmark; indicates the lack of inclusion of that aspect. The ``-'' symbol is used in the event modality columns when the MIMIC-III dataset is not used as a data source by the method.}
\label{tab:literature}
\resizebox{1.0\columnwidth}{!}{
\begin{tabular}{cccc|cc|cccc|cc|cc|cc|ccc}
\toprule
\multicolumn{1}{c}{References} & \multicolumn{1}{c}{Model}                                                                        & \multicolumn{1}{c}{Type}                                              & \multicolumn{1}{c}{Year} & \multicolumn{1}{|c}{Conditional} & \multicolumn{1}{c|}{Longitudinal} & \multicolumn{4}{c|}{Fidelity}                               & \multicolumn{2}{c|}{Utility} & \multicolumn{2}{c|}{Privacy} & \multicolumn{1}{c}{Open-Source} & \multicolumn{1}{c}{Use MIMIC-III} & \multicolumn{3}{|c}{Event Modality} \\
 & & & & & & DWS & CMS & DP & Human & TSTR & TSRTR & MIR & AIR & & & \multicolumn{1}{c}{Diagnosis} & \multicolumn{1}{c}{Procedure} & \multicolumn{1}{c}{Medication}\\ \midrule
\cite{buczak2010data} & EMERGE & Rule & 2010 & \cmark & \cmark & \xmark & \xmark & \xmark & \xmark & \xmark & \xmark & \xmark & \xmark & \xmark & \xmark & - & - & - \\                           
\cite{park2013perturbed} & PeGS & non-param & 2013 & \xmark & \xmark & \xmark & \xmark & \xmark & \xmark & \xmark & \xmark & \xmark & \cmark & \xmark & \xmark & - & - & - \\
\cite{franklin2014plasmode} & Plasmode & Rule & 2014 & \cmark & \cmark & \cmark & \xmark & \xmark & \xmark &\xmark &\xmark & \xmark & \xmark & \xmark & \xmark & - & - & - \\
\cite{dube2014approach} & PADARSER/GRiSER & Rule & 2014 & \cmark & \cmark & \cmark & \xmark & \xmark & \cmark & \xmark & \xmark &\xmark &\xmark & \xmark & \xmark & - & - & - \\
\cite{mclachlan2016using} & CoMSER & Rule & 2016 & \cmark & \cmark & \xmark & \xmark & \xmark & \cmark & \xmark & \xmark &\xmark &\xmark & \xmark & \xmark & - & - & - \\
\cite{esteban2017real} & RGA/RCGAN & GAN & 2017 & \cmark & \cmark & \xmark  & \xmark & \xmark & \xmark & \cmark & \xmark & \xmark & \xmark & \cmark & \xmark & - & - & - \\
\cite{choi2017generating} & MedGAN & GAN & 2017 & \xmark & \xmark & \cmark & \xmark & \xmark & \cmark   & \cmark & \xmark & \cmark & \cmark & \cmark & \cmark & \cmark & \cmark & \cmark \\
\cite{walonoski2018synthea} & Synthea & Rule & 2018 & \cmark & \cmark & \xmark & \xmark & \xmark & \xmark & \xmark & \xmark & \xmark & \xmark & \cmark & \xmark & - & - & - \\
\cite{mclachlan2018aten} & ATEN & Rule & 2018 & \xmark & \xmark & \xmark & \xmark & \xmark & \xmark & \xmark & \xmark & \xmark & \xmark & \xmark & \xmark & - & - & - \\
\cite{baowaly2019synthesizing} & medBGAN/medWGAN & GAN & 2019 & \xmark & \xmark & \cmark & \cmark  & \xmark & \xmark & \cmark & \xmark & \xmark & \xmark & \cmark & \cmark & \cmark & \cmark & \xmark \\
\cite{yang2019grouped} & GcGAN & GAN & 2019 & \xmark & \xmark & \cmark & \cmark & \xmark & \xmark & \cmark & \xmark & \xmark & \xmark & \xmark & \xmark & - & - & - \\
\cite{yale2020generation} & HealthGAN & GAN & 2020 & \xmark & \xmark & \xmark  & \xmark   & \xmark & \xmark & \xmark & \xmark & \xmark & \xmark & \cmark & \cmark  & \cmark & \xmark & \cmark  \\
\cite{lee2020generating} & DAAE & GAN & 2020 & \xmark & \cmark & \xmark  & \xmark   & \cmark & \cmark & \xmark & \xmark & \xmark & \xmark & \cmark & \cmark & \cmark & \xmark & \xmark \\
\cite{zhang2020ensuring} & EMR-WGAN/EMR-CWGAN & GAN & 2020 & \cmark & \xmark & \cmark & \xmark & \xmark & \xmark & \xmark & \xmark & \cmark & \cmark & \xmark & \xmark & - & - & - \\
\cite{yan2020generating} & HGAN & GAN & 2020 & \cmark & \xmark & \cmark & \cmark  & \xmark & \xmark & \xmark & \xmark & \cmark & \cmark & \xmark & \xmark & - & - & - \\
\cite{torfi2020corgan} & CorGAN & GAN & 2020 & \xmark & \xmark & \cmark & \xmark & \xmark & \xmark & \cmark & \xmark & \cmark & \xmark & \cmark & \cmark & \cmark & \cmark & \xmark \\
\cite{yoon2020anonymization} & ADS-GAN & GAN & 2020 & \xmark & \xmark & \cmark & \cmark & \xmark & \xmark & \cmark & \xmark & \xmark & \xmark & \cmark & \xmark & - & - & - \\
\cite{rashidian2020smooth} & SMOOTH-GAN & GAN & 2020 & \cmark & \xmark & \cmark & \cmark & \xmark & \xmark & \cmark & \xmark & \xmark & \xmark & \xmark & \xmark & - & - & - \\
\cite{kaur2021application} & Bayesian Network & Bayesian Network & 2021 & \cmark & \xmark & \cmark & \cmark  & \cmark & \xmark & \cmark & \xmark & \cmark & \cmark & \xmark & \cmark & \cmark & \xmark & \xmark \\
\cite{zhang2021synteg} & SynTEG & GAN & 2021 & \cmark & \cmark & \cmark & \xmark & \xmark & \xmark & \xmark & \xmark & \cmark & \cmark & \xmark & \xmark & - & - & - \\
\cite{biswal2021eva} & EVA & VAE & 2021 & \cmark & \cmark & \xmark & \xmark & \xmark & \cmark   & \xmark & \xmark & \cmark & \xmark & \xmark & \xmark & - & - & - \\
\cite{wang2022promptehr} & PromptEHR & Transformer & 2022 & \cmark & \cmark  & \xmark  & \xmark   & \xmark & \xmark & \xmark & \xmark & \cmark & \cmark & \cmark  & \cmark & \cmark & \cmark & \cmark \\
\cite{sun2021generating} & LongGAN & GAN & 2022 & \cmark & \cmark  & \xmark  & \xmark   & \xmark & \xmark & \cmark & \xmark & \xmark & \cmark & \xmark  & \xmark & - & - & - \\
\textcolor{black}{\cite{bing2022conditional}} & \textcolor{black}{HealthGen} & \textcolor{black}{VAE} & \textcolor{black}{2022} & \textcolor{black}{\cmark} & \textcolor{black}{\cmark}  & \textcolor{black}{\xmark}  & \textcolor{black}{\xmark}   & \textcolor{black}{\xmark} & \textcolor{black}{\xmark} & \textcolor{black}{\cmark} & \textcolor{black}{\cmark} & \textcolor{black}{\xmark} & \textcolor{black}{\xmark} & \textcolor{black}{\cmark}  & \textcolor{black}{\cmark} & \textcolor{black}{\xmark} & \textcolor{black}{\xmark} & \textcolor{black}{\xmark} \\
\textcolor{black}{\cite{SUN2023104404}} & \textcolor{black}{DP-CGANS} & \textcolor{black}{GAN} & \textcolor{black}{2023} & \textcolor{black}{\cmark} & \textcolor{black}{\xmark}  & \textcolor{black}{\cmark}  & \textcolor{black}{\cmark}   & \textcolor{black}{\xmark} & \textcolor{black}{\xmark} & \textcolor{black}{\cmark} & \textcolor{black}{\xmark} & \textcolor{black}{\cmark} & \textcolor{black}{\cmark} & \textcolor{black}{\cmark}  & \textcolor{black}{\xmark} & \textcolor{black}{-} & \textcolor{black}{-} & \textcolor{black}{-} \\
\textcolor{black}{\cite{shankar2023clinical}} & \textcolor{black}{Clinical-GAN} & \textcolor{black}{Transformer + GAN} & \textcolor{black}{2023} & \textcolor{black}{\cmark} & \textcolor{black}{\cmark}  & \textcolor{black}{\xmark}  & \textcolor{black}{\xmark}   & \textcolor{black}{\xmark} & \textcolor{black}{\xmark} & \textcolor{black}{\xmark} & \textcolor{black}{\xmark} & \textcolor{black}{\xmark} & \textcolor{black}{\xmark} & \textcolor{black}{\cmark}  & \textcolor{black}{$\text{\cmark}^{\dag}$} & \textcolor{black}{\cmark} & \textcolor{black}{\cmark} & \textcolor{black}{\cmark} \\
\textcolor{black}{\cite{lu2023multi}} & \textcolor{black}{MT-GAN} & \textcolor{black}{GAN} & \textcolor{black}{2023} & \textcolor{black}{\cmark} & \textcolor{black}{\cmark}  & \textcolor{black}{\xmark}  & \textcolor{black}{\xmark}   & \textcolor{black}{\xmark} & \textcolor{black}{\xmark} & \textcolor{black}{\xmark} & \textcolor{black}{\cmark} & \textcolor{black}{\xmark} & \textcolor{black}{\xmark} & \textcolor{black}{\cmark}  & \textcolor{black}{\cmark} & \textcolor{black}{\cmark} & \textcolor{black}{\xmark} & \textcolor{black}{\xmark} \\
\cite{nikolentzos2023synthetic} & Graph-VAE & VAE (Graph) & 2023 & \xmark & \cmark & \cmark & \xmark   & \xmark & \xmark & \cmark & \xmark & \xmark & \xmark & \xmark & \xmark & - & - & - \\
\cite{li2023generating} & EHR-M-GAN & GAN & 2023 & \cmark & \cmark  & \cmark & \cmark  & \cmark & \xmark & \cmark & \cmark & \cmark & \xmark & \cmark & \cmark & - & - & - \\
\cite{ceritli2023synthesizing} & EHR-TabDDPM & Diffusion & 2023 & \xmark & \xmark & \cmark & \xmark & \xmark & \xmark & \cmark & \cmark & \cmark & \xmark & \xmark & \cmark & \cmark & \xmark & \xmark \\
\cite{he2023meddiff} & MedDiff & Diffusion & 2023 & \cmark & \xmark & \cmark & \cmark  & \xmark & \xmark & \xmark & \xmark & \xmark & \xmark & \xmark & \cmark & \cmark & \xmark & \xmark \\
\cite{mosquera2023method} & LSTM-EHR & LSTM & 2023 & \cmark & \cmark  & \cmark  & \cmark   & \xmark & \xmark & \xmark & \xmark & \xmark & \cmark & \xmark  & \xmark & - & - & - \\
\cite{naseer2023scoehr} & ScoEHR & Diffusion & 2023 & \xmark & \xmark & \cmark & \cmark & \xmark & \cmark & \xmark & \xmark & \cmark & \xmark & \cmark  & \cmark & \cmark & \xmark & \xmark  \\
\cite{theodorou2023synthesize} & HALO & Transformer & 2023 & \cmark & \cmark & \cmark & \cmark & \xmark & \xmark & \cmark & \cmark & \cmark & \cmark & \cmark  & \cmark & \cmark & \cmark & \cmark \\
\cite{tian2023fast} & TimeDiff & Diffusion & 2023 & \cmark & \cmark & \xmark & \xmark & \cmark & \xmark & \cmark & \cmark & \cmark & \xmark & \cmark  & \cmark & \xmark & \xmark & \xmark \\
\cite{yoon2023ehr} & EHR-Safe & GAN & 2023 & \xmark & \cmark & \cmark & \xmark & \xmark & \xmark & \cmark & \xmark & \cmark & \cmark & \xmark  & \cmark & \xmark & \xmark & \xmark \\
\cite{yuan2023ehrdiff} & EHRDiff & Diffusion & 2024 & \xmark & \xmark & \cmark & \cmark & \xmark & \xmark & \cmark & \xmark & \cmark & \cmark & \cmark & \cmark  & \cmark & \cmark & \xmark \\
\cite{sun2024collaborative} & MSIC & VAE & 2024 & \cmark & \cmark & \cmark & \cmark  & \xmark & \xmark & \xmark & \xmark & \xmark & \xmark & \cmark & \cmark & \cmark & \xmark & \cmark \\
\cite{chen2024guided} & EHR-D3PM & Diffusion & 2024 & \cmark & \xmark & \cmark & \cmark  & \xmark & \xmark & \cmark & \xmark & \cmark & \xmark & \xmark & \cmark & \cmark & \cmark & \xmark \\
\cite{pang2024cehr} & CEHR-GPT & Transformer & 2024 & \xmark & \cmark & \cmark & \cmark  & \xmark & \xmark & \xmark & \xmark & \cmark & \cmark & \xmark & \xmark & - & - & - \\ 
\cite{heflexible} & FlexGen-EHR & Diffusion & 2024 & \cmark & \cmark & \cmark & \xmark & \xmark & \xmark & \cmark & \xmark & \cmark & \xmark & \xmark & \cmark & - & - & - \\
\cite{ramachandranpillai2024bt} & Bt-GAN & GAN & 2024 & \xmark & \xmark & \xmark & \xmark & \xmark & \xmark & \cmark & \xmark & \xmark & \xmark & \xmark & \cmark & - & - & - \\
\cite{wang2024igamt} & IGAMT & GAN & 2024 & \cmark & \cmark & \xmark & \xmark & \xmark & \xmark & \cmark & \xmark & \xmark & \xmark & \cmark & \; \cmark$^\dag$ & - & - & - \\
\cite{vardhan2024large} & LLM-EHR & Transformer/LLM & 2024 & \xmark & \xmark & \cmark & \cmark & \xmark & \xmark & \xmark & \xmark & \xmark & \xmark & \xmark & \cmark & \cmark & \cmark & \xmark\\
\cite{gwon2024ldp} & LDP-GAN & GAN & 2024 & \xmark & \xmark & \cmark & \cmark & \xmark & \xmark & \cmark & \xmark & \cmark & \xmark & \cmark & \xmark & - & - & - \\
\cite{zhong2024synthesizing} & EHRPD & Diffusion & 2024 & \cmark & \cmark & \cmark & \cmark & \xmark & \xmark & \cmark & \cmark & \xmark & \xmark & \cmark & \cmark & \cmark & \cmark & \cmark \\ 
\textcolor{black}{\cite{lee2024leveraging}} & \textcolor{black}{CodeAR} & \textcolor{black}{VAE} & \textcolor{black}{2024} & \textcolor{black}{\cmark} & \textcolor{black}{\cmark}  & \textcolor{black}{\cmark}  & \textcolor{black}{\xmark}   & \textcolor{black}{\xmark} & \textcolor{black}{\xmark} & \textcolor{black}{\cmark} & \textcolor{black}{\xmark} & \textcolor{black}{\cmark} & \textcolor{black}{\xmark} & \textcolor{black}{\cmark}  & \textcolor{black}{\cmark} & \textcolor{black}{\xmark} & \textcolor{black}{\xmark} & \textcolor{black}{\xmark} \\ 
\textcolor{black}{\cite{karami2024timehr}} & \textcolor{black}{TimEHR} & \textcolor{black}{GAN} & \textcolor{black}{2024} & \textcolor{black}{\cmark} & \textcolor{black}{\cmark}  & \textcolor{black}{\xmark}  & \textcolor{black}{\cmark}   & \textcolor{black}{\xmark} & \textcolor{black}{\xmark} & \textcolor{black}{\cmark} & \textcolor{black}{\xmark} & \textcolor{black}{\cmark} & \textcolor{black}{\xmark} & \textcolor{black}{\cmark}  & \textcolor{black}{\cmark} & \textcolor{black}{\xmark} & \textcolor{black}{\xmark} & \textcolor{black}{\xmark} \\ \bottomrule
\end{tabular}}
\end{table}
\end{landscape}

\begin{itemize}
    \item \textit{Rule-based method}: These approaches generate synthetic EHR data using predefined algorithms and rules, often incorporating real-world disease prevalence and national census data as hyperparameters to simulate disease occurrence and population demographics. From our review, 6 out of \textcolor{black}{48} methods are rule-based methods\cite{buczak2010data, franklin2014plasmode, dube2014approach, mclachlan2016using, mclachlan2018aten}.

    \item \textit{GAN-based method}:  With the rise of deep learning, GANs have become widely adopted for synthetic EHR generation. \textcolor{black}{21} out of \textcolor{black}{48} methods are GAN-based methods\cite{baowaly2019synthesizing, yang2019grouped, yale2020generation, lee2020generating, zhang2020ensuring, yan2020generating, torfi2020corgan, yoon2020anonymization, rashidian2020smooth, zhang2021synteg, sun2021generating, li2023generating, ramachandranpillai2024bt, wang2024igamt, gwon2024ldp}.

    \item \textit{VAE-based method}: As another strand of deep generative modeling that are popular in the artificial intelligence (AI) community, VAEs are applied to generate synthetic EHR data from latent embeddings\cite{biswal2021eva, bing2022conditional, lee2024leveraging}. Our review shows that \textcolor{black}{5} out of \textcolor{black}{48} methods apply VAEs to generate synthetic EHR data.

    \item \textit{Transformer-based method}: Transformers\cite{vaswani2017attention}, known for their ability to process sequential data\cite{li2019neural} and their success in natural language processing\cite{devlin2018bert, brown2020gpt3}, have also been explored for longitudinal synthetic EHR generation.
    In our review, \textcolor{black}{5} out of \textcolor{black}{48} methods learn to generate synthetic EHR data with the transformer structures.
    
    \item \textit{Diffusion-based method}: Most recently, diffusion models\cite{sohl2015deep, ho2020denoising}, which have demonstrated strong performance in image and video generation\cite{ho2022imagen, ho2022video, saharia2022photorealistic}, have been applied for synthetic EHR generation. 8 out of the \textcolor{black}{48} studies have proposed diffusion-based models for this purpose\cite{ceritli2023synthesizing, he2023meddiff, naseer2023scoehr, tian2023fast, yuan2023ehrdiff, chen2024guided, heflexible, zhong2024synthesizing}.
\end{itemize}

In addition to the five main categories above, conventional statistical methods were also used to generate synthetic EHR data, for example,  \textcolor{black}{Gibbs sampling-based algorithms\cite{park2013perturbed}};  \textcolor{black}{Bayesian Networks\cite{kaur2021application}. Additionally, other neural network architectures are also explored\cite{mosquera2023method}.} 

Based on these literature, we identify three key functionalities that are critical for the design of synthetic data generation algorithms, which we leave in Section~\ref{secsupp:results}.


\subsection{\textcolor{black}{Methods selected for benchmarking}}
\label{sec:method-selected-for-benchmarking}

\textcolor{black}{Despite the recent development of synthetic EHR methods, \textcolor{black}{31} out of \textcolor{black}{48} studies \textcolor{black}{ identified in our scoping review} remain closed-source \textcolor{black}{in terms of the codebases} at the time of this scoping review (Table~\ref{tab:literature}).} \textcolor{black}{To provide a benchmarking, we select seven generation methods with high impact and publicly available repositories with annotated code and detailed instruction to enable implementation on the MIMIC datasets without significantly additional efforts, chosen from the five identified methodological categories identified in our scoping review.} Specifically, for rule-based methods, we select Plasmode\cite{franklin2014plasmode} and Synthea\cite{walonoski2018synthea}; for GAN-based methods, we choose MedGAN\cite{choi2017generating} and CorGAN\cite{torfi2020corgan}; for VAE-based methods, we use a baseline implemented by CorGAN's authors due to the lack of easily reproducible codebases in existing VAE-based methods; for transformer-based methods, we select PromptEHR\cite{wang2022promptehr}; and for diffusion-based methods, we choose EHRDiff\cite{yuan2023ehrdiff}. For all the existing methods, we follow their default settings, except necessary minor changes for debugging.

\textcolor{black}{Note that among the included methods, PromptEHR and Synthea generate EHR data longitudinally. Specifically, PromptEHR is a Transformer-based model that generates subsequent visits conditional on the sequence of prior visits; Synthea simulates full patient trajectories using disease progression modules. For evaluation, we generate data longitudinally and then aggregate it into binary phenotypes-labeling a disease as 1 if it occurs at least once. Summary statistics for the number of visits generated in the longitudinal synthetic data can be found in Table~\ref{tab:longitudinal_generation_statistics} in the Supplementary Materials.}

Additionally, we include two simple baselines for comparison. The first baseline, \textit{Prevalence-based Random (PBR)}, randomly generates synthetic EHR data based on the estimated marginal phenotype prevalences without taking phenotype correlation into account. The second baseline, \textit{Resample}, bootstraps the real EHR dataset to ``synthesize'' data. We leave the detailed description of the code preprocessing pipeline including mapping to a common \textcolor{black}{coding} system in Section~\ref{secsupp:method}.

\subsection{Evaluation results}
\label{sec:evaluation-results}

In this subsection, we introduce the benchmarking results of key evaluation metrics. Full descriptions, including the results of computational cost \textcolor{black}{(Table~\ref{tab:computational-cost})} and numerical studies \textcolor{black}{(Figure~\ref{fig:gen_sample_size},~\ref{fig:train_size})}, \textcolor{black}{sensitivity analysis on coding systems, and sub-experiment results on evaluating the methods in different sub-populations,} are presented in Section~\ref{secsupp:results} \textcolor{black}{of the Supplementary Materials}.

\vspace{0.5cm}

\noindent
\textit{Fidelity}

\begin{figure}[!t]
    \centering
    \includegraphics[width=0.6\linewidth]{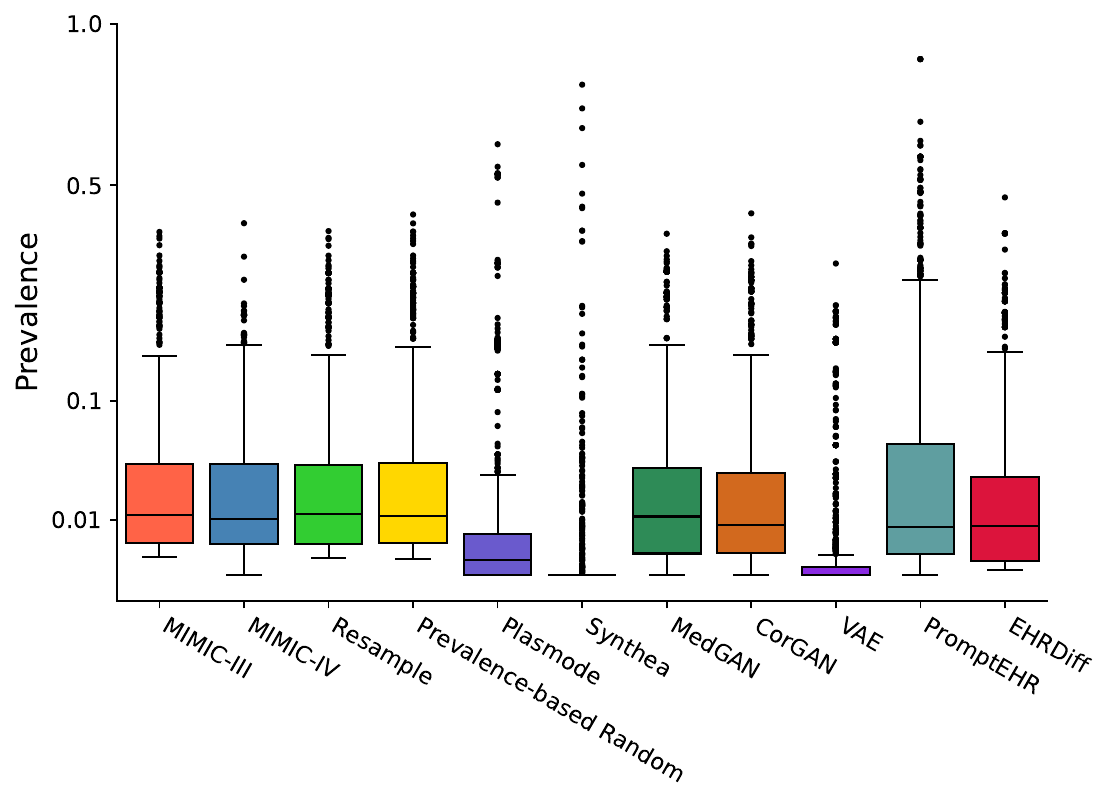}
    \caption{Boxplots of the estimated phecode-wise prevalences (Y-axis) across distinct phecodes in the real datasets or synthetic datasets generated from the selected methods shown on the X-axis. Only the $K = 1,773$ distinct phecodes occurring in more than 50 patients in MIMIC-III are included.}
    \label{fig:prevalence-boxplot}
\end{figure}

\begin{figure}[H]
    \centering
    \includegraphics[width=0.6\linewidth]{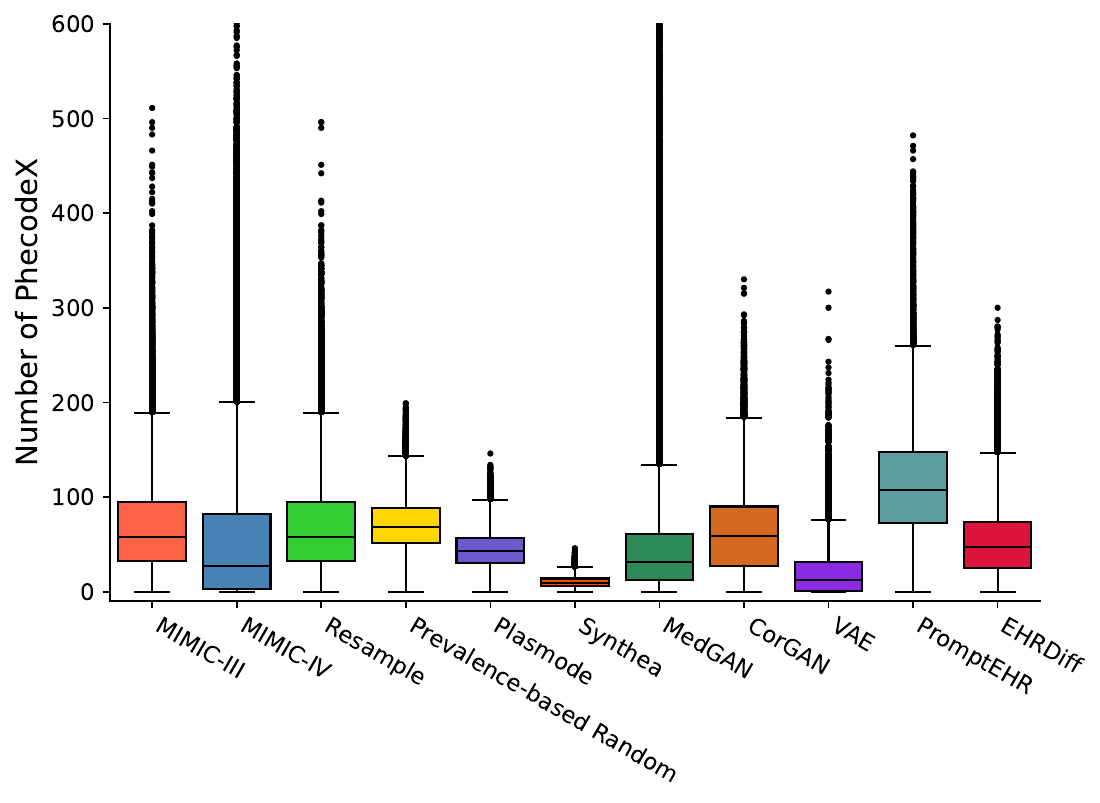}
    \caption{Boxplots of the number of unique phecodes per patient (Y-axis) based on the real datasets or synthetic datasets generated from selected methods shown on X-axis. Only the $K = 1,773$ distinct phecodes occurring in more than 50 patients in MIMIC-III are included.}
    \label{fig:num_of_phecode-boxplot}
\end{figure}

Figure~\ref{fig:prevalence-boxplot} \textcolor{black}{presents boxplots of the marginal prevalences in the real and synthetic EHR data.} The \textit{PBR} baseline, MedGAN, CorGAN, and EHRDiff \textcolor{black}{effectively preserve the prevalence distribution in MIMIC-III, while Plasmode, Synthea, and VAE consistently underestimate it. The MIMIC-IV dataset exhibits a similar prevalence distribution to MIMIC-III, though with a slightly lower median prevalence across all phenotypes.}

Figure~\ref{fig:num_of_phecode-boxplot} displays the number of unique phecodes per patient in EHR data. \textcolor{black}{CorGAN closely matches MIMIC-III, whereas most other methods underestimate the number of unique phecodes. In contrast, PromptEHR consistently overestimates it.}

\begin{figure}[!t]
    \centering
    \includegraphics[width=0.6\linewidth]{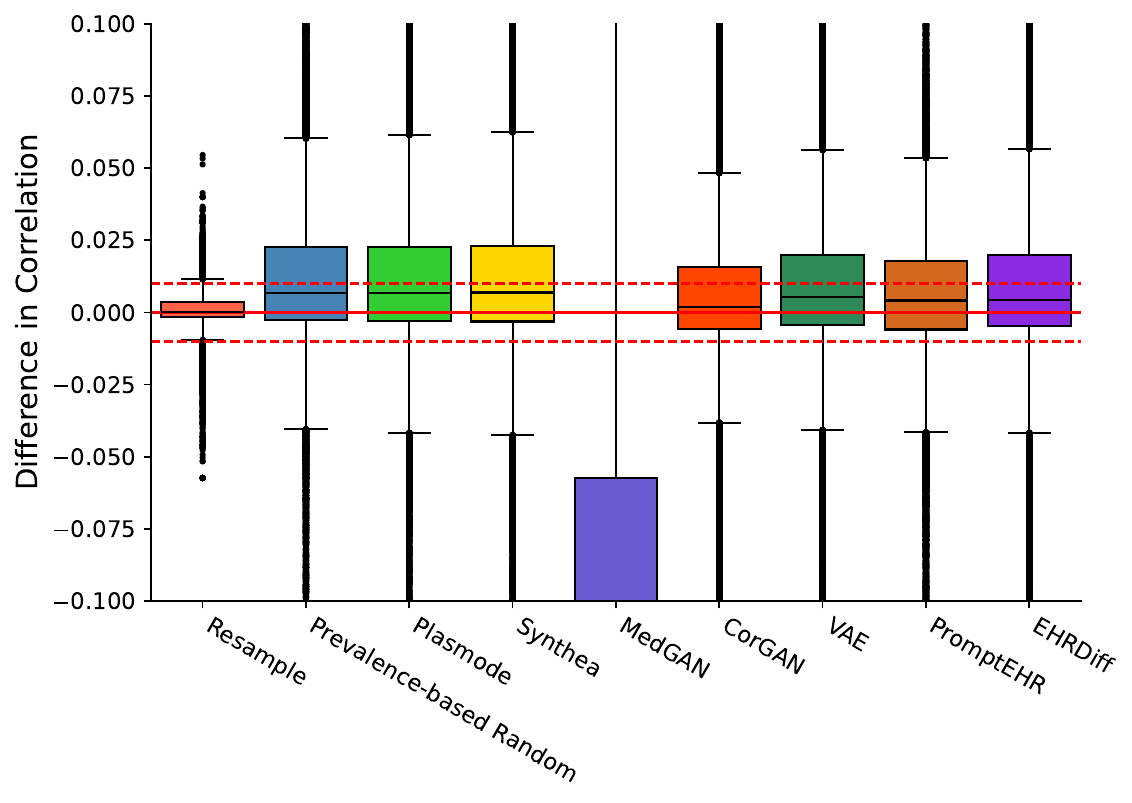}
    \caption{Boxplots of the differences in the estimated Pearson correlation for $100,000$ randomly selected pairs of phecodes (out of all ${1,773\choose2}$ pairs) comparing the ones obtained from synthetic data with those from MIMIC-III for each of the 9 selected methods. The solid red line represents an exact match in correlation; the area within the red dotted lines indicates a correlation difference of less than 0.01.}
    \label{fig:mimic3-corr-diff}
\end{figure}

\begin{figure}[!t]
    \centering
    \includegraphics[width=0.6\linewidth]{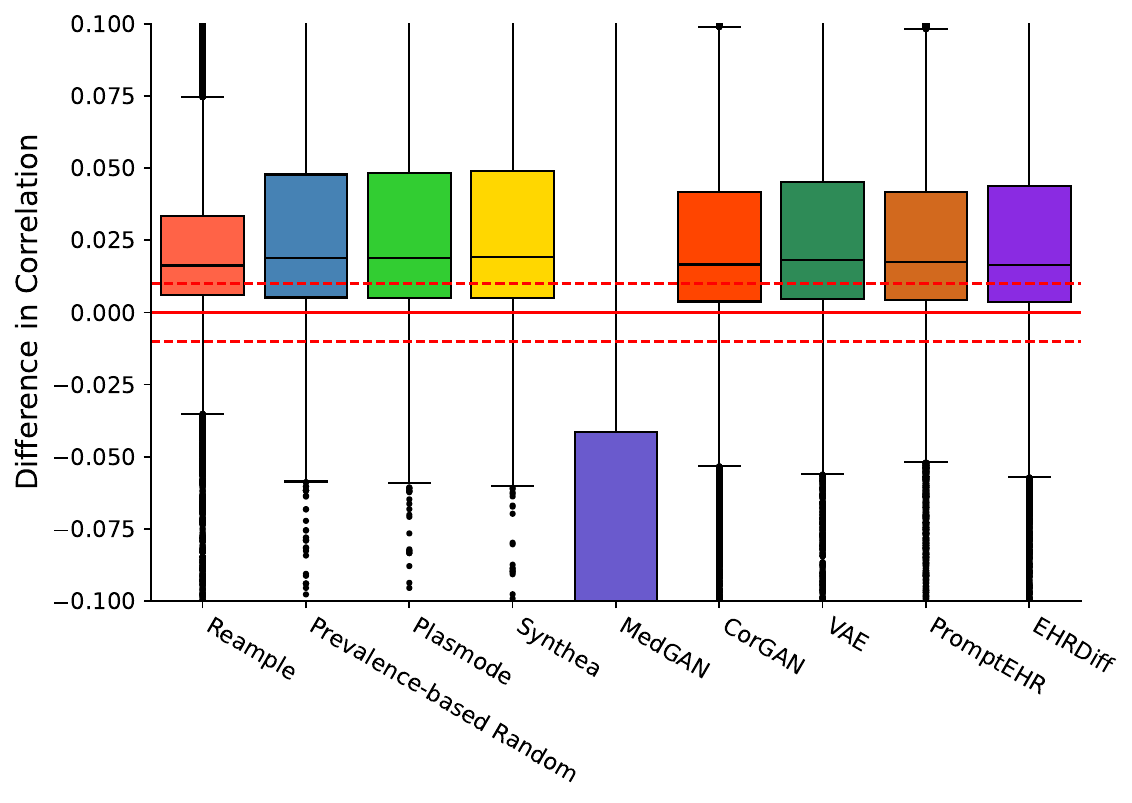}
    \caption{Boxplots of the differences in the estimated Pearson correlation for $100,000$ randomly selected pairs of phecodes (out of all ${1,773\choose2}$ pairs) comparing the ones obtained from synthetic data with those from MIMIC-IV for each of the 9 selected methods. The solid red line represents an exact match in correlation; the area within the red dotted lines indicates a correlation difference of less than 0.01.}
    \label{fig:mimic4-corr-diff}
\end{figure}

\textcolor{black}{For bivariate relationships preservation,} Figure~\ref{fig:mimic3-corr-diff},~\ref{fig:mimic4-corr-diff} shows the boxplots of the element-wise differences in the correlation matrix. \textcolor{black}{Most methods overestimate correlations, except MedGAN, which underestimates them. This overestimation is even more pronounced in MIMIC-IV, where all methods, except MedGAN, consistently overestimate the correlations.}

\begin{landscape}
\begin{table}[]
\centering
\caption{Quantitative evaluation results on synthetic data quality when evaluated using the MIMIC-III data (top block) or MIMIC-IV data (bottom block). For fidelity, Maximum Mean Discrepancy (MMD),  Root Mean Squared Percentage Error (RMSPE), Mean Absolute Percentage Error (MAPE), Correlation Frobenius Distance (CFD), Co-occurrence matrix Frobenius distance (COFD), and the Area Under the Curve (AUC) and Accuracy (ACC) of discriminative prediction are reported. For utility, the performance gap of AUC and ACC between Train on Synthetic, Test on Real (TSTR) and Train on Real, Test on Real (TRTR) are reported, and similarly for the Train on Synthetic + Real, Test on Real (TSRTR) one. For privacy, the mean and median of minimum Euclidean distance between each real medical record and the synthetic EHR dataset are reported for Membership Inference Risk (MIR); the F1 score of predictive performance of 1-Nearest Neighbor (1-NN) are reported for Attribute Inference Risk (AIR). \reddown \; represents the lower the better; \greenup \; represents the higher the better. For each metric, the overall top two methods are  in \textbf{black boldface}; the best method excluding {\it Resample} and {\it Prevalence-based Random} is in \bluebf{blue boldface}. \redbf{Red boldface} indicates the method is in overall top-two and also ranks the best among all but {\it Resample} and {\it Prevalence-based Random}.}
\label{tab:quantitative}
\resizebox{1.0\columnwidth}{!}{%
\begin{tabular}{l|ccccccc|cccc|ccc} \toprule
                          & \multicolumn{7}{c|}{Fidelity}                           & \multicolumn{4}{c|}{Utility}     & \multicolumn{3}{c}{Privacy} \\
 &
  \multicolumn{1}{c}{MMD} &
  \multicolumn{1}{c}{RMSPE} &
  \multicolumn{1}{c}{MAPE} & 
  \multicolumn{1}{c}{CFD} &
  \multicolumn{1}{c}{COFD} &
  \multicolumn{1}{c}{AUC} &
  \multicolumn{1}{c|}{ACC} &
  \multicolumn{1}{c}{TSTR (AUC)} &
  \multicolumn{1}{c}{TSTR (ACC)} &
  \multicolumn{1}{c}{TSRTR (AUC)} &
  \multicolumn{1}{c|}{TSRTR (ACC)} &
  \multicolumn{1}{c}{MIR (mean)} &
  \multicolumn{1}{c}{MIR (median)} &
  \multicolumn{1}{c}{AIR} \\
 & \reddown & \reddown & \reddown & \reddown & \reddown & \reddown & \reddown & \greenup & \greenup & \greenup & \greenup & \greenup & \greenup & \reddown \\ \midrule
\multicolumn{15}{c}{MIMIC-III} \\ \midrule
\rowcolor{gray!50} \textit{Resample} & \textbf{0.0040} & \textbf{0.17} & \textbf{7.18}     & \textbf{4.42}  & \textbf{29.50}  & \textbf{0.476}  & \textbf{0.500}  & \textbf{0.007}   & \textbf{0.005}   & \textbf{0.008}  & \textbf{0.006}  & 0.97    & 0.00    & 0.952   \\
\textit{Prevalence-based Random} & \textbf{0.0044} & \textbf{0.14}    & \textbf{6.99}     & 46.94  & 114.11 & \textbf{0.494}  & \textbf{0.421}  & -0.287 & -0.275 & -0.014 & -0.010 & \textbf{5.47}    & \textbf{5.29}    & 0.622   \\ \midrule
Synthea & 0.7206 & 11.56 & 222.96 & 47.34 & 469.45 & 0.999  & 0.995  & -0.149 & -0.115 & -0.004 & -0.002 & \redbf{4.86} & \redbf{4.80} & \redbf{0.571}   \\ 
Plasmode & 0.5945 & 127.41 & 1951.77 & 39.01 & 596.81 & 0.999 & 0.996 & -0.268 & -0.333 & \redbf{0.0003} & 0.001 & 4.53 & 4.58  & \textbf{0.595} \\ \midrule
MedGAN                    & 0.1430 & 0.58    & 46.85    & 120.89 & 252.11 & 0.913  & 0.840  & \redbf{-0.063}  & \redbf{-0.038}  & -0.003 & \redbf{0.003} & 2.93    & 2.83    & 0.874   \\
CorGAN                    & \bluebf{0.0478} & \bluebf{0.57}    & \bluebf{44.30}    & \redbf{22.20}  & \redbf{84.48} & \bluebf{0.720}  & \bluebf{0.669}  & -0.078  & -0.056  & -0.020 & -0.013 & 3.06 & 3.16 & 0.838   \\ \midrule
VAE & 0.3083 & 0.94    & 92.67    & 44.05  & 345.86 & 0.984  & 0.943  & -0.086  & -0.063  & -0.010 & -0.002 & 3.88    & 4.00    & 0.845   \\ \midrule
PromptEHR & 0.8593 & 2.47    & 91.31    & 24.03  & 762.25 & 0.990  & 0.957  & -0.081  & -0.049  & -0.014 & -0.011 & 3.48    & 3.46    & 0.839   \\ \midrule
EHRDiff & 0.3381 & 18.30 & 245.55 & 31.98 & 337.31 & 0.995 & 0.967 & -0.178 & -0.216 & -0.029 & -0.024 & 3.66 & 3.61 & 0.737 \\ \midrule
\multicolumn{15}{c}{MIMIC-IV} \\ \midrule

\rowcolor{gray!50} \textit{Resample}  & 0.2363 & 1047.70 & 11849.43 & \textbf{30.15} & 1103.51 & \textbf{0.898} & \textbf{0.854} & \textbf{-0.066} & \textbf{-0.028} & -0.0009  & \textbf{0.005}  & 2.58 & 2.00 & 0.766 \\
\textit{Prevalence-based Random} & 0.3740 & 1092.77 & 12540.96 & 44.92 & 1153.67 & 0.956  & 0.914  & -0.254 & -0.168 & -0.005 & -0.001 & 2.96 & 2.45 & 0.694   \\ \midrule 
Synthea & 0.7216 & 727.39 & \textbf{5689.49} & 58.61 & 1491.42 & 0.999 & 0.993 & -0.127 & -0.092 & -0.001 & -0.002 & \textbf{3.80} & \redbf{3.46} & \textbf{0.597} \\ 
Plasmode & 0.5970 & 5844.53 & 84377.27  & 53.23 & 1535.78 & 0.999 & 0.999 & -0.218 & -0.206 & \redbf{0.000} & -0.0002 & \redbf{3.98} & \redbf{3.46} & \redbf{0.497} \\ \midrule
MedGAN & 0.2377 & 797.89 & 9259.80 & 107.59 & \textbf{1017.83} & 0.962 & 0.916 & \redbf{-0.076}  & \redbf{-0.059}  & \redbf{-0.0007} & \redbf{0.004} & 2.83 & 2.45 & 0.714 \\
CorGAN & \redbf{0.2335} & 959.54 & 10549.75 & \redbf{38.92} & 1152.84 & \redbf{0.927} & \redbf{0.883}  & -0.131  & -0.081  & -0.004 & 0.001 & 2.86 & 2.45 & 0.732   \\ \midrule
VAE & \textbf{0.2360} & \textbf{680.68} & 6566.97 & 53.89 & 1430.53 & 0.978  & 0.923 & -0.123 & -0.074 & -0.004 & \redbf{0.004} & 3.44 & 3.16 & 0.765 \\ \midrule
PromptEHR & 0.8686 & \redbf{113.46} & \redbf{1418.71} & 39.09  & \redbf{887.76} & 0.995 & 0.980  & -0.117  & -0.066  & -0.003 & \redbf{0.004} & 3.09 & 2.83 & 0.750 \\ \midrule
EHRDiff & 0.3176 & 1274.05 & 14207.65 & 44.11 & 1309.36 & 0.995 & 0.981 & -0.218 & -0.152 & -0.007 & 0.0008 & 3.04 & 2.65 & 0.675 \\ \bottomrule
\end{tabular}%
}
\end{table}
\end{landscape}

Table~\ref{tab:quantitative} presents the quantitative evaluation of key fidelity metrics. \textit{PBR} outperforms other methods \textcolor{black}{on most} metrics, except for its higher CFD score (46.9), \textcolor{black}{which reflects its inability to model bivariate correlations by design.} CorGAN preserves distributional information best among the selected methods, particularly excelling in CFD due to its design for modeling associations.

When evaluated on the MIMIC-IV data, CorGAN outperforms \textcolor{black}{the baselines} in preserving the phecode-wise prevalence, achieving an MMD of 0.234 for absolute differences. \textcolor{black}{However, PromptEHR performs best in relative difference metrics} (RMSPE: 113.5; MAPE: 1418.7). Additionally, \textcolor{black}{since synthetic data were trained on MIMIC-III, it is significantly easier to distinguish them from real MIMIC-IV data. Among all the selected methods, CorGAN remains the most effective, achieving lowest AUC of 0.927.}

\vspace{0.5cm}

\noindent
\textit{Utility}

\begin{table}[!t]
\centering
\caption{Estimated $\beta$ from the real data and synthetic data generated by selected methods in two bivariate logistic regression models: 1) Obesity (\texttt{EM\_236}) v.s. with any type of cnacers (all phecodeX starting with ``\texttt{CA}''); 2) Diabetes (\texttt{EM\_202}) v.s. Hypertension (\texttt{CV\_401}). Synthetic data generated by VAE fails to converge for the first task as no patients with obesity and absent any type of cancer are generated.}
\label{tab:analytical-utility}
\resizebox{0.85\columnwidth}{!}{%
\begin{tabular}{c|l|cc|cc} \toprule
& & \multicolumn{2}{c}{Obesity v.s. All Cancers} & \multicolumn{2}{|c}{Diabetes v.s. Hypertension} \\
 & & $\hat{\beta}$ & 95\% CI  & $\hat{\beta}$ & 95\% CI       \\ \midrule
\multirow{2}{*}{Real Data} & MIMIC-III     & 0.871                              & (0.782 0.960)   & 1.219                              & (1.174 1.263)  \\
 & MIMIC-IV  &  0.810 & (0.782 0.838) & 1.412 & (1.388 1.437)   \\ \midrule
\multirow{9}{*}{Synthetic Data} & \textit{Resample}        & 0.868                              & (0.781 0.955)  & 1.194                              & (1.151 1.237)  \\
& \textit{Prevalence-based Random} & -0.125                             & (-0.316 0.066) & 0.003                              & (-0.038 0.043) \\ \cmidrule(r){2-6}
& Synthea & 1.685 & (1.614 1.756) & 3.590 & (3.512 3.668) \\
& Plasmode      & -0.039                             & (-0.143 0.066) & 0.027                              & (-0.118 0.171) \\  \cmidrule(r){2-6}
& MedGAN        & 3.636                              & (3.453 3.819)  & 0.864                              & (0.822 0.905)  \\
& CorGAN        & 1.080                               & (0.970 1.191)   & 1.165                              & (1.120 1.209)  \\  \cmidrule(r){2-6}
& VAE & - & - & 3.190                              & (3.063 3.318)  \\  \cmidrule(r){2-6}
& PromptEHR     & 0.514                              & (0.420 0.608)   & 0.597                              & (0.553 0.642)  \\  \cmidrule(r){2-6}
& EHRDiff       & 0.565                              & (0.527 0.603)  & 0.946                              & (0.891 1.000)  \\ \bottomrule
\end{tabular}%
}
\end{table}

We first \textcolor{black}{assess} the analytical utility of the synthetic EHR data. Table~\ref{tab:analytical-utility} presents the logistic regression results. In the Obesity vs. All Cancers \textcolor{black}{setting, the estimated log-odds ratios for the real data are} 0.87 (95\% CI: [0.78, 0.96]) in MIMIC-III and 0.81 (95\% CI: [0.78, 0.84]) in MIMIC-IV. \textcolor{black}{For Diabetes vs. Hypertension, the estimates are 1.22 (95\% CI: [1.17, 1.26]) in MIMIC-III and 1.41 (95\% CI: [1.39, 1.44]) in MIMIC-IV}. Although \textcolor{black}{the estimated directions of the two datasets are consistent, their magnitudes differ in the second task.} For the synthetic data, \textit{PBR} yields insignificant estimates for both tasks. Notably, CorGAN has estimated log-odds ratio of 1.08 (95\% CI: [0.97, 1.19]) and 1.17 (95\% CI: [1.12, 1.21]) in the two tasks, with confidence intervals substantially overlapping those of MIMIC-III in the second task. Notably, VAE encounters positivity issues in the first task, \textcolor{black}{failing} to generate obesity \textcolor{black}{cases without any type of cancer}, leading to failure in \textcolor{black}{model} convergence. This issue highlights \textcolor{black}{its limitations} in generating reliable synthetic data. 

We then examine the predictive utility of synthetic EHR data (Table~\ref{tab:quantitative}). \textcolor{black}{In TSTR, all methods show performance degradation compared to TRTR on both datasets.} MedGAN shows the smallest performance drop in AUC on both datasets, with a decrease of 0.063 on MIMIC-III and 0.076 on MIMIC-IV. In contrast, Plasmode and EHRDiff exhibit the most significant declines, with AUC decreases of 0.268 and 0.178, respectively, on MIMIC-III, and 0.218 for both on MIMIC-IV.

For TSRTR, on MIMIC-III, \textcolor{black}{Plasmode becomes competitive with MedGAN, showing minor improvements in AUC (+0.0003) and ACC (+0.003). On MIMIC-IV, MedGAN, VAE, and PromptEHR perform competitively, each achieving a marginal ACC improvement of 0.004 in the downstream task.}

\textcolor{black}{Figure~\ref{fig:tstr-code-frequency} and Table~\ref{tab:disease_comparison} presents results examining the performance in TSTR with varying prevalence of the phecodes. See Sections~\ref{secsupp:method} and~\ref{secsupp:results} for a more comprehensive discussion.}

\vspace{0.5cm}

\noindent
\textit{Privacy}

For the MIR (Table~\ref{tab:quantitative}), \textit{PBR} and Synthea demonstrate the strongest performance on MIMIC-III, because these methods do not rely on training data to generate synthetic EHRs. \textcolor{black}{Regarding the} AIR, Synthea and Plasmode perform best, with predictive performance by attackers at 0.571 and 0.595, respectively. \textit{PBR} follows closely, with a predictive performance of 0.622. Plasmode’s superior performance can be attributed to its dimension-by-dimension data generation approach, which omits the associations in the real dataset. 

When evaluated on the MIMIC-IV dataset (assuming subjects in MIMIC-IV can be attacked), Plasmode and Synthea achieve the best results in lowering the risk of both membership and attribute inference attacks, mostly consistent with findings if MIMIC-III subjects are attacked.

\subsection{SynthEHRella: A package for benchmarking synthetic EHR performance}
\label{sec:package}

\textcolor{black}{A} major contribution of this paper is to consolidate all \textcolor{black}{seven} synthetic data generation pipelines under one unified coding framework. \textcolor{black}{To achieve this, we developed ``SynthEHRella'', a Python package that integrates various generation approaches and evaluation metrics, facilitating future streamlined evaluation and methods selection. This also reduces the barrier to benchmarking multiple methods, making it a valuable tool for developing new approaches.} The package is available at \href{https://github.com/chenxran/synthEHRella}{https://github.com/chenxran/synthEHRella}.

\section{DISCUSSION}
\label{sec:discussion}

\textcolor{black}{This} scoping review and benchmarking provide several key insights. First, \textit{PBR} \textcolor{black}{best resembles real MIMIC-III data although, by design, it does not capture bivariate associations.} This finding indicates a \textcolor{black}{major gap} in synthetic data fidelity to be improved, and is consistent with previous work\cite{yan2022multifaceted}. Second, \textcolor{black}{CorGAN best preserves data distributions and analytical utility. However, it consistently underestimates prevalence (Figure~\ref{fig:phewas-boxplot})}. Third, \textcolor{black}{Synthea offers strong privacy protection by not relying on training data but offers limited utility due to its restricted disease modules.} Fourth, \textcolor{black}{GAN-based methods are the most efficient, likely due to their lightweight, non-longitudinal design.}

\begin{figure}[!tp]
    \centering
    \includegraphics[width=1.0\columnwidth]{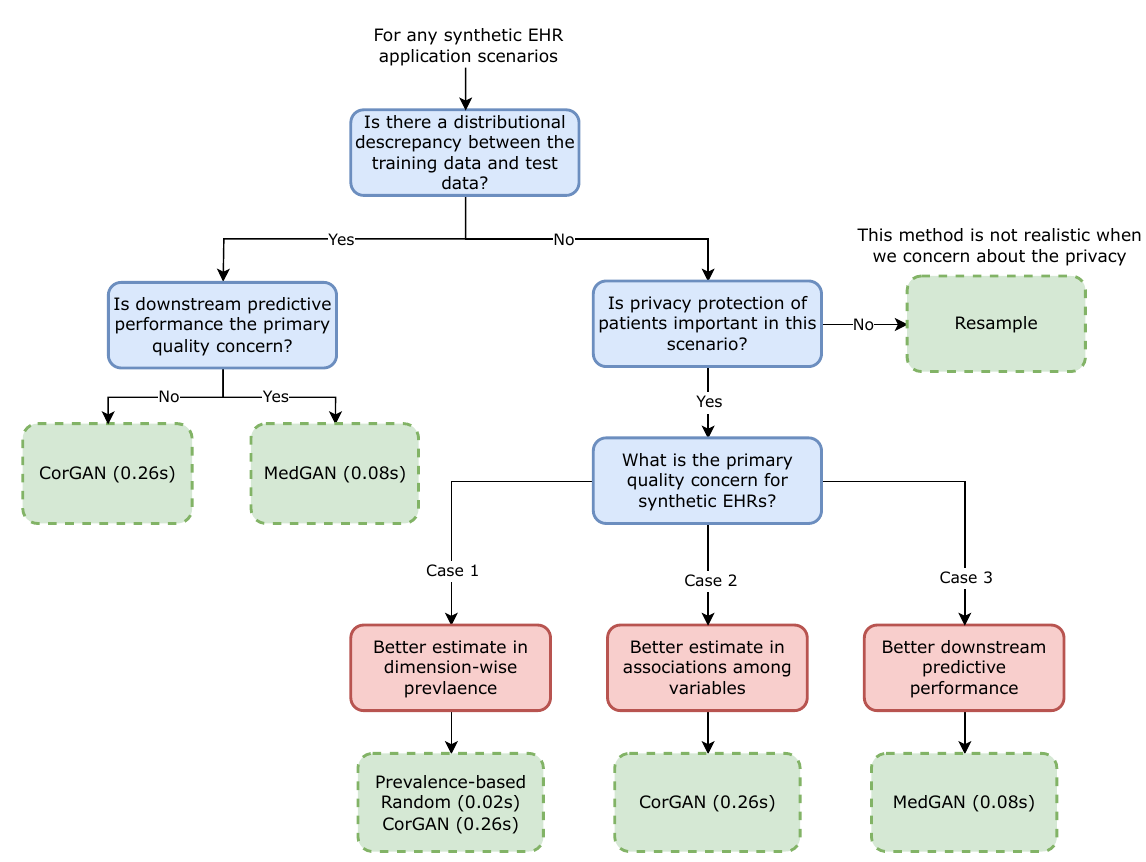}
    \caption{Decision tree for selecting synthetic EHR generation methods to generate phenotype EHR data as informed by our evaluation results of the representative methods studied in this paper. The numbers in parentheses indicate the computational time cost of corresponding methods for generating 100 samples.}
    \label{fig:decision-tree}
\end{figure}

Based on our evaluation, we offer recommendations for practitioners to select synthetic EHR generation methods \textcolor{black}{(Figure~\ref{fig:decision-tree})}. \textcolor{black}{When distributional shifts exist between training and testing samples,} CorGAN is the preferred method \textcolor{black}{unless} downstream predictive performance is the \textcolor{black}{priority}, in which case MedGAN is the best choice. When the training and the testing populations are identical, we recommend MedGAN for downstream predictive tasks, CorGAN and {\it PBR} for better estimation of dimension-wise prevalence, and CorGAN for better estimation of associations among the phenotypes. \textcolor{black}{Despite the inclusion of diverse model types, the top-performing methods are mostly GAN-based or {\it PBR}, which is consistent with previous benchmarks\cite{yan2022multifaceted}. There are two possible explanations. First, methods like Synthea and PromptEHR generate data longitudinally, which may underperform on cross-sectional phenotype tasks. Second, emerging architectures (e.g., Transformer/Diffusion models) may not yet realize their full potential. However, these methods show promise for generating multimodal EHR data.}

\textcolor{black}{Figures~\ref{fig:correlation-evaluation},~\ref{fig:correlation-evaluation-cross} further examine trade-offs between fidelity/utility and privacy, and transportability across MIMIC-III and MIMIC-IV.} The findings suggest that the trade-offs between non-privacy and privacy related metrics exist and are consistent across datasets, despite the systematic performance degradation caused by distributional differences. These align with results in the previous benchmarking study\cite{yan2022multifaceted}. See Section~\ref{secsupp:tradeoff} for detailed descriptions.

Our evaluation identifies several open opportunities. First, we showed that marginal \textit{PBR} can generate synthetic data that resemble the real data, despite failing to capture correlations. However, \textcolor{black}{even CorGAN with best fidelity performance}, consistently underestimated these prevalence, leaving gaps for improvements. Second, as even the current most complicated methods did not consider the combination of both the structured (e.g., tabular code events and lab results) and unstructured (e.g., imaging, clinical notes) data, developing methods that account for the true multimodal complexity of EHR data, taking account of the temporal nature is a pressing open problem in synthetic EHR generation. Third, as \textcolor{black}{from} Figure~\ref{fig:decision-tree}, further investigation is required to exploit \textcolor{black}{emerging 
architectures (e.g., Transformer/Diffusion models)}. Fourth, as generating synthetic multimodal EHR data with temporal features will become a common practice, \textcolor{black}{new benchmarks and evaluation metrics are needed to assess such methods effectively.} Finally, new methods are needed to incorporate formal privacy protections, e.g, differential privacy, to improve flexible control over the privacy-fidelity/utility balance in synthetic EHR generation. Other open opportunities are discussed in Section~\ref{secsupp:discussion}.

\textcolor{black}{A major limitation of our study is our focus on only cross-sectional phenotype codes, excluding longitudinal evaluation of EHR data. We also did not evaluate conditional generation based on sociodemographic variables. Despite these gaps,} our contribution is important for evaluating and inspiring further development and benchmarking of synthetic EHR methods. For example, our software platform SynthEHRella that implements all these generation methods can be extended to other data modalities and provide evaluation metrics in the expanded context. The remaining limitations of this paper are discussed in Section~\ref{secsupp:discussion}.

\section{CONCLUSION}
\label{sec:conclusion}

This paper presents a scoping review and a benchmarking of the existing representative methods with open-source codebases for generating synthetic EHR data using benchmark datasets. Under privacy concern, when distributional discrepancy exists between the training and testing samples, CorGAN is the best choice for dimension-wise prevalence estimation and downstream analytical tasks; MedGAN is recommended for training ML models for downstream use. Otherwise, when the distribution of the testing population is similar to that of the training data, MedGAN, \textit{PBR}, and CorGAN are preferred choices for predictive modeling, dimension-wise prevalence estimation, and modeling associations among phenotypes, respectively. Despite the significant increase in the number of publications in the recent literature on synthetic EHR generation methods, we identify a few critical areas for future research. The benchmarking results highlight a substantial performance gap in fidelity between the existing methods and \textit{PBR}, indicating the need for future methods research to narrow this gap. Finally, comprehensive benchmarking of the existing synthetic data generation methods that account for the longitudinal nature of EHR data and support conditional data generation is needed.

\section*{\textcolor{black}{Funding Statement}}

\textcolor{black}{This work was supported through grant DMS1712933 from the National Science Foundation and MI-CARES grant 1UG3CA267907 from the National Cancer Institute. The funders had no role in the design of the study; collection, analysis, or interpretation of the data; writing of the report; or the decision to submit the manuscript for publication.}

\section*{\textcolor{black}{Competing Interests Statement}}

\textcolor{black}{The authors of this study have no competing interests to declare.}

\section*{\textcolor{black}{Contributorship Statement}}

\textcolor{black}{The contributions of authors in this paper are summarized as follows: Xingran Chen (data curation, methodology, formal analysis, visualization, open-source package, writing), Zhenke Wu (formal analysis, methodology, supervision, writing), Xu Shi (critical revisions), Hyunghoon Cho (critical revisions), Bhramar Mukherjee (Conceptualization, methodology, supervision, writing, funding acquisition). All authors approved the final version of the manuscript.}

\section*{\textcolor{black}{Data Availability statement}}

\textcolor{black}{The data used for benchmarking evaluation in this study were provided by the authors of MIMIC-III and MIMIC-IV and are available on PhysioNet upon request. The codebase used to implemented the benchmarking evaluation in this study is available at \href{https://github.com/chenxran/synthEHRella}{https://github.com/ chenxran/synthEHRella}. There is no new dataset generated and released in this study.}

\section*{Acknowledgment}

\textcolor{black}{Large language models (LLM)-assisted technologies, such as ChatGPT-4o, were used to improve the writing of this paper. The primary use of LLMs was to refine the language and grammar. All LLM suggestions were evaluated by the first author prior to their incorporation into the manuscript. LLMs were strictly limited to proposing edits to the initial drafts and were not used to generate any new materials.}

\printbibliography

@article{yuan2023ehrdiff,
  title={{EHRDiff}: Exploring Realistic {EHR} Synthesis with Diffusion Models},
  author={Yuan, Hongyi and Zhou, Songchi and Yu, Sheng},
  journal={arXiv preprint arXiv:2303.05656},
  year={2023}
}

@inproceedings{sun2024collaborative,
  title={Collaborative Synthesis of Patient Records through Multi-Visit Health State Inference},
  author={Sun, Hongda and Lin, Hongzhan and Yan, Rui},
  booktitle={Proceedings of the AAAI Conference on Artificial Intelligence},
  volume={38},
  number={17},
  pages={19044--19052},
  year={2024}
}

@article{chen2024guided,
  title={Guided Discrete Diffusion for Electronic Health Record Generation},
  author={Chen, Zixiang and Han, Jun and Li, Yongqian and Kou, Yiwen and Halperin, Eran and Tillman, Robert E and Gu, Quanquan},
  journal={arXiv preprint arXiv:2404.12314},
  year={2024}
}

@article{pang2024cehr,
  title={{CEHR-GPT}: Generating Electronic Health Records with Chronological Patient Timelines},
  author={Pang, Chao and Jiang, Xinzhuo and Pavinkurve, Nishanth Parameshwar and Kalluri, Krishna S and Minto, Elise L and Patterson, Jason and Zhang, Linying and Hripcsak, George and Elhadad, No{\'e}mie and Natarajan, Karthik},
  journal={arXiv preprint arXiv:2402.04400},
  year={2024}
}

@article{he2023meddiff,
  title={{MedDiff}: Generating electronic health records using accelerated denoising diffusion model},
  author={He, Huan and Zhao, Shifan and Xi, Yuanzhe and Ho, Joyce C},
  journal={arXiv preprint arXiv:2302.04355},
  year={2023}
}

@article{ceritli2023synthesizing,
  title={Synthesizing mixed-type electronic health records using diffusion models},
  author={Ceritli, Taha and Ghosheh, Ghadeer O and Chauhan, Vinod Kumar and Zhu, Tingting and Creagh, Andrew P and Clifton, David A},
  journal={arXiv preprint arXiv:2302.14679},
  year={2023}
}

@article{li2023generating,
  title={Generating synthetic mixed-type longitudinal electronic health records for artificial intelligent applications},
  author={Li, Jin and Cairns, Benjamin J and Li, Jingsong and Zhu, Tingting},
  journal={NPJ Digital Medicine},
  volume={6},
  number={1},
  pages={98},
  year={2023},
  publisher={Nature Publishing Group UK London}
}

@article{nikolentzos2023synthetic,
  title={Synthetic electronic health records generated with variational graph autoencoders},
  author={Nikolentzos, Giannis and Vazirgiannis, Michalis and Xypolopoulos, Christos and Lingman, Markus and Brandt, Erik G},
  journal={NPJ Digital Medicine},
  volume={6},
  number={1},
  pages={83},
  year={2023},
  publisher={Nature Publishing Group UK London}
}

@article{wang2022promptehr,
  title={{PromptEHR}: Conditional electronic healthcare records generation with prompt learning},
  author={Wang, Zifeng and Sun, Jimeng},
  journal={arXiv preprint arXiv:2211.01761},
  year={2022}
}

@inproceedings{biswal2021eva,
  title={{EVA}: Generating longitudinal electronic health records using conditional variational autoencoders},
  author={Biswal, Siddharth and Ghosh, Soumya and Duke, Jon and Malin, Bradley and Stewart, Walter and Xiao, Cao and Sun, Jimeng},
  booktitle={Machine Learning for Healthcare Conference},
  pages={260--282},
  year={2021},
  organization={PMLR}
}

@article{zhang2021synteg,
  title={{SynTEG}: a framework for temporal structured electronic health data simulation},
  author={Zhang, Ziqi and Yan, Chao and Lasko, Thomas A and Sun, Jimeng and Malin, Bradley A},
  journal={Journal of the American Medical Informatics Association},
  volume={28},
  number={3},
  pages={596--604},
  year={2021},
  publisher={Oxford University Press}
}

@article{kaur2021application,
  title={Application of Bayesian networks to generate synthetic health data},
  author={Kaur, Dhamanpreet and Sobiesk, Matthew and Patil, Shubham and Liu, Jin and Bhagat, Puran and Gupta, Amar and Markuzon, Natasha},
  journal={Journal of the American Medical Informatics Association},
  volume={28},
  number={4},
  pages={801--811},
  year={2021},
  publisher={Oxford University Press}
}

@article{torfi2020corgan,
title={{COR-GAN}: Correlation-Capturing Convolutional Neural Networks for Generating Synthetic Healthcare Records},
author={Torfi, Amirsina and Fox, Edward A},
journal={arXiv preprint arXiv:2001.09346},
year={2020}
}

@inproceedings{yan2020generating,
  title={Generating electronic health records with multiple data types and constraints},
  author={Yan, Chao and Zhang, Ziqi and Nyemba, Steve and Malin, Bradley A},
  booktitle={AMIA annual symposium proceedings},
  volume={2020},
  pages={1335},
  year={2020},
  organization={American Medical Informatics Association}
}

@article{zhang2020ensuring,
  title={Ensuring electronic medical record simulation through better training, modeling, and evaluation},
  author={Zhang, Ziqi and Yan, Chao and Mesa, Diego A and Sun, Jimeng and Malin, Bradley A},
  journal={Journal of the American Medical Informatics Association},
  volume={27},
  number={1},
  pages={99--108},
  year={2020},
  publisher={Oxford University Press}
}

@article{lee2020generating,
  title={Generating sequential electronic health records using dual adversarial autoencoder},
  author={Lee, Dongha and Yu, Hwanjo and Jiang, Xiaoqian and Rogith, Deevakar and Gudala, Meghana and Tejani, Mubeen and Zhang, Qiuchen and Xiong, Li},
  journal={Journal of the American Medical Informatics Association},
  volume={27},
  number={9},
  pages={1411--1419},
  year={2020},
  publisher={Oxford University Press}
}

@article{yale2020generation,
  title={Generation and evaluation of privacy preserving synthetic health data},
  author={Yale, Andrew and Dash, Saloni and Dutta, Ritik and Guyon, Isabelle and Pavao, Adrien and Bennett, Kristin P},
  journal={Neurocomputing},
  volume={416},
  pages={244--255},
  year={2020},
  publisher={Elsevier}
}

@article{baowaly2019synthesizing,
  title={Synthesizing electronic health records using improved generative adversarial networks},
  author={Baowaly, Mrinal Kanti and Lin, Chia-Ching and Liu, Chao-Lin and Chen, Kuan-Ta},
  journal={Journal of the American Medical Informatics Association},
  volume={26},
  number={3},
  pages={228--241},
  year={2019},
  publisher={Oxford University Press}
}

@inproceedings{choi2017generating,
  title={Generating multi-label discrete patient records using generative adversarial networks},
  author={Choi, Edward and Biswal, Siddharth and Malin, Bradley and Duke, Jon and Stewart, Walter F and Sun, Jimeng},
  booktitle={Machine Learning for Healthcare Conference},
  pages={286--305},
  year={2017},
  organization={PMLR}
}

@article{walonoski2018synthea,
  title={Synthea: An approach, method, and software mechanism for generating synthetic patients and the synthetic electronic health care record},
  author={Walonoski, Jason and Kramer, Mark and Nichols, Joseph and Quina, Andre and Moesel, Chris and Hall, Dylan and Duffett, Carlton and Dube, Kudakwashe and Gallagher, Thomas and McLachlan, Scott},
  journal={Journal of the American Medical Informatics Association},
  volume={25},
  number={3},
  pages={230--238},
  year={2018},
  publisher={Oxford University Press}
}

@article{esteban2017real,
  title={Real-valued (medical) time series generation with recurrent conditional gans},
  author={Esteban, Crist{\'o}bal and Hyland, Stephanie L and R{\"a}tsch, Gunnar},
  journal={arXiv preprint arXiv:1706.02633},
  year={2017}
}

@article{franklin2014plasmode,
  title={Plasmode simulation for the evaluation of pharmacoepidemiologic methods in complex healthcare databases},
  author={Franklin, Jessica M and Schneeweiss, Sebastian and Polinski, Jennifer M and Rassen, Jeremy A},
  journal={Computational Statistics \& Data Analysis},
  volume={72},
  pages={219--226},
  year={2014},
  publisher={Elsevier}
}

@inproceedings{park2013perturbed,
  title={Perturbed gibbs samplers for generating large-scale privacy-safe synthetic health data},
  author={Park, Yubin and Ghosh, Joydeep and Shankar, Mallikarjun},
  booktitle={2013 IEEE International Conference on Healthcare Informatics},
  pages={493--498},
  year={2013},
  organization={IEEE}
}

@article{buczak2010data,
  title={Data-driven approach for creating synthetic electronic medical records},
  author={Buczak, Anna L and Babin, Steven and Moniz, Linda},
  journal={BMC Medical Informatics and Decision Making},
  volume={10},
  pages={1--28},
  year={2010},
  publisher={Springer}
}

@article{10.1093/bioinformatics/btad655,
    author = {Shuey, Megan M and Stead, William W and Aka, Ida and Barnado, April L and Bastarache, Julie A and Brokamp, Elly and Campbell, Meredith and Carroll, Robert J and Goldstein, Jeffrey A and Lewis, Adam and Malow, Beth A and Mosley, Jonathan D and Osterman, Travis and Padovani-Claudio, Dolly A and Ramirez, Andrea and Roden, Dan M and Schuler, Bryce A and Siew, Edward and Sucre, Jennifer and Thomsen, Isaac and Tinker, Rory J and Van Driest, Sara and Walsh, Colin and Warner, Jeremy L and Wells, Quinn S and Wheless, Lee and Bastarache, Lisa},
    title = "{Next-generation phenotyping: introducing phecodeX for enhanced discovery research in medical phenomics}",
    journal = {Bioinformatics},
    volume = {39},
    number = {11},
    pages = {btad655},
    year = {2023},
    month = {11},
    issn = {1367-4811},
    doi = {10.1093/bioinformatics/btad655},
    url = {https://doi.org/10.1093/bioinformatics/btad655},
    eprint = {https://academic.oup.com/bioinformatics/article-pdf/39/11/btad655/52829330/btad655.pdf},
}

@article{johnson2016mimic,
  title={{MIMIC-III}, a freely accessible critical care database},
  author={Johnson, Alistair EW and Pollard, Tom J and Shen, Lu and Lehman, Li-wei H and Feng, Mengling and Ghassemi, Mohammad and Moody, Benjamin and Szolovits, Peter and Anthony Celi, Leo and Mark, Roger G},
  journal={Scientific Data},
  volume={3},
  number={1},
  pages={1--9},
  year={2016},
  publisher={Nature Publishing Group}
}

@InProceedings{sohl2015deep,
  title = 	 {Deep Unsupervised Learning using Nonequilibrium Thermodynamics},
  author = 	 {Sohl-Dickstein, Jascha and Weiss, Eric and Maheswaranathan, Niru and Ganguli, Surya},
  booktitle = 	 {Proceedings of the 32nd International Conference on Machine Learning},
  pages = 	 {2256--2265},
  year = 	 {2015},
  editor = 	 {Bach, Francis and Blei, David},
  volume = 	 {37},
  series = 	 {Proceedings of Machine Learning Research},
  address = 	 {Lille, France},
  month = 	 {07--09 Jul},
  publisher =    {PMLR},
  url = 	 {https://proceedings.mlr.press/v37/sohl-dickstein15.html}
}

@inproceedings{ho2020denoising,
 author = {Ho, Jonathan and Jain, Ajay and Abbeel, Pieter},
 booktitle = {Advances in Neural Information Processing Systems},
 editor = {H. Larochelle and M. Ranzato and R. Hadsell and M.F. Balcan and H. Lin},
 pages = {6840--6851},
 publisher = {Curran Associates, Inc.},
 title = {Denoising Diffusion Probabilistic Models},
 url = {https://proceedings.neurips.cc/paper_files/paper/2020/file/4c5bcfec8584af0d967f1ab10179ca4b-Paper.pdf},
 volume = {33},
 year = {2020}
}

@inproceedings{vaswani2017attention,
 author = {Vaswani, Ashish and Shazeer, Noam and Parmar, Niki and Uszkoreit, Jakob and Jones, Llion and Gomez, Aidan N and Kaiser, \L ukasz and Polosukhin, Illia},
 booktitle = {Advances in Neural Information Processing Systems},
 editor = {I. Guyon and U. Von Luxburg and S. Bengio and H. Wallach and R. Fergus and S. Vishwanathan and R. Garnett},
 pages = {},
 publisher = {Curran Associates, Inc.},
 title = {Attention is All you Need},
 url = {https://proceedings.neurips.cc/paper_files/paper/2017/file/3f5ee243547dee91fbd053c1c4a845aa-Paper.pdf},
 volume = {30},
 year = {2017}
}

@inproceedings{dube2014approach,
  title={Approach and method for generating realistic synthetic electronic healthcare records for secondary use},
  author={Dube, Kudakwashe and Gallagher, Thomas},
  booktitle={Foundations of Health Information Engineering and Systems: Third International Symposium, FHIES 2013, Macau, China, August 21-23, 2013. Revised Selected Papers 3},
  pages={69--86},
  year={2014},
  organization={Springer}
}

@inproceedings{mclachlan2016using,
  title={Using the caremap with health incidents statistics for generating the realistic synthetic electronic healthcare record},
  author={McLachlan, Scott and Dube, Kudakwashe and Gallagher, Thomas},
  booktitle={2016 IEEE International Conference on Healthcare Informatics (ICHI)},
  pages={439--448},
  year={2016},
  organization={IEEE}
}

@article{mclachlan2018aten,
  title={The {ATEN} framework for creating the realistic synthetic electronic health record},
  author={McLachlan, Scott and Dube, Kudakwashe and Gallagher, Thomas and Daley, Bridget and Walonoski, Jason and others},
  year={2018}
}

@inproceedings{yang2019grouped,
  title={Grouped correlational generative adversarial networks for discrete electronic health records},
  author={Yang, Fan and Yu, Zhongping and Liang, Yunfan and Gan, Xiaolu and Lin, Kaibiao and Zou, Quan and Zeng, Yifeng},
  booktitle={2019 IEEE International Conference on Bioinformatics and Biomedicine (BIBM)},
  pages={906--913},
  year={2019},
  organization={IEEE}
}

@article{yoon2020anonymization,
  title={Anonymization through data synthesis using generative adversarial networks ({ADS-GAN})},
  author={Yoon, Jinsung and Drumright, Lydia N and Van Der Schaar, Mihaela},
  journal={IEEE Journal of Biomedical and Health Informatics},
  volume={24},
  number={8},
  pages={2378--2388},
  year={2020},
  publisher={IEEE}
}

@inproceedings{rashidian2020smooth,
  title={{SMOOTH-GAN}: towards sharp and smooth synthetic {EHR} data generation},
  author={Rashidian, Sina and Wang, Fusheng and Moffitt, Richard and Garcia, Victor and Dutt, Anurag and Chang, Wei and Pandya, Vishwam and Hajagos, Janos and Saltz, Mary and Saltz, Joel},
  booktitle={Artificial Intelligence in Medicine: 18th International Conference on Artificial Intelligence in Medicine, AIME 2020, Minneapolis, MN, USA, August 25--28, 2020, Proceedings 18},
  pages={37--48},
  year={2020},
  organization={Springer}
}

@inproceedings{sun2021generating,
  title={Generating longitudinal synthetic {EHR} data with recurrent autoencoders and generative adversarial networks},
  author={Sun, Siao and Wang, Fusheng and Rashidian, Sina and Kurc, Tahsin and Abell-Hart, Kayley and Hajagos, Janos and Zhu, Wei and Saltz, Mary and Saltz, Joel},
  booktitle={Heterogeneous Data Management, Polystores, and Analytics for Healthcare: VLDB Workshops, Poly 2021 and DMAH 2021, Virtual Event, August 20, 2021, Revised Selected Papers 7},
  pages={153--165},
  year={2021},
  organization={Springer}
}

@article{mosquera2023method,
  title={A method for generating synthetic longitudinal health data},
  author={Mosquera, Lucy and El Emam, Khaled and Ding, Lei and Sharma, Vishal and Zhang, Xue Hua and Kababji, Samer El and Carvalho, Chris and Hamilton, Brian and Palfrey, Dan and Kong, Linglong and others},
  journal={BMC Medical Research Methodology},
  volume={23},
  number={1},
  pages={67},
  year={2023},
  publisher={Springer}
}

@inproceedings{naseer2023scoehr,
  title={{ScoEHR}: Generating Synthetic Electronic Health Records using Continuous-time Diffusion Models},
  author={Naseer, Ahmed Ammar and Walker, Benjamin and Landon, Christopher and Ambrosy, Andrew and Fudim, Marat and Wysham, Nicholas and Toro, Botros and Swaminathan, Sumanth and Lyons, Terry},
  booktitle={Machine Learning for Healthcare Conference},
  pages={489--508},
  year={2023},
  organization={PMLR}
}

@article{theodorou2023synthesize,
  title={Synthesize high-dimensional longitudinal electronic health records via hierarchical autoregressive language model},
  author={Theodorou, Brandon and Xiao, Cao and Sun, Jimeng},
  journal={Nature Communications},
  volume={14},
  number={1},
  pages={5305},
  year={2023},
  publisher={Nature Publishing Group UK London}
}

@inproceedings{heflexible,
  title={A Flexible Generative Model for Heterogeneous Tabular {EHR} with Missing Modality},
  author={He, Huan and Xi, Yuanzhe and Chen, Yong and Malin, Bradley and Ho, Joyce and others},
  booktitle={The Twelfth International Conference on Learning Representations},
  year={2024}
}

@article{ramachandranpillai2024bt,
  title={{Bt-GAN}: Generating Fair Synthetic Healthdata via Bias-transforming Generative Adversarial Networks},
  author={Ramachandranpillai, Resmi and Sikder, Md Fahim and Bergstr{\"o}m, David and Heintz, Fredrik},
  journal={Journal of Artificial Intelligence Research},
  volume={79},
  pages={1313--1341},
  year={2024}
}

@inproceedings{wang2024igamt,
  title={{IGAMT}: Privacy-Preserving Electronic Health Record Synthesization with Heterogeneity and Irregularity},
  author={Wang, Wenjie and Tang, Pengfei and Lou, Jian and Shao, Yuanming and Waller, Lance and Ko, Yi-an and Xiong, Li},
  booktitle={Proceedings of the AAAI Conference on Artificial Intelligence},
  volume={38},
  number={14},
  pages={15634--15643},
  year={2024}
}

@inproceedings{vardhan2024large,
  title={Large language models as synthetic electronic health record data generators},
  author={Vardhan, Madhurima and Nathani, Deepak and Vardhan, Swarnima and Aggarwal, Abhinav and Simini, Filippo},
  booktitle={2024 IEEE Conference on Artificial Intelligence (CAI)},
  pages={804--810},
  year={2024},
  organization={IEEE}
}

@article{gwon2024ldp,
  title={{LDP-GAN}: Generative adversarial networks with local differential privacy for patient medical records synthesis},
  author={Gwon, Hansle and Ahn, Imjin and Kim, Yunha and Kang, Hee Jun and Seo, Hyeram and Choi, Heejung and Cho, Ha Na and Kim, Minkyoung and Han, JiYe and Kee, Gaeun and others},
  journal={Computers in Biology and Medicine},
  volume={168},
  pages={107738},
  year={2024},
  publisher={Elsevier}
}

@article{tian2023fast,
  title={Fast and reliable generation of {EHR} time series via diffusion models},
  author={Tian, Muhang and Chen, Bernie and Guo, Allan and Jiang, Shiyi and Zhang, Anru R},
  journal={arXiv preprint arXiv:2310.15290},
  year={2023}
}

@inproceedings{zhong2024synthesizing,
  title={Synthesizing Multimodal Electronic Health Records via Predictive Diffusion Models},
  author={Zhong, Yuan and Wang, Xiaochen and Wang, Jiaqi and Zhang, Xiaokun and Wang, Yaqing and Huai, Mengdi and Xiao, Cao and Ma, Fenglong},
  booktitle={Proceedings of the 30th ACM SIGKDD Conference on Knowledge Discovery and Data Mining},
  pages={4607--4618},
  year={2024}
}

@article{yan2022multifaceted,
  title={A multifaceted benchmarking of synthetic electronic health record generation models},
  author={Yan, Chao and Yan, Yao and Wan, Zhiyu and Zhang, Ziqi and Omberg, Larsson and Guinney, Justin and Mooney, Sean D and Malin, Bradley A},
  journal={Nature Communications},
  volume={13},
  number={1},
  pages={7609},
  year={2022},
  publisher={Nature Publishing Group UK London}
}

@article{tarczy2013survey,
  title={A survey of informatics approaches to whole-exome and whole-genome clinical reporting in the electronic health record},
  author={Tarczy-Hornoch, Peter and Amendola, Laura and Aronson, Samuel J and Garraway, Levi and Gray, Stacy and Grundmeier, Robert W and Hindorff, Lucia A and Jarvik, Gail and Karavite, Dean and Lebo, Matthew and others},
  journal={Genetics in Medicine},
  volume={15},
  number={10},
  pages={824--832},
  year={2013},
  publisher={Nature Publishing Group}
}

@article{linder2021role,
  title={The role of electronic health records in advancing genomic medicine},
  author={Linder, Jodell E and Bastarache, Lisa and Hughey, Jacob J and Peterson, Josh F},
  journal={Annual Review of Genomics and Human Genetics},
  volume={22},
  number={1},
  pages={219--238},
  year={2021},
  publisher={Annual Reviews}
}

@article{yadav2018mining,
  title={Mining electronic health records ({EHR}s) A survey},
  author={Yadav, Pranjul and Steinbach, Michael and Kumar, Vipin and Simon, Gyorgy},
  journal={ACM Computing Surveys (CSUR)},
  volume={50},
  number={6},
  pages={1--40},
  year={2018},
  publisher={ACM New York, NY, USA}
}

@article{beesley2020emerging,
  title={The emerging landscape of health research based on biobanks linked to electronic health records: Existing resources, statistical challenges, and potential opportunities},
  author={Beesley, Lauren J and Salvatore, Maxwell and Fritsche, Lars G and Pandit, Anita and Rao, Arvind and Brummett, Chad and Willer, Cristen J and Lisabeth, Lynda D and Mukherjee, Bhramar},
  journal={Statistics in Medicine},
  volume={39},
  number={6},
  pages={773--800},
  year={2020},
  publisher={Wiley Online Library}
}

@article{sudlow2015uk,
  title={{UK biobank}: an open access resource for identifying the causes of a wide range of complex diseases of middle and old age},
  author={Sudlow, Cathie and Gallacher, John and Allen, Naomi and Beral, Valerie and Burton, Paul and Danesh, John and Downey, Paul and Elliott, Paul and Green, Jane and Landray, Martin and others},
  journal={PLOS Medicine},
  volume={12},
  number={3},
  pages={e1001779},
  year={2015},
  publisher={Public Library of Science San Francisco, CA USA}
}

@article{all2019all,
  title={{The “All of Us” research program}},
  author={All of Us Research Program Investigators},
  journal={New England Journal of Medicine},
  volume={381},
  number={7},
  pages={668--676},
  year={2019},
  publisher={Mass Medical Soc}
}

@article{zhou2022global,
  title={Global Biobank Meta-analysis Initiative: Powering genetic discovery across human disease},
  author={Zhou, Wei and Kanai, Masahiro and Wu, Kuan-Han H and Rasheed, Humaira and Tsuo, Kristin and Hirbo, Jibril B and Wang, Ying and Bhattacharya, Arjun and Zhao, Huiling and Namba, Shinichi and others},
  journal={Cell Genomics},
  volume={2},
  number={10},
  year={2022},
  publisher={Elsevier}
}

@article{murcia2024automating,
  title={Automating Clinical Trial Matches Via Natural Language Processing of Synthetic Electronic Health Records and Clinical Trial Eligibility Criteria},
  author={Murcia, Victor M and Aggarwal, Vinod and Pesaladinne, Nikhil and Thammineni, Ram and Do, Nhan and Alterovitz, Gil and Fricks, Rafael B},
  journal={AMIA Summits on Translational Science Proceedings},
  volume={2024},
  pages={125},
  year={2024},
  publisher={American Medical Informatics Association}
}

@article{rankin2020reliability,
  title={Reliability of supervised machine learning using synthetic data in health care: Model to preserve privacy for data sharing},
  author={Rankin, Debbie and Black, Michaela and Bond, Raymond and Wallace, Jonathan and Mulvenna, Maurice and Epelde, Gorka and others},
  journal={JMIR Medical Informatics},
  volume={8},
  number={7},
  pages={e18910},
  year={2020},
  publisher={JMIR Publications Inc., Toronto, Canada}
}

@article{budu2024evaluation,
  title={Evaluation of synthetic electronic health records: A systematic review and experimental assessment},
  author={Budu, Emmanuella and Etminani, Kobra and Soliman, Amira and R{\"o}gnvaldsson, Thorsteinn},
  journal={Neurocomputing},
  pages={128253},
  year={2024},
  publisher={Elsevier}
}

@inproceedings{tall2020generating,
  title={Generating Connected Synthetic Electronic Health Records and Social Media Data for Modeling and Simulation},
  author={Tall, Anne M and Zou, Cliff C and Wang, Jun},
  booktitle={Interservice/Industry Training, Simulation and Education Conference (I/ITSEC)},
  year={2020}
}

@article{shi2022generating,
  title={Generating high-fidelity privacy-conscious synthetic patient data for causal effect estimation with multiple treatments},
  author={Shi, Jingpu and Wang, Dong and Tesei, Gino and Norgeot, Beau},
  journal={Frontiers in Artificial Intelligence},
  volume={5},
  pages={918813},
  year={2022},
  publisher={Frontiers Media SA}
}

@article{chen2022simulation,
  title={Simulation of a machine learning enabled learning health system for risk prediction using synthetic patient data},
  author={Chen, Anjun and Chen, Drake O},
  journal={Scientific Reports},
  volume={12},
  number={1},
  pages={17917},
  year={2022},                                                    
  publisher={Nature Publishing Group UK London}
}

@article{perets2023subpopulation,
  title={Subpopulation-Specific Synthetic {EHR} for Better Mortality Prediction},
  author={Perets, Oriel and Rappoport, Nadav},
  journal={arXiv preprint arXiv:2305.16363},
  year={2023}
}

@article{muller2022synthesising,
  title={Synthesising Electronic Health Records: Cystic Fibrosis Patient Group},
  author={Muller, Emily and Zheng, Xu and Hayes, Jer},
  journal={arXiv preprint arXiv:2201.05400},
  year={2022}
}

@article{goncalves2020generation,
  title={Generation and evaluation of synthetic patient data},
  author={Goncalves, Andre and Ray, Priyadip and Soper, Braden and Stevens, Jennifer and Coyle, Linda and Sales, Ana Paula},
  journal={BMC Medical Research Methodology},
  volume={20},
  pages={1--40},
  year={2020},
  publisher={Springer}
}

@article{mendelevitch2021fidelity,
  title={Fidelity and privacy of synthetic medical data},
  author={Mendelevitch, Ofer and Lesh, Michael D},
  journal={arXiv preprint arXiv:2101.08658},
  year={2021}
}

@article{ghosheh2022review,
  title={A review of Generative Adversarial Networks for Electronic Health Records: applications, evaluation measures and data sources},
  author={Ghosheh, Ghadeer and Li, Jin and Zhu, Tingting},
  journal={arXiv preprint arXiv:2203.07018},
  year={2022}
}

@article{hernandez2022synthetic,
  title={Synthetic data generation for tabular health records: A systematic review},
  author={Hernandez, Mikel and Epelde, Gorka and Alberdi, Ane and Cilla, Rodrigo and Rankin, Debbie},
  journal={Neurocomputing},
  volume={493},
  pages={28--45},
  year={2022},
  publisher={Elsevier}
}

@article{achterberg2024evaluation,
  title={On the evaluation of synthetic longitudinal electronic health records},
  author={Achterberg, Jim L and Haas, Marcel R and Spruit, Marco R},
  journal={BMC Medical Research Methodology},
  volume={24},
  number={1},
  pages={181},
  year={2024},
  publisher={Springer}
}

@article{donnelly2006snomed,
  title={{SNOMED-CT}: The advanced terminology and coding system for eHealth},
  author={Donnelly, Kevin and others},
  journal={Studies in Health Technology and Informatics},
  volume={121},
  pages={279},
  year={2006},
  publisher={IOS Press; 1999}
}

@article{birkhead2015uses,
  title={Uses of electronic health records for public health surveillance to advance public health},
  author={Birkhead, Guthrie S and Klompas, Michael and Shah, Nirav R},
  journal={Annual Review of Public Health},
  volume={36},
  number={1},
  pages={345--359},
  year={2015},
  publisher={Annual Reviews}
}

@article{friedman2013electronic,
  title={Electronic health records and US public health: current realities and future promise},
  author={Friedman, Daniel J and Parrish, R Gibson and Ross, David A},
  journal={American Journal of Public Health},
  volume={103},
  number={9},
  pages={1560--1567},
  year={2013},
  publisher={American Public Health Association}
}

@article{kruse2018use,
  title={The use of Electronic Health Records to Support Population Health: A Systematic Review of the Literature},
  author={Clemens Scott Kruse and Anna Stein and Heather Thomas and Harman D Kaur},
  journal={Journal of Medical Systems},
  year={2018},
  volume={42},
  url={https://api.semanticscholar.org/CorpusID:52889203}
}

@article{fodeh2016mining,
  title={Mining Big Data in biomedicine and health care},
  author={Samah Jamal Fodeh and Qing Zeng-Treitler},
  journal={Journal of Biomedical Informatics},
  year={2016},
  volume={63},
  pages={
          400-403
        },
  url={https://api.semanticscholar.org/CorpusID:205714982}
}

@article{johnson2023mimic,
  title={{MIMIC-IV}, a freely accessible electronic health record dataset},
  author={Johnson, Alistair EW and Bulgarelli, Lucas and Shen, Lu and Gayles, Alvin and Shammout, Ayad and Horng, Steven and Pollard, Tom J and Hao, Sicheng and Moody, Benjamin and Gow, Brian and others},
  journal={Scientific Data},
  volume={10},
  number={1},
  pages={1},
  year={2023},
  publisher={Nature Publishing Group UK London}
}

@inproceedings{li2019neural,
author = {Li, Naihan and Liu, Shujie and Liu, Yanqing and Zhao, Sheng and Liu, Ming},
title = {Neural speech synthesis with transformer network},
year = {2019},
isbn = {978-1-57735-809-1},
publisher = {AAAI Press},
url = {https://doi.org/10.1609/aaai.v33i01.33016706},
doi = {10.1609/aaai.v33i01.33016706},
booktitle = {Proceedings of the Thirty-Third AAAI Conference on Artificial Intelligence and Thirty-First Innovative Applications of Artificial Intelligence Conference and Ninth AAAI Symposium on Educational Advances in Artificial Intelligence},
articleno = {823},
numpages = {8},
location = {Honolulu, Hawaii, USA}
}

@article{devlin2018bert,
  title={Bert: Pre-training of deep bidirectional transformers for language understanding},
  author={Devlin, Jacob},
  journal={arXiv preprint arXiv:1810.04805},
  year={2018}
}

@inproceedings{brown2020gpt3,
author = {Brown, Tom B. and Mann, Benjamin and Ryder, Nick and Subbiah, Melanie and Kaplan, Jared and Dhariwal, Prafulla and Neelakantan, Arvind and Shyam, Pranav and Sastry, Girish and Askell, Amanda and Agarwal, Sandhini and Herbert-Voss, Ariel and Krueger, Gretchen and Henighan, Tom and Child, Rewon and Ramesh, Aditya and Ziegler, Daniel M. and Wu, Jeffrey and Winter, Clemens and Hesse, Christopher and Chen, Mark and Sigler, Eric and Litwin, Mateusz and Gray, Scott and Chess, Benjamin and Clark, Jack and Berner, Christopher and McCandlish, Sam and Radford, Alec and Sutskever, Ilya and Amodei, Dario},
title = {Language models are few-shot learners},
year = {2020},
isbn = {9781713829546},
publisher = {Curran Associates Inc.},
address = {Red Hook, NY, USA},
booktitle = {Proceedings of the 34th International Conference on Neural Information Processing Systems},
articleno = {159},
numpages = {25},
location = {Vancouver, BC, Canada},
series = {NIPS '20}
}

@article{ho2022imagen,
  title={Imagen video: High definition video generation with diffusion models},
  author={Ho, Jonathan and Chan, William and Saharia, Chitwan and Whang, Jay and Gao, Ruiqi and Gritsenko, Alexey and Kingma, Diederik P and Poole, Ben and Norouzi, Mohammad and Fleet, David J and others},
  journal={arXiv preprint arXiv:2210.02303},
  year={2022}
}

@article{ho2022video,
  title={Video diffusion models},
  author={Ho, Jonathan and Salimans, Tim and Gritsenko, Alexey and Chan, William and Norouzi, Mohammad and Fleet, David J},
  journal={Advances in Neural Information Processing Systems},
  volume={35},
  pages={8633--8646},
  year={2022}
}

@inproceedings{saharia2022photorealistic,
author = {Saharia, Chitwan and Chan, William and Saxena, Saurabh and Lit, Lala and Whang, Jay and Denton, Emily and Ghasemipour, Seyed Kamyar Seyed and Ayan, Burcu Karagol and Mahdavi, S. Sara and Gontijo-Lopes, Raphael and Salimans, Tim and Ho, Jonathan and Fleet, David J and Norouzi, Mohammad},
title = {Photorealistic text-to-image diffusion models with deep language understanding},
year = {2024},
isbn = {9781713871088},
publisher = {Curran Associates Inc.},
address = {Red Hook, NY, USA},
booktitle = {Proceedings of the 36th International Conference on Neural Information Processing Systems},
articleno = {2643},
numpages = {16},
location = {New Orleans, LA, USA},
series = {NIPS '22}
}

@article{laderas2017teaching,
  title={Teaching data science fundamentals through realistic synthetic clinical cardiovascular data},
  author={Laderas, Ted and Vasilevsky, Nicole and Pederson, Bjorn and Haendel, Melissa and McWeeney, Shannon and Dorr, David A},
  journal={BioRxiv},
  pages={232611},
  year={2017},
  publisher={Cold Spring Harbor Laboratory}
}

@article{bing2022conditional,
  title={Conditional generation of medical time series for extrapolation to underrepresented populations},
  author={Bing, Simon and Dittadi, Andrea and Bauer, Stefan and Schwab, Patrick},
  journal={PLOS Digital Health},
  volume={1},
  number={7},
  pages={e0000074},
  year={2022},
  publisher={Public Library of Science San Francisco, CA USA}
}

@article{yoon2023ehr,
  title={{EHR-Safe}: generating high-fidelity and privacy-preserving synthetic electronic health records},
  author={Yoon, Jinsung and Mizrahi, Michel and Ghalaty, Nahid Farhady and Jarvinen, Thomas and Ravi, Ashwin S and Brune, Peter and Kong, Fanyu and Anderson, Dave and Lee, George and Meir, Arie and others},
  journal={NPJ Digital Medicine},
  volume={6},
  number={1},
  pages={141},
  year={2023},
  publisher={Nature Publishing Group UK London}
}

@inproceedings{zheng2023toward,
 author = {Zheng, Chenyu and Wu, Guoqiang and LI, Chongxuan},
 booktitle = {Advances in Neural Information Processing Systems},
 editor = {A. Oh and T. Naumann and A. Globerson and K. Saenko and M. Hardt and S. Levine},
 pages = {54046--54060},
 publisher = {Curran Associates, Inc.},
 title = {Toward Understanding Generative Data Augmentation},
 url = {https://proceedings.neurips.cc/paper_files/paper/2023/file/a94a8800a4b0af45600bab91164849df-Paper-Conference.pdf},
 volume = {36},
 year = {2023}
}

@article{tsiklidis2022predicting,
  title={Predicting risk for trauma patients using static and dynamic information from the {MIMIC III} database},
  author={Tsiklidis, Evan J and Sinno, Talid and Diamond, Scott L},
  journal={PLOS ONE},
  volume={17},
  number={1},
  pages={e0262523},
  year={2022},
  publisher={Public Library of Science San Francisco, CA USA}
}

@article{huang2021nomogram,
  title={A nomogram to predict in-hospital mortality of neonates admitted to the intensive care unit},
  author={Huang, Xihua and Liang, Zhenyu and Li, Tang and Lingna, Yu and Zhu, Wei and Li, Huiyi},
  journal={International Health},
  volume={13},
  number={6},
  pages={633--639},
  year={2021},
  publisher={Oxford University Press}
}

@article{shi2022evaluation,
  title={Evaluation of the neonatal sequential organ failure assessment and mortality risk in neonates with respiratory distress syndrome: A retrospective cohort study},
  author={Shi, Shanshan and Guo, Jie and Fu, Minqiang and Liao, Lihua and Tu, Jiabin and Xiong, Jialing and Liao, Quanwang and Chen, Weihua and Chen, Kaihong and Liao, Ying},
  journal={Frontiers in Pediatrics},
  volume={10},
  pages={911444},
  year={2022},
  publisher={Frontiers Media SA}
}

@article{hou2020predicting,
  title={Predicting 30-days mortality for {MIMIC-III} patients with sepsis-3: a machine learning approach using XGboost},
  author={Hou, Nianzong and Li, Mingzhe and He, Lu and Xie, Bing and Wang, Lin and Zhang, Rumin and Yu, Yong and Sun, Xiaodong and Pan, Zhengsheng and Wang, Kai},
  journal={Journal of Translational Medicine},
  volume={18},
  pages={1--14},
  year={2020},
  publisher={Springer}
}

@article{dai2020analysis,
  title={Analysis of adult disease characteristics and mortality on {MIMIC-III}},
  author={Dai, Zheng and Liu, Siru and Wu, Jinfa and Li, Mengdie and Liu, Jialin and Li, Ke},
  journal={PLOS ONE},
  volume={15},
  number={4},
  pages={e0232176},
  year={2020},
  publisher={Public Library of Science San Francisco, CA USA}
}

@inproceedings{chen2016xgboost,
author = {Chen, Tianqi and Guestrin, Carlos},
title = {{XGBoost}: A Scalable Tree Boosting System},
year = {2016},
isbn = {9781450342322},
publisher = {Association for Computing Machinery},
address = {New York, NY, USA},
url = {https://doi.org/10.1145/2939672.2939785},
doi = {10.1145/2939672.2939785},
booktitle = {Proceedings of the 22nd ACM SIGKDD International Conference on Knowledge Discovery and Data Mining},
pages = {785–794},
numpages = {10},
location = {San Francisco, California, USA},
series = {KDD '16}
}

@inproceedings{ke2017lightgbm,
 author = {Ke, Guolin and Meng, Qi and Finley, Thomas and Wang, Taifeng and Chen, Wei and Ma, Weidong and Ye, Qiwei and Liu, Tie-Yan},
 booktitle = {Advances in Neural Information Processing Systems},
 editor = {I. Guyon and U. Von Luxburg and S. Bengio and H. Wallach and R. Fergus and S. Vishwanathan and R. Garnett},
 pages = {},
 publisher = {Curran Associates, Inc.},
 title = {{LightGBM}: A Highly Efficient Gradient Boosting Decision Tree},
 url = {https://proceedings.neurips.cc/paper_files/paper/2017/file/6449f44a102fde848669bdd9eb6b76fa-Paper.pdf},
 volume = {30},
 year = {2017}
}

@inproceedings{ho1995random,
  title={Random decision forests},
  author={Ho, Tin Kam},
  booktitle={Proceedings of 3rd International Conference on Document Analysis and Recognition},
  volume={1},
  pages={278--282},
  year={1995},
  organization={IEEE}
}

@article{friedman2001greedy,
  title={Greedy function approximation: a gradient boosting machine},
  author={Friedman, Jerome H},
  journal={Annals of Statistics},
  pages={1189--1232},
  year={2001},
  publisher={JSTOR}
}

@InProceedings{pmlr-v206-breugel23a,
  title = 	 {Membership Inference Attacks against Synthetic Data through Overfitting Detection},
  author =       {van Breugel, Boris and Sun, Hao and Qian, Zhaozhi and van der Schaar, Mihaela},
  booktitle = 	 {Proceedings of The 26th International Conference on Artificial Intelligence and Statistics},
  pages = 	 {3493--3514},
  year = 	 {2023},
  editor = 	 {Ruiz, Francisco and Dy, Jennifer and van de Meent, Jan-Willem},
  volume = 	 {206},
  series = 	 {Proceedings of Machine Learning Research},
  month = 	 {25--27 Apr},
  publisher =    {PMLR},
  url = 	 {https://proceedings.mlr.press/v206/breugel23a.html}
}

@inproceedings{ganju2018property,
  title={Property inference attacks on fully connected neural networks using permutation invariant representations},
  author={Ganju, Karan and Wang, Qi and Yang, Wei and Gunter, Carl A and Borisov, Nikita},
  booktitle={Proceedings of the 2018 ACM SIGSAC conference on computer and communications security},
  pages={619--633},
  year={2018}
}

@inproceedings{shokri2017membership,
  title={Membership inference attacks against machine learning models},
  author={Shokri, Reza and Stronati, Marco and Song, Congzheng and Shmatikov, Vitaly},
  booktitle={2017 IEEE Symposium on Security and Privacy (SP)},
  pages={3--18},
  year={2017},
  organization={IEEE}
}

@article{zhang2022keeping,
  title={Keeping synthetic patients on track: feedback mechanisms to mitigate performance drift in longitudinal health data simulation},
  author={Zhang, Ziqi and Yan, Chao and Malin, Bradley A},
  journal={Journal of the American Medical Informatics Association},
  volume={29},
  number={11},
  pages={1890--1898},
  year={2022},
  publisher={Oxford University Press}
}

@article{xie2018differentially,
  title={Differentially private generative adversarial network},
  author={Xie, Liyang and Lin, Kaixiang and Wang, Shu and Wang, Fei and Zhou, Jiayu},
  journal={arXiv preprint arXiv:1802.06739},
  year={2018}
}

@article{tricco2018prisma,
  title={PRISMA extension for scoping reviews (PRISMA-ScR): checklist and explanation},
  author={Tricco, Andrea C and Lillie, Erin and Zarin, Wasifa and O'Brien, Kelly K and Colquhoun, Heather and Levac, Danielle and Moher, David and Peters, Micah DJ and Horsley, Tanya and Weeks, Laura and others},
  journal={Annals of Internal Medicine},
  volume={169},
  number={7},
  pages={467--473},
  year={2018},
  publisher={American College of Physicians}
}

@article{bastarache2021using,
  title={Using phecodes for research with the electronic health record: from PheWAS to PheRS},
  author={Bastarache, Lisa},
  journal={Annual review of biomedical data science},
  volume={4},
  number={1},
  pages={1--19},
  year={2021},
  publisher={Annual Reviews}
}

@article{alessandrini2010new,
  title={A new diagnosis grouping system for child emergency department visits},
  author={Alessandrini, Evaline A and Alpern, Elizabeth R and Chamberlain, James M and Shea, Judy A and Gorelick, Marc H},
  journal={Academic Emergency Medicine},
  volume={17},
  number={2},
  pages={204--213},
  year={2010},
  publisher={Wiley Online Library}
}

@article{rassekh2010reclassification,
  title={Reclassification of ICD-9 Codes into Meaningful Categories for Oncology Survivorship Research},
  author={Rassekh, SR and Lorenzi, M and Lee, L and Devji, S and McBride, M and Goddard, K},
  journal={Journal of cancer epidemiology},
  volume={2010},
  number={1},
  pages={569517},
  year={2010},
  publisher={Wiley Online Library}
}

@article{denny2010phewas,
  title={PheWAS: demonstrating the feasibility of a phenome-wide scan to discover gene--disease associations},
  author={Denny, Joshua C and Ritchie, Marylyn D and Basford, Melissa A and Pulley, Jill M and Bastarache, Lisa and Brown-Gentry, Kristin and Wang, Deede and Masys, Dan R and Roden, Dan M and Crawford, Dana C},
  journal={Bioinformatics},
  volume={26},
  number={9},
  pages={1205--1210},
  year={2010},
  publisher={Oxford University Press}
}

@article{turer2015icd,
  title={ICD-10-CM Crosswalks in the primary care setting: assessing reliability of the GEMs and reimbursement mappings},
  author={Turer, Robert W and Zuckowsky, Theresa D and Causey, H Jennifer and Rosenbloom, S Trent},
  journal={Journal of the American Medical Informatics Association},
  volume={22},
  number={2},
  pages={417--425},
  year={2015},
  publisher={Oxford University Press}
}

@article{shankar2023clinical,
title = {Clinical-GAN: Trajectory Forecasting of Clinical Events using Transformer and Generative Adversarial Networks},
journal = {Artificial Intelligence in Medicine},
volume = {138},
pages = {102507},
year = {2023},
issn = {0933-3657},
doi = {https://doi.org/10.1016/j.artmed.2023.102507},
url = {https://www.sciencedirect.com/science/article/pii/S0933365723000210},
author = {Vignesh Shankar and Elnaz Yousefi and Alireza Manashty and Dayne Blair and Deepika Teegapuram}
}

@article{SUN2023104404,
title = {Generating synthetic personal health data using conditional generative adversarial networks combining with differential privacy},
journal = {Journal of Biomedical Informatics},
volume = {143},
pages = {104404},
year = {2023},
issn = {1532-0464},
doi = {https://doi.org/10.1016/j.jbi.2023.104404},
url = {https://www.sciencedirect.com/science/article/pii/S1532046423001259},
author = {Chang Sun and Johan {van Soest} and Michel Dumontier}
}

@article{lee2024leveraging,
  title={Leveraging VQ-VAE tokenization for autoregressive modeling of medical time series},
  author={Lee, Yoonhyung and Chae, Younhyung and Jung, Kyomin},
  journal={Artificial Intelligence in Medicine},
  volume={154},
  pages={102925},
  year={2024},
  publisher={Elsevier}
}

@article{lu2023multi,
  title={Multi-label clinical time-series generation via conditional GAN},
  author={Lu, Chang and Reddy, Chandan K and Wang, Ping and Nie, Dong and Ning, Yue},
  journal={IEEE Transactions on Knowledge and Data Engineering},
  volume={36},
  number={4},
  pages={1728--1740},
  year={2023},
  publisher={IEEE}
}

@article{karami2024timehr,
  title={TimEHR: Image-based Time Series Generation for Electronic Health Records},
  author={Karami, Hojjat and Hartley, Mary-Anne and Atienza, David and Ionescu, Anisoara},
  journal={arXiv preprint arXiv:2402.06318},
  year={2024}
}

@misc{harzing2007pop,
  author       = {Harzing, A. W.},
  title        = {Publish or Perish},
  year         = {2007},
  howpublished = {\url{https://harzing.com/resources/publish-or-perish}}
}

\appendix



\renewcommand{\thesection}{S\arabic{section}}
\renewcommand\thefigure{S\arabic{figure}}    
\renewcommand\thetable{S\arabic{table}}    
\numberwithin{equation}{section}
\makeatletter 
\renewcommand\theequation{\thesection.\arabic{equation}}    
\newcommand{\section@cntformat}{Supplement \thesection:\ }
\makeatother

\newpage
\section*{Supplementary Materials}

\setcounter{figure}{0}
\setcounter{table}{0}

\begin{figure}[H]
    \centering
    \includegraphics[width=0.75\linewidth]{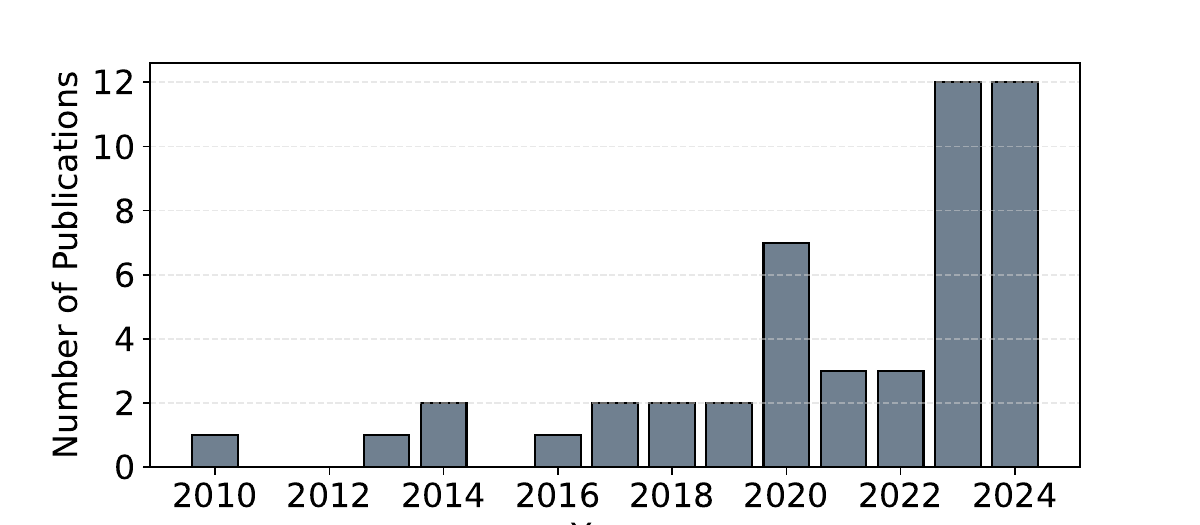}
    \caption{The number of publications on synthetic EHR generation methods in recent years.}
    \label{fig:num-of-pub}
\end{figure}

\begin{figure}[H]
    \centering
    \includegraphics[width=0.9\linewidth]{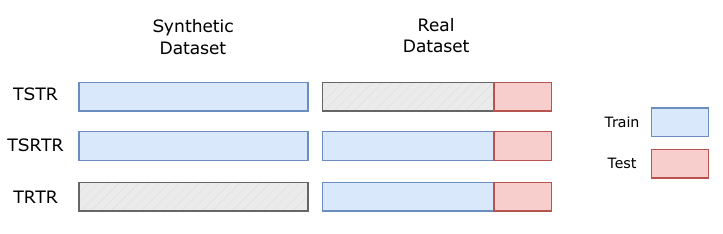}
    \caption{Three different training approaches described in predictive utility. Real data (either MIMIC-III or MIMIC-IV) are split into training and testing subsets with 4:1 ratio. For TRTR, ML models are trained on the training subset of the real data, and are tested on the real testing subset. For TSTR, ML models are trained on all the synthetic data, and tested on the real testing subset. For TSRTR, the real training subset is stacked with all the synthetic data to train ML models, and the testing subset of the real data is used to evaluate the trained ML models.}
    \label{fig:pred-utility}
\end{figure}

\begin{table}[H]
\centering
\caption{Summary statistics of the MIMIC-III and MIMIC-IV data. For age, gender, ethnicity, and diseases, the values represent percentages; counts are shown in parentheses.}
\label{tab:summary}
\resizebox{0.58\columnwidth}{!}{
\begin{tabular}{lcc} \toprule
 & MIMIC-III & MIMIC-IV                \\ \midrule
\textbf{\# of patients}                       & 46,520  &      180,733       \\
\textbf{\# of ICU patients} & 46,520 & 50,920 \\
\textbf{\# of Admissions} & 58,976 & 431,231 \\
\textbf{Demographics}     & \multicolumn{1}{l}{} & \\
Age  ($\%$ (count))                     & \multicolumn{1}{l}{} & \\
\quad 0-9                       & 16.9 (7,874)       &  0.0 (0) \\
\quad 10-19                     & 0.8 (371)          &  1.8 (3,210) \\
\quad 20-29                     & 3.6 (1,672)        &  12.6 (22,711) \\
\quad 30-39                     & 4.4 (2,029)        &  11.6 (20,937) \\
\quad 40-49                     & 8.7 (4,040)        &  12.2 (21,965) \\
\quad 50-59                     & 14.2 (6,586)       &  16.7 (30,167) \\
\quad 60-69                     & 16.9 (7,845)       &  17.4 (31,474) \\
\quad 70-79                     & 17.0 (7,918)       &  14.0 (25,323) \\
\quad 80+                       & 17.6 (8,185)       &  13.8 (24,946) \\
Gender  ($\%$ (count))                  & \multicolumn{1}{l}{}     & \\
\quad Female                    & 43.9 (20,399)      & 53.0 (95,729) \\
\quad Male                      & 56.2 (26,121)      &  47.0 (85,004) \\
Ethnicity  ($\%$ (count))                & \multicolumn{1}{l}{} 
   & \\
\quad Asian                     & 3.6 (1,690)        & 4.2 (7,583) \\
\quad Black/African American    & 8.3 (3,864)        &  13.0 (23,538) \\
\quad Hispanic                  & 3.5 (1,642)        &  5.4 (9,766) \\
\quad White                     & 69.6 (32,372)      &  67.2 (121,421) \\
\quad Other/Unknown             & 14.9 (6,952)       &  10.2 (18,425) \\
\textbf{Diseases} ($\%$ (count))             &                    &  \\
Anxiety                         & 11.0 (5,112)              & 13.0 (23,412) \\
Cancer                          & 52.9 (24,620)              & 40.2 (72,683) \\    
Depression                      & 16.0 (7,431)              & 20.8 (37,551) \\
Diabetes                        & 25.4 (11,811)             & 21.7 (39.211) \\
Hypertension                    & 47.6 (22,147)              & 46.0 (83,107) \\
Obesity                         & 5.4 (2,518)                & 12.6 (22,855) \\
\textbf{EHR Characteristics}       & \multicolumn{1}{l}{} & \\
Encounter per person      & \multicolumn{1}{l}{} & \\
\quad Minimum                   & 1.0                &  1.0  \\
\quad Median                    & 1.0                &  1.0  \\
\quad Mean                      & 1.3                &  2.4 \\
\quad Maximum                   & 42.0               &  238.0  \\
Unique phecode per person & \multicolumn{1}{l}{} & \\
\quad Minimum                   & 0.0                &  0.0  \\
\quad Median                    & 26.0               &  14.0  \\
\quad Mean                      & 31.3               &  26.0 \\
\quad Maximum                   & 203.0              &  260.0  \\
Length of follow-up (years)       & \multicolumn{1}{l}{} & \\
\quad Minimum                   & 0.0               &   0.0  \\
\quad Median                    & 0.0                &  0.0  \\
\quad Mean                      & 0.3               &   1.2 \\
\quad Maximum                   & 11.5              &   12.0 \\ \bottomrule
\end{tabular}}
\end{table}

\begin{figure}[H]
    \centering
    \includegraphics[width=1.0\linewidth]{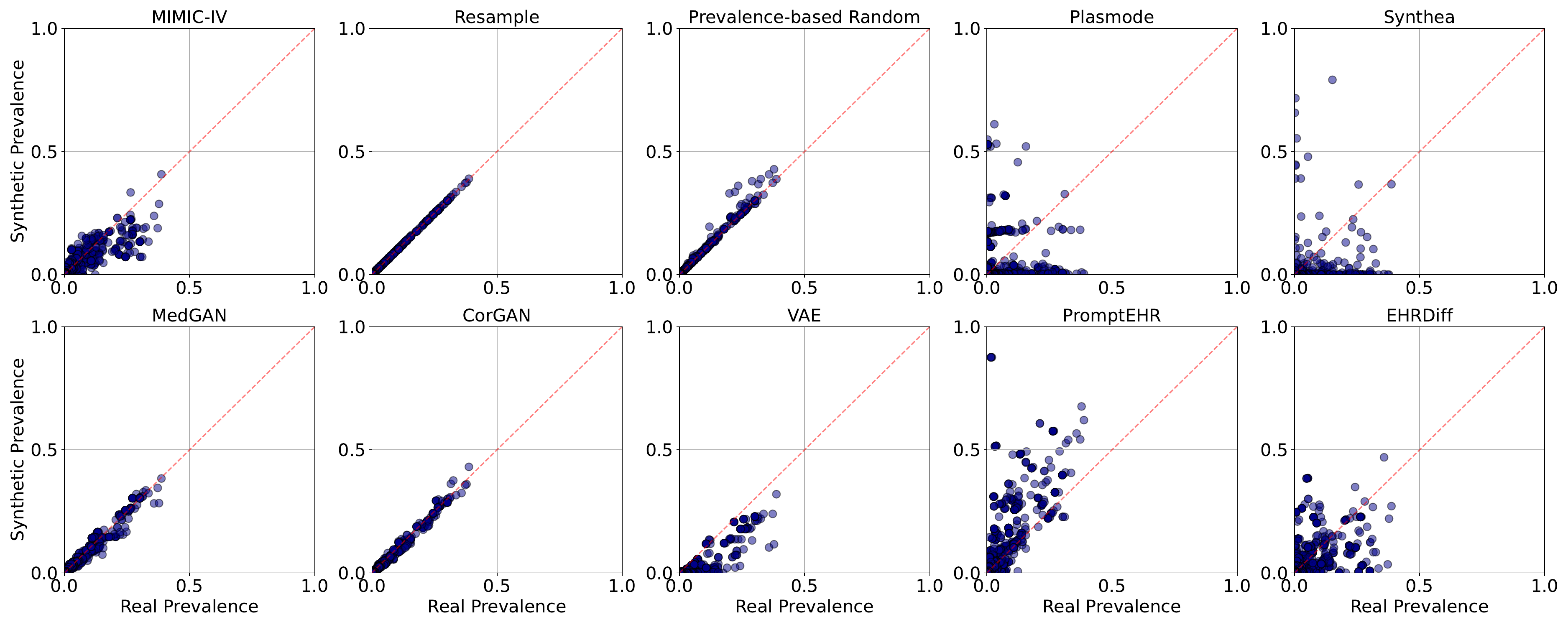}
    \caption{Scatterplots comparing the phecode-wise prevalences estimated from the MIMIC-IV data or synthetic data with the one from the real data in the MIMIC-III dataset. The red dashed lines indicate the perfect agreement of estimated prevalence between the synthetic and the MIMIC-III real data.}
    \label{fig:real-versus-synthetic-prevalence}
\end{figure}

\begin{figure}[H]
    \centering
    \begin{subfigure}[b]{0.48\textwidth}
        \centering
        \includegraphics[width=\linewidth]{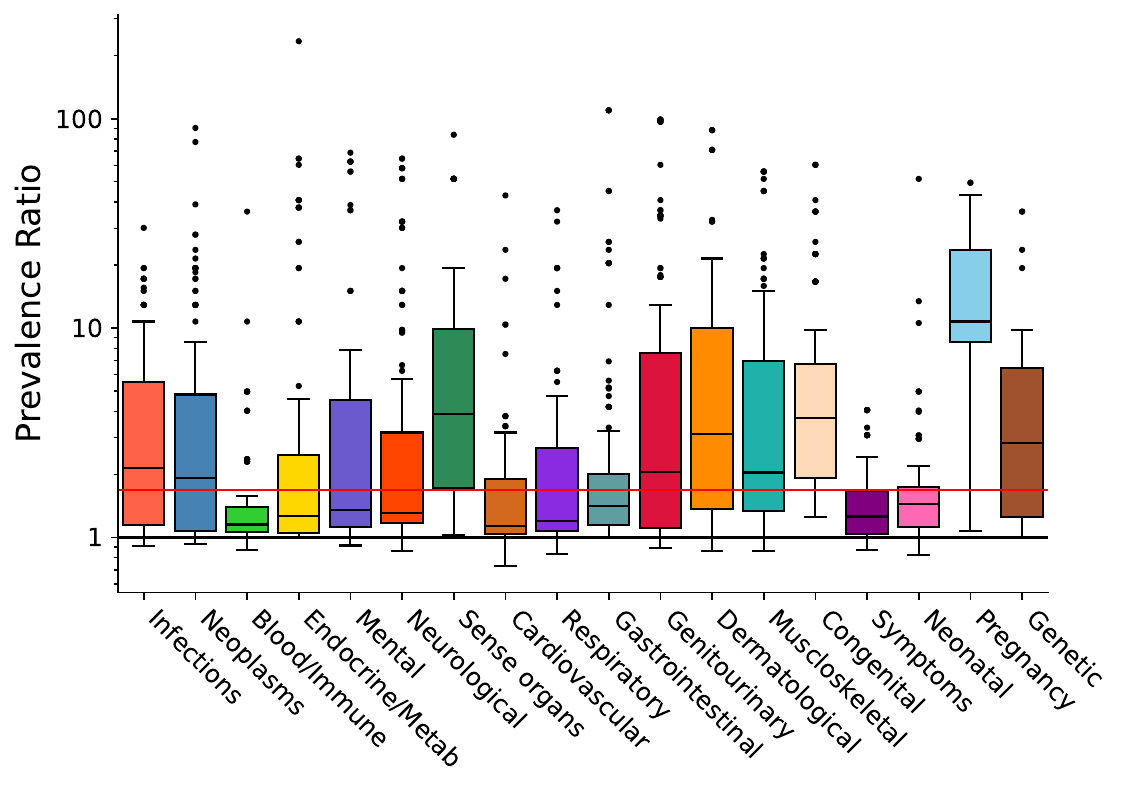}
        \caption{MIMIC-III.}
        \label{fig:phewas-boxplot-mimic3}
    \end{subfigure}
    \hfill 
    \begin{subfigure}[b]{0.48\textwidth}
        \centering
        \includegraphics[width=\linewidth]{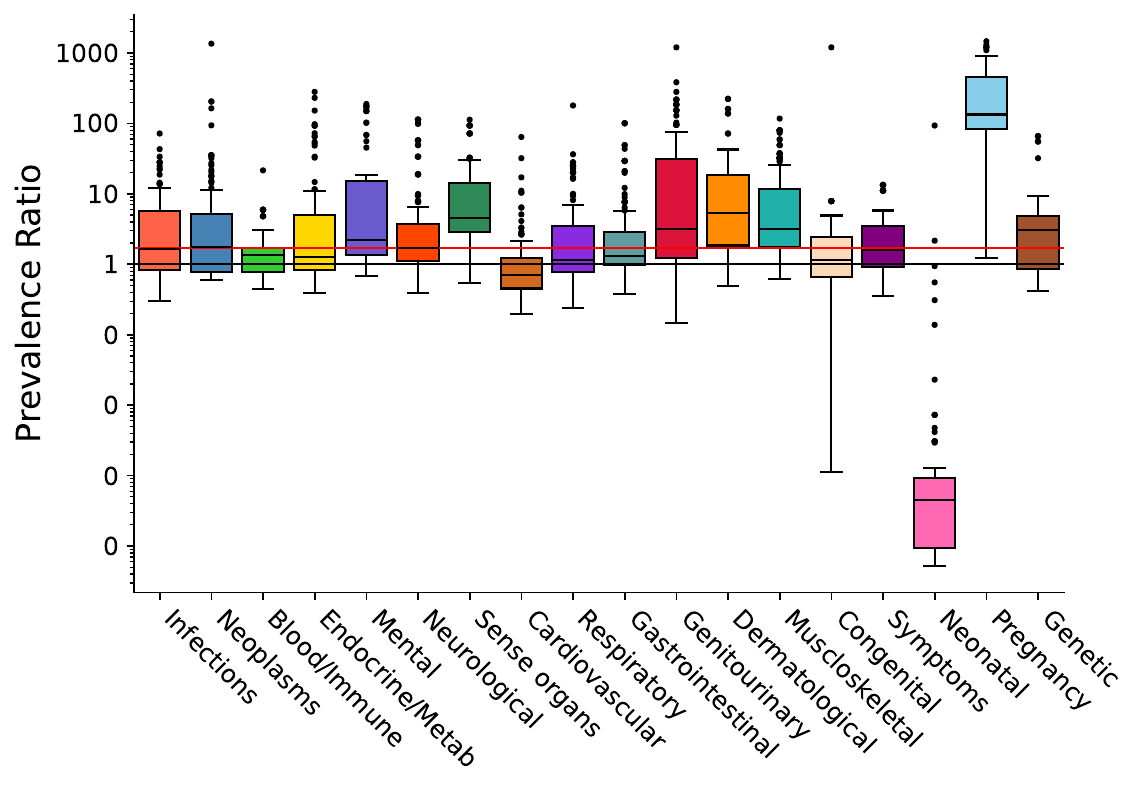}
        \caption{MIMIC-IV.}
        \label{fig:phewas-boxplot-mimic4}
    \end{subfigure}
    \caption{Boxplots of the ratio of the estimated PhecodeX-wise prevalence between the real and the synthetic data generated by CorGAN, grouped by PhecodeX categories defined in \cite{10.1093/bioinformatics/btad655}. Y-axis represents the ratio of estimated prevalence, less than 1 means underestimation by the synthetic data. X-axis represents all the PhecodeX categories. Red line indicates the median of prevalence ratio of all phecodes. The significant over-estimation within the neonatal phenotype group in the MIMIC-IV dataset is because of its lack of neonatal population.}
    \label{fig:phewas-boxplot}
\end{figure}



\begin{figure}[H]
    \centering
    \begin{subfigure}[b]{0.8\textwidth}
        \centering
        \includegraphics[width=\linewidth]{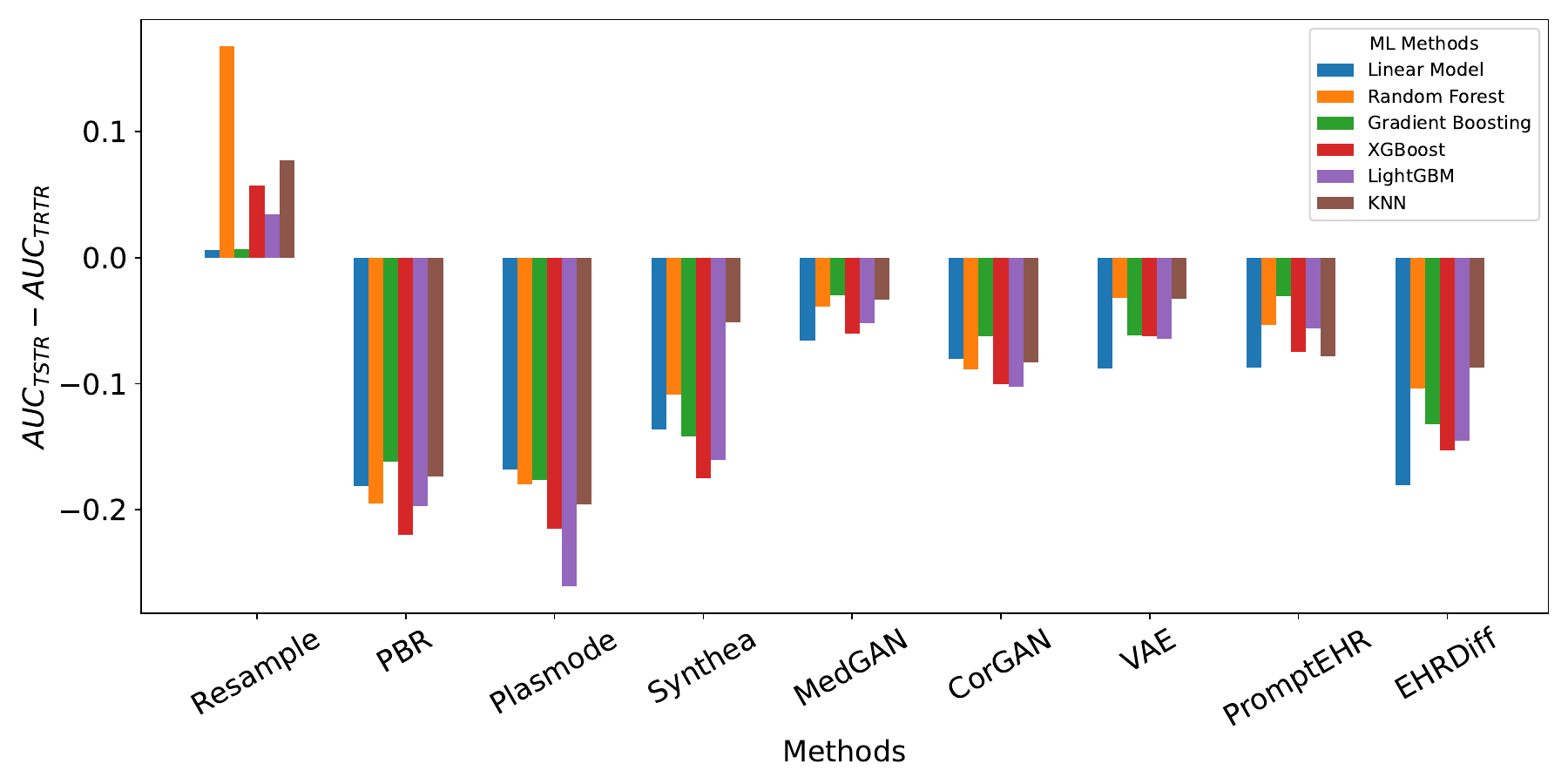}
        \caption{Evaluated on MIMIC-III.}
        \label{fig:mimic3-tstr-more-ml}
    \end{subfigure}
    \begin{subfigure}[b]{0.8\textwidth}
        \centering
        \includegraphics[width=\linewidth]{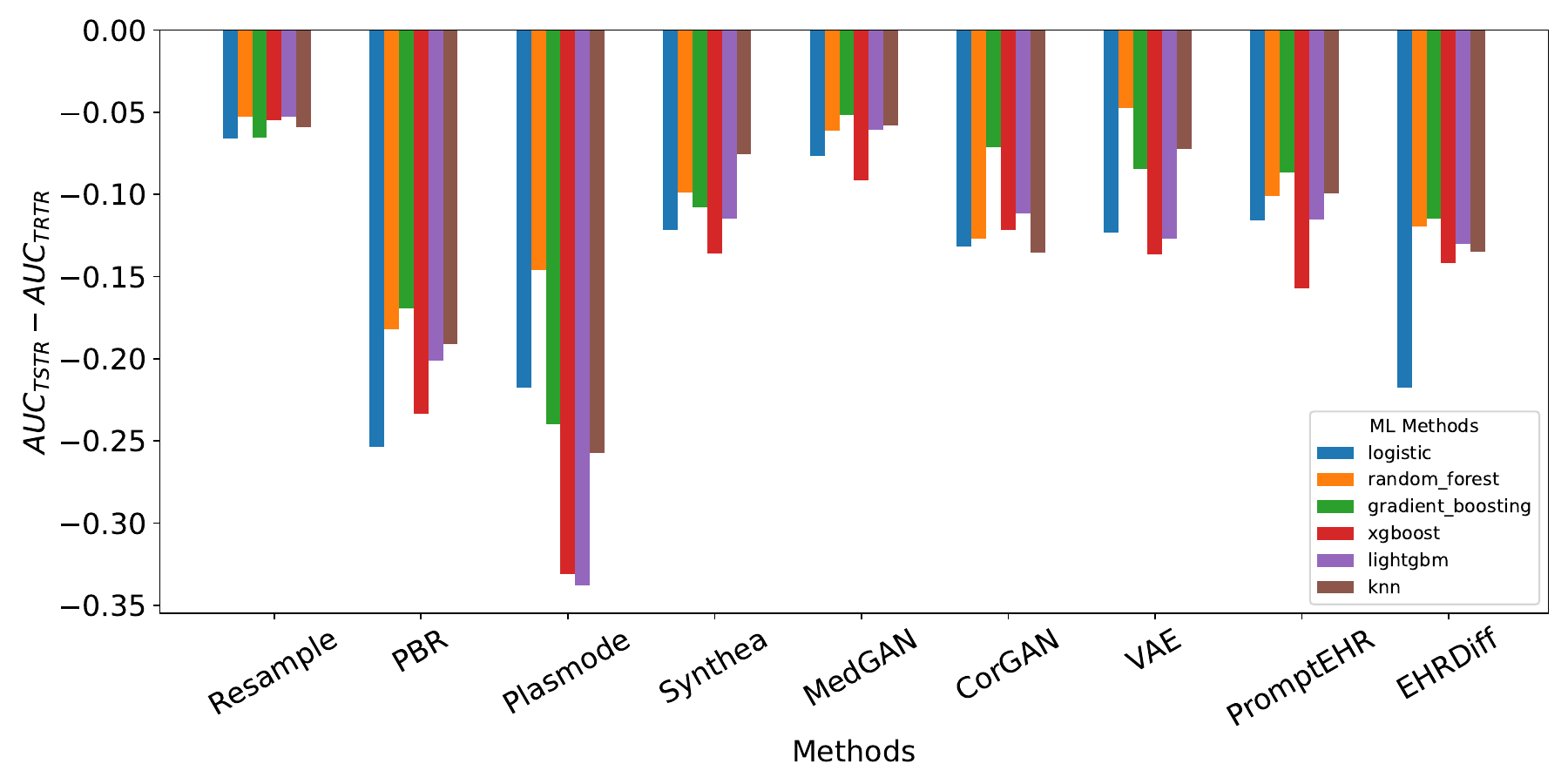}
        \caption{Evaluated on MIMIC-IV.}
        \label{fig:mimic4-tstr-more-ml}
    \end{subfigure}
    \caption{The difference of AUC between TSTR and TRTR based on different ML methods trained on either MIMIC-III or MIMIC-IV. The large performance gap in {\it Resample} is because of the fact that the TRTR baseline is trained on the training subset of the MIMIC-III/IV dataset. On the other hand, {\it Resample} bootstrap over the entire MIMIC-III/IV datasets to ``synthesize'' data.}
    \label{fig:advanced-ml-tstr}
\end{figure}

\begin{figure}[H]
    \centering
    \begin{subfigure}[b]{0.8\textwidth}
        \centering
        \includegraphics[width=\linewidth]{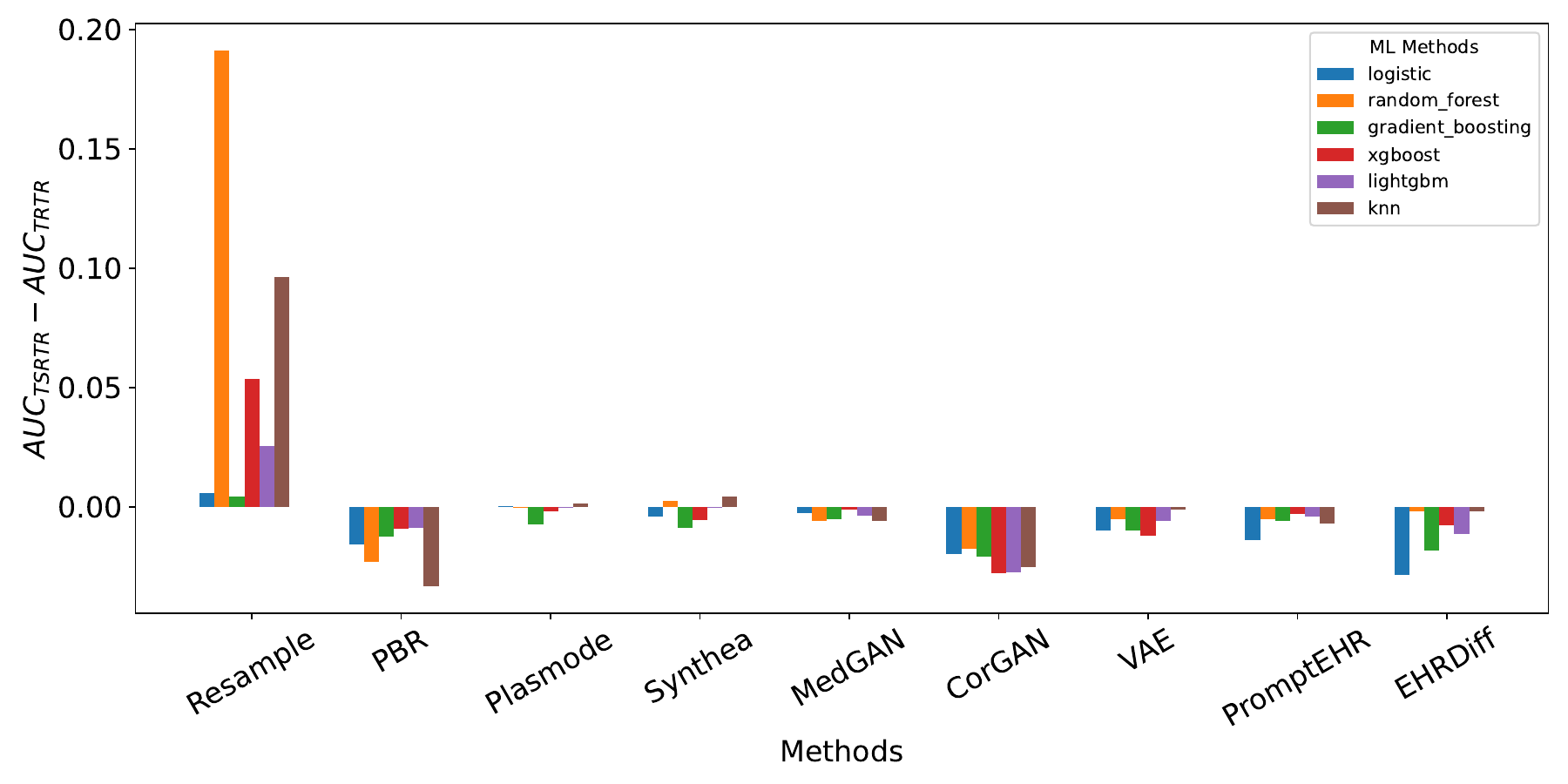}
        \caption{Evaluated on MIMIC-III.}
        \label{fig:mimic3-tsrtr-more-ml}
    \end{subfigure}
    \begin{subfigure}[b]{0.8\textwidth}
        \centering
        \includegraphics[width=\linewidth]{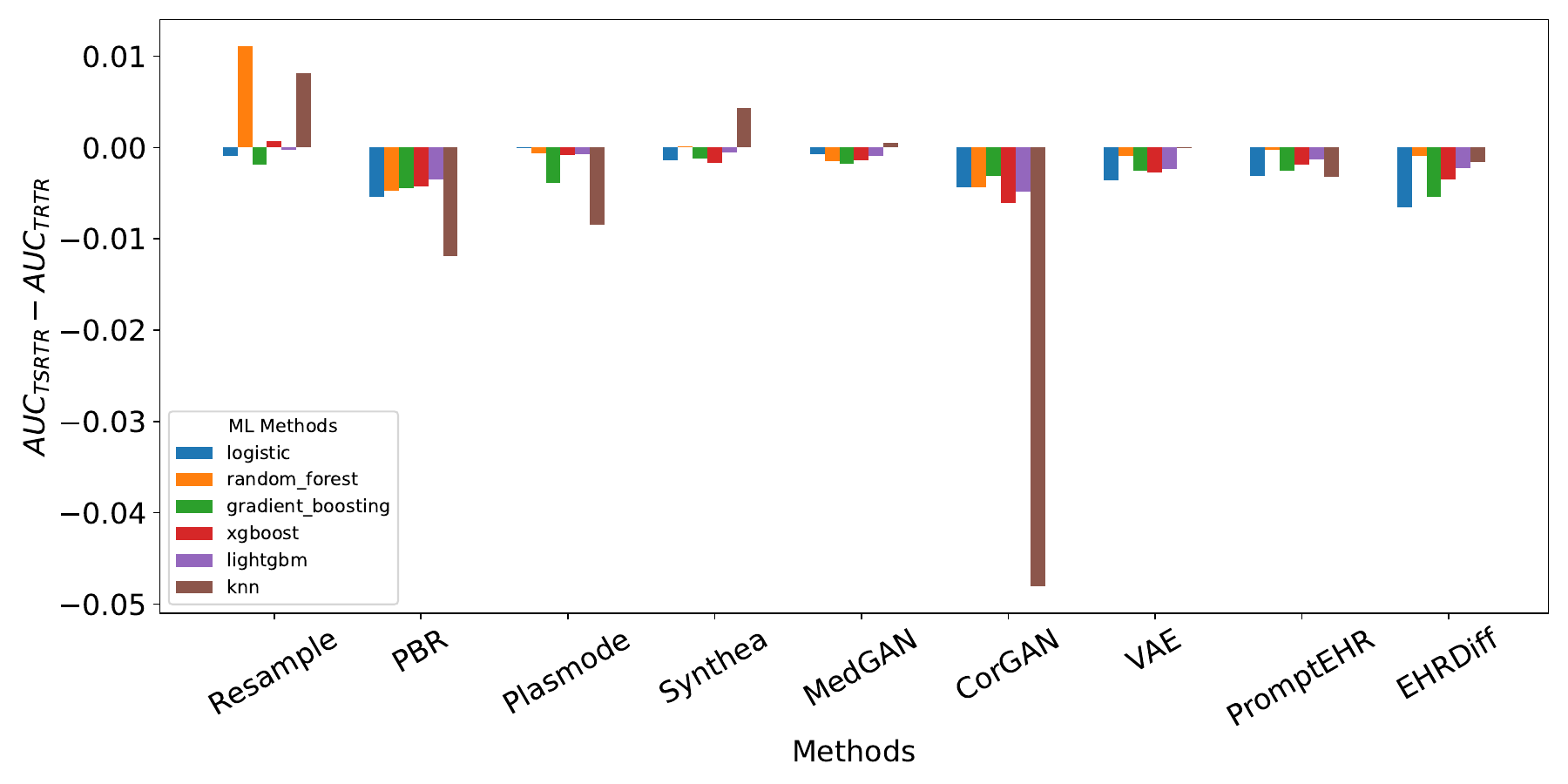}
        \caption{Evaluated on MIMIC-IV.}
        \label{fig:mimic4-tsrtr-more-ml}
    \end{subfigure}
    \caption{The difference of AUC between TSRTR and TRTR based on different ML methods trained on either MIMIC-III or MIMIC-IV. The large performance gap in {\it Resample} is because of the fact that the TRTR baseline is trained on the training subset of the MIMIC-III/IV dataset. On the other hand, {\it Resample} bootstrap over the entire MIMIC-III/IV datasets to ``synthesize'' data.}
    \label{fig:advanced-ml-tsrtr}
\end{figure}



\begin{figure}[H]
    \centering
    \begin{subfigure}[b]{1.0\textwidth}
        \centering
        \includegraphics[width=\linewidth]{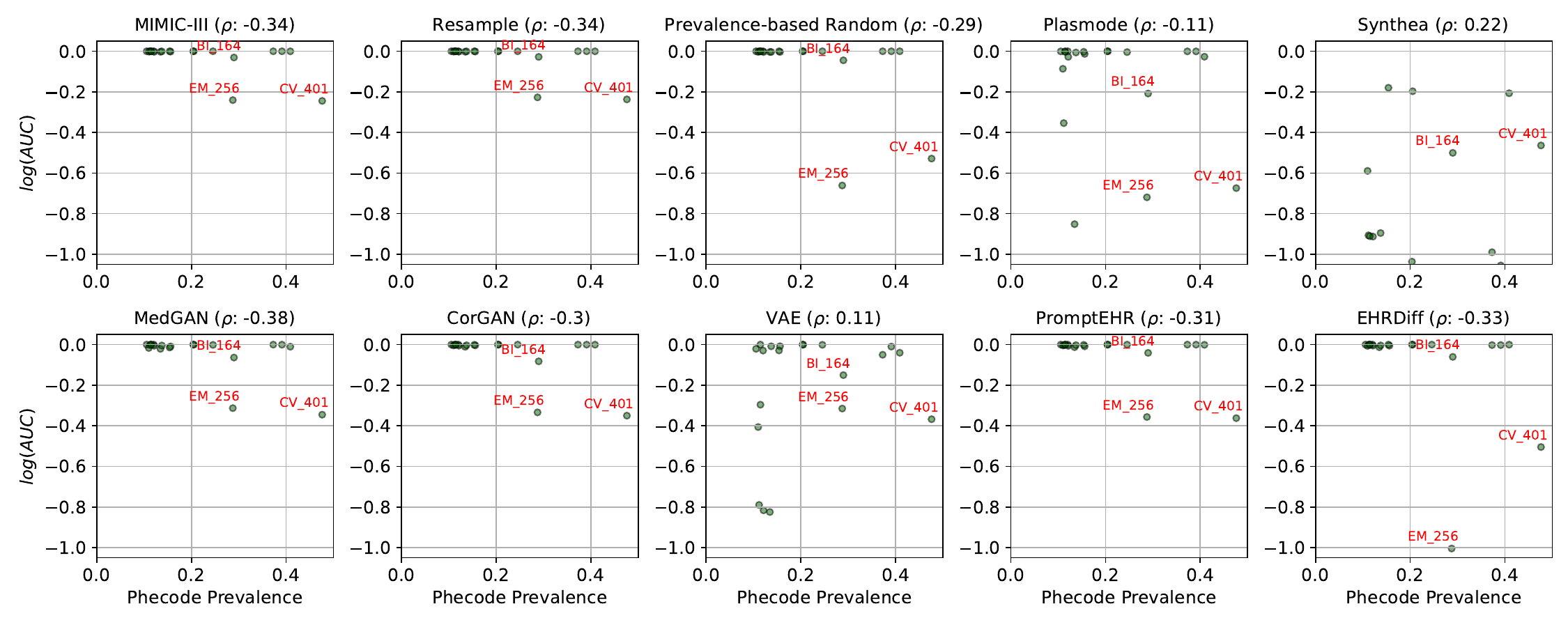}
        \caption{MIMIC-III.}
        \label{fig:tstr-code-frequency-a}
    \end{subfigure}
    \begin{subfigure}[b]{1.0\textwidth}
        \centering
        \includegraphics[width=\linewidth]{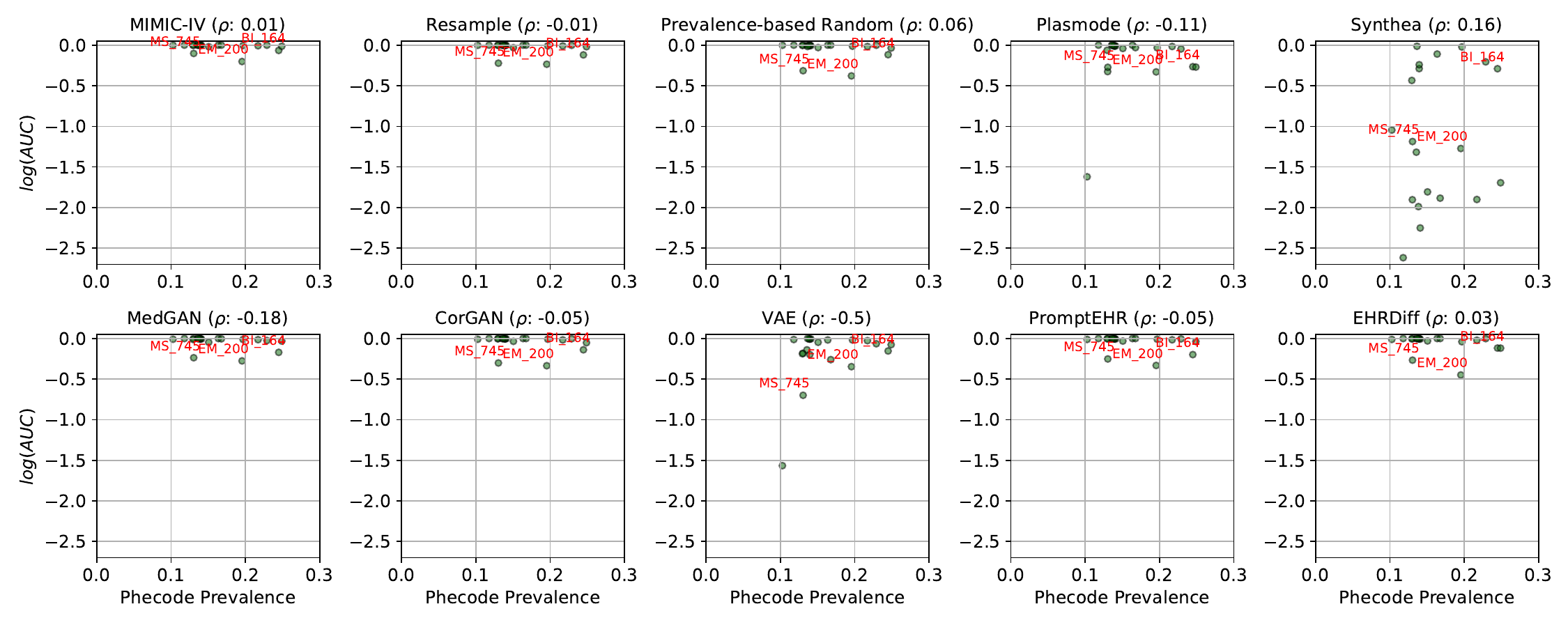}
        \caption{MIMIC-IV.}
        \label{fig:tstr-code-frequency-b}
    \end{subfigure}
    \caption{\textcolor{black}{Scatterplots comparing the AUC$_{TSTR}$ and code frequency of selected phecodes in the MIMIC-III and MIMIC-IV datasets. Each scatterplot corresponds to a generation method and includes 20 phenotypes randomly selected from those with a prevalence greater than 0.1. Spearman’s $\rho$ correlation is reported for each plot. The top-left scatterplots display TRTR performance based on real data. Outlier phecodes in the TRTR evaluation from real data are annotated across all TSTR-versus-prevalence scatterplots for MIMIC-III and MIMIC-IV, respectively. Specifically, the outliers are Hypertension (\texttt{CV\_401}), Disorders of fluid, electrolyte, and acid-base balance (\texttt{EM\_256}), and Anemia (\texttt{BI\_164}) in MIMIC-III; Anemia (\texttt{BI\_164}), Fractures (\texttt{MS\_745}), and Disorders of the thyroid gland (\texttt{EM\_200}) in MIMIC-IV.}}
    \label{fig:tstr-code-frequency}
\end{figure}

\begin{table}[H]
\centering
\caption{\textcolor{black}{AUC$_{TSTR}$ performance comparison of different methods for additional diseases in MIMIC-III and MIMIC-IV. PhecodeX codes, prevalence, and the AUC$_{TRTR}$ results of the selected diseases are reported in the table for reference.}}
\resizebox{1.0\columnwidth}{!}{
\begin{tabular}{l|ccccc|ccccc}
\toprule
 & \multicolumn{5}{c|}{\textbf{MIMIC-III}} & \multicolumn{5}{c}{\textbf{MIMIC-IV}} \\ 
\midrule
Diseases & Hypertension & Diabetes & Anxiety & Depression & Obesity 
& Hypertension & Diabetes & Anxiety & Depression & Obesity \\ 
PhecodeX & \texttt{CV\_401} & \texttt{EM\_202} & \texttt{MB\_288} & \texttt{MB\_286} & \texttt{EM\_236} & \texttt{CV\_401} & \texttt{EM\_202} & \texttt{MB\_288} & \texttt{MB\_286} & \texttt{EM\_236} \\ 
Prevalence (\%)& 47.6 & 25.4 & 11.0 & 16.0 & 5.4 
& 46.0 & 21.7 & 13.0 & 20.8 & 12.6 \\ 
TRTR Results & 78.3 & 99.6 & 99.8 & 88.1 & 98.9 & 84.6 & 96.2 & 99.9 & 93.4 & 90.7 \\ 
\midrule
\textit{Resample}  & 78.9 & 99.7 & 99.9 & 89.2 & 99.4 & 77.9 & 90.0 & 99.9 & 91.5 & 80.5 \\
\textit{PBR}       & 59.4 & 99.5 & 99.7 & 80.6 & 98.7 & 59.3 & 90.4 & 99.7 & 80.0 & 80.1 \\
Synthea   & 63.1 & 67.9 & 53.1 & 53.1 & 54.9 & 72.6 & 75.8 & 64.8 & 57.1 & 65.0 \\
Plasmode  & 48.8 & 97.6 & 91.4 & 81.6 & 49.3 & 48.5 & 90.9 & 93.9 & 81.4 & 50.1 \\
MedGAN    & 71.9 & 98.8 & 98.4 & 84.6 & 95.5 & 77.0 & 92.4 & 98.9 & 89.7 & 75.7 \\
CorGAN    & 70.3 & 99.3 & 99.6 & 83.9 & 97.7 & 71.4 & 89.0 & 99.6 & 85.7 & 77.2 \\
VAE       & 69.5 & 97.4 & 66.3 & 58.7 & 52.1 & 72.3 & 88.3 & 82.9 & 69.3 & 34.0 \\
PromptEHR & 69.9 & 99.0 & 99.7 & 82.6 & 97.0 & 72.9 & 89.6 & 99.6 & 80.0 & 80.9 \\
EHRDiff   & 61.1 & 99.1 & 98.2 & 79.4 & 96.8 & 62.9 & 89.6 & 99.5 & 79.7 & 77.8 \\ \midrule
Average & 65.9 & 95.4 & 89.6 & 77.1 & 82.4 & 68.3 & 88.5 & 93.2 & 79.4 & 69.1 \\
\bottomrule 
\end{tabular}}
\label{tab:disease_comparison}
\end{table}

\begin{figure}[H]
    \centering
    \begin{subfigure}[b]{0.8\textwidth}
        \centering
        \includegraphics[width=\linewidth]{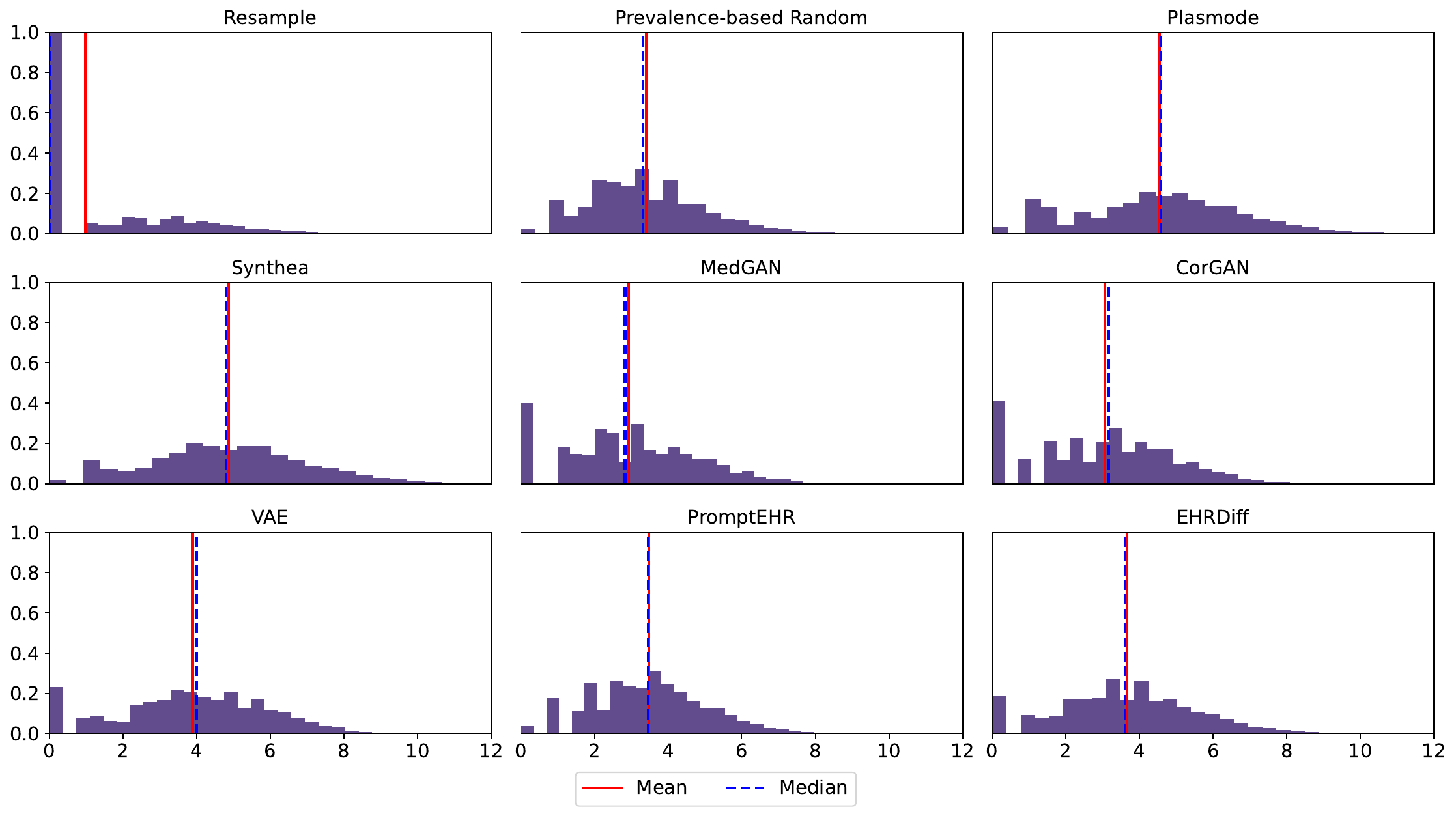}
        \caption{Evaluated on MIMIC-III.}
        \label{fig:mimic3-mir-hist}
    \end{subfigure}
    \begin{subfigure}[b]{0.8\textwidth}
        \centering
        \includegraphics[width=\linewidth]{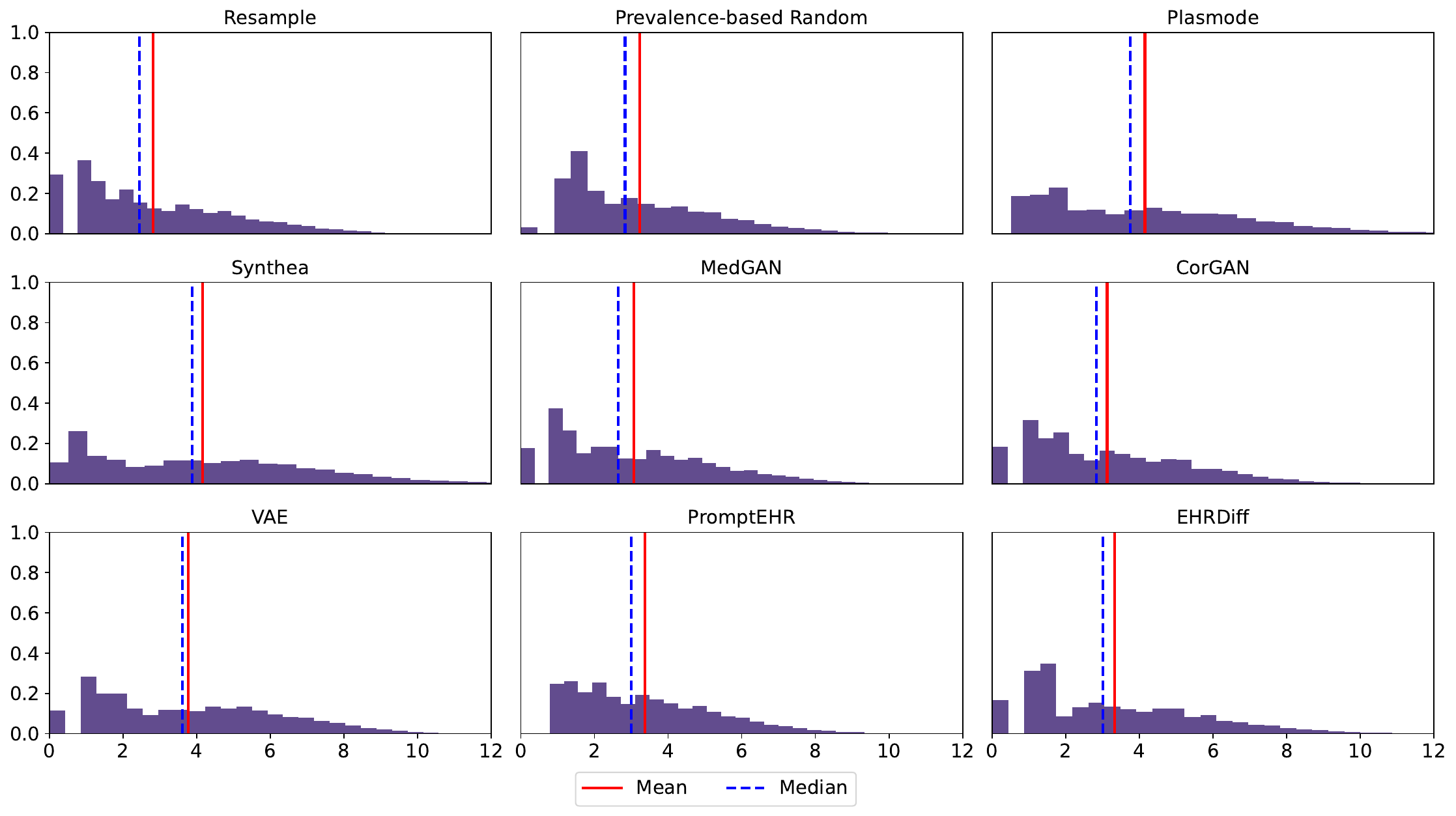}
        \caption{Evaluated on MIMIC-IV.}
        \label{fig:mimic4-mir-hist}
    \end{subfigure}
    \caption{\textcolor{black}{Histogram of the Euclidean distances from each real EHR sample in MIMIC-III or MIMIC-IV datasets to its nearest synthetic EHR sample generated by included generation methods. The red solid line represents the sample mean, and the blue dotted line represents the sample median. Patients with no diagnosed phecode are excluded from the calculation.}}
    \label{fig:mir-hist}
\end{figure}

\begin{figure}[H]
    \centering
    \begin{subfigure}[b]{0.48\textwidth}
        \centering
        \includegraphics[width=\linewidth]{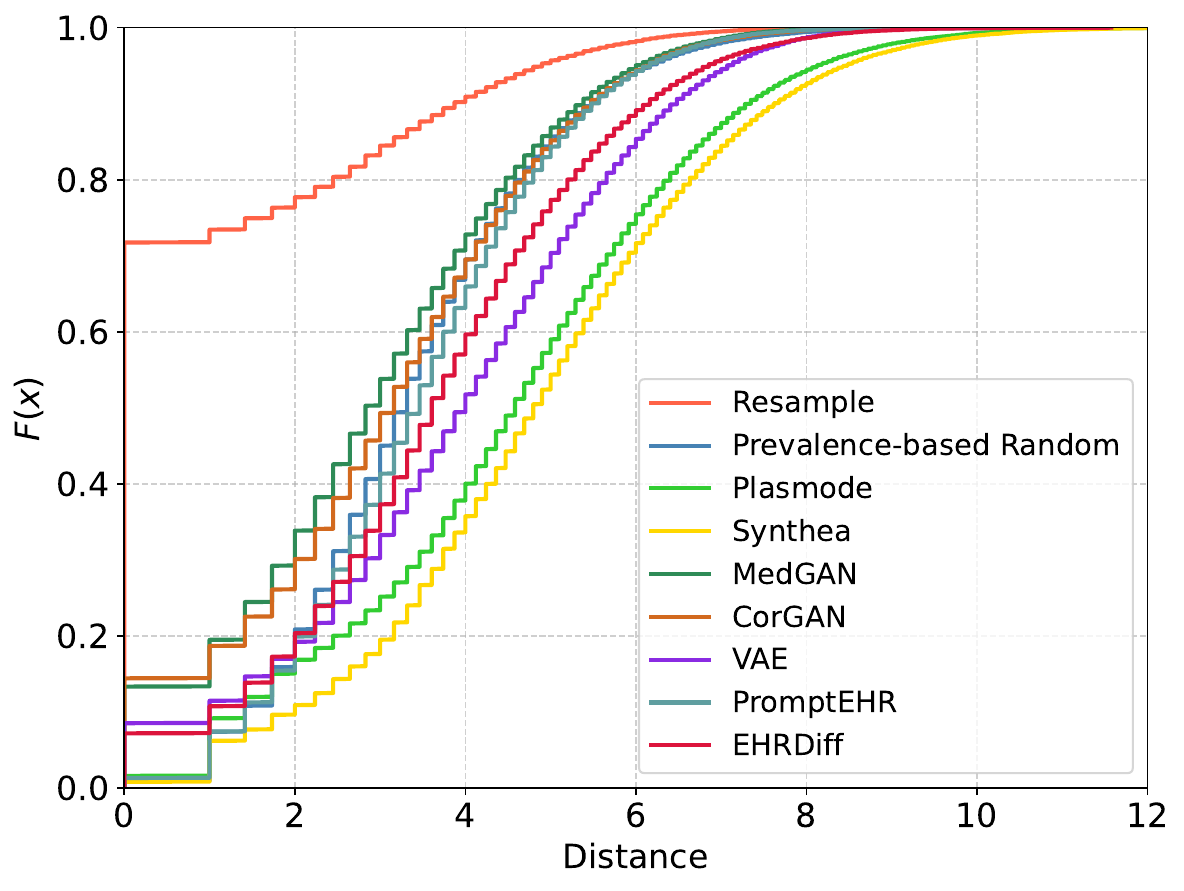}
        \caption{Evaluated on MIMIC-III.}
        \label{fig:mimic3-mir-cdf}
    \end{subfigure}
    \hfill 
    \begin{subfigure}[b]{0.48\textwidth}
        \centering
        \includegraphics[width=\linewidth]{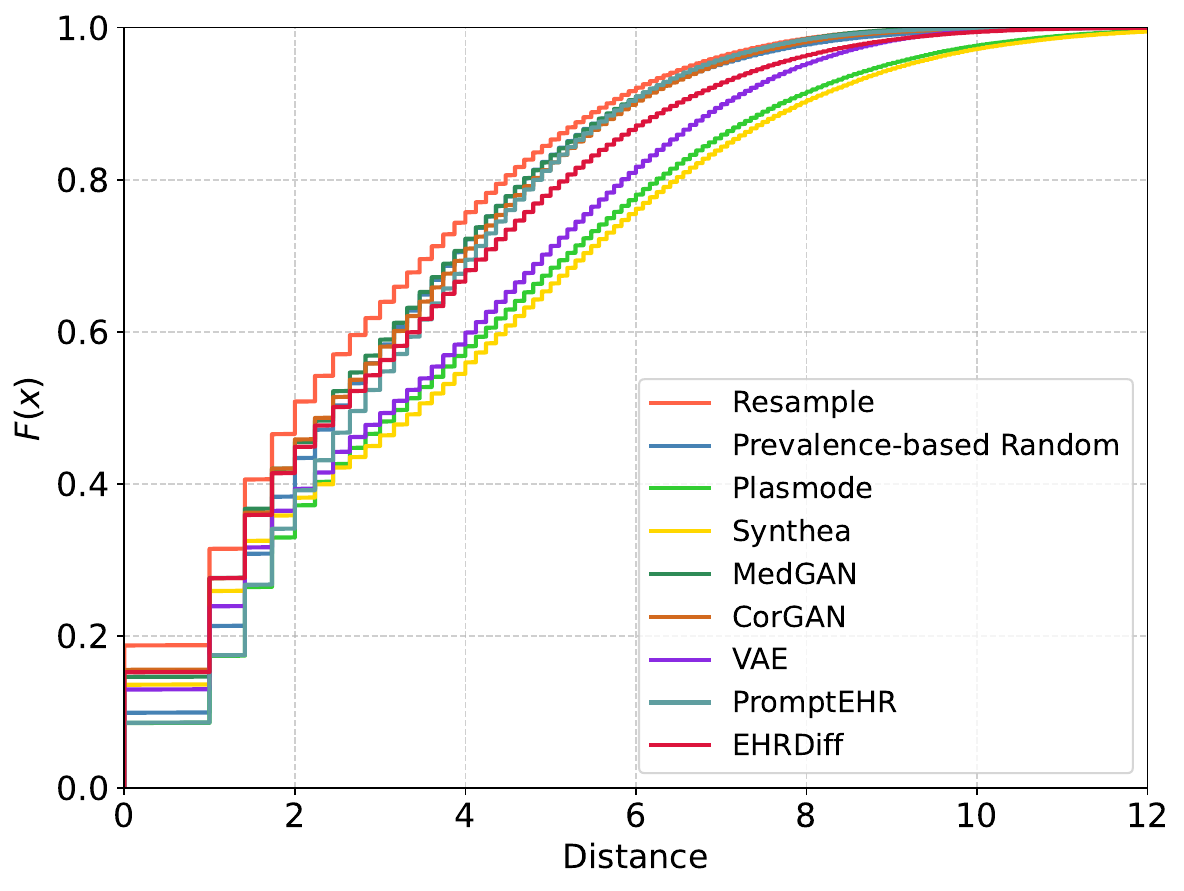}
        \caption{Evaluated on MIMIC-IV.}
        \label{fig:mimic4-mir-cdf}
    \end{subfigure}
    \caption{\textcolor{black}{Cumulative distribution function (CDF) plots of the Euclidean distances from each real EHR sample in MIMIC-III or MIMIC-IV datasets to its nearest synthetic EHR sample generated by included generation methods. Higher curve values at a given distance cutoff indicate a greater proportion of real EHR samples with synthetic samples within that distance. Patients with no diagnosed phecodes are excluded from the calculation.}}
    \label{fig:mir-cdf}
\end{figure}

\begin{table}[H]
\centering
\caption{\textcolor{black}{Proportion of real EHR samples in MIMIC-III and MIMIC-IV that are perfectly matched with at least one synthetic EHR sample generated by selected methods. The values represent percentage (\%). Patients with no diagnosed phecode are excluded from the calculation.}}
\resizebox{0.6\columnwidth}{!}{
\begin{tabular}{l|c|c}
\toprule
\textbf{Method}   & \textbf{MIMIC-III} & \textbf{MIMIC-IV} \\ 
\midrule
\textit{Resample}       & 71.74  & 11.17  \\ 
\textit{PBR}            & 0.90   & 1.50   \\ 
Synthea        & 0.78   & 5.52  \\
Plasmode       & 1.56   & 0.02   \\ 
MedGAN         & 13.28  & 6.65  \\ 
Corgan         & 14.38  & 7.66  \\ 
VAE            & 8.49   & 4.83  \\ 
PromptEHR      & 1.28   & 0.06   \\ 
EHRDiff        & 7.17   & 7.32  \\ 
\bottomrule 
\end{tabular}}
\label{tab:exact_match}
\end{table}

\begin{table}[]
\centering
\caption{Computational time cost of different synthetic EHR generation methods. Time is recorded for generating 100 samples. \dag: the generation time of Plasmode does not increase linearly with size of generation (e.g., Plasmode spends 17,225 seconds in generating 50,000 synthetic samples). The top-2 fastest results are \textbf{bolded}.}
\label{tab:computational-cost}
\resizebox{0.6\columnwidth}{!}{%
\begin{tabular}{lr} \toprule
 & \multicolumn{1}{l}{Computational cost (sec/100 samples)} \\ \midrule
PromptEHR & 71.0 \\
MedGAN & \textbf{0.08} \\
CorGAN & 0.26 \\
VAE & \textbf{0.16} \\
Synthea & 24.0 \\
Plasmode & $12291.5^\dag$ \\
EHRDiff & 0.65 \\ \bottomrule
\end{tabular}%
}
\end{table}

\begin{figure}[H]
    \centering    
    \includegraphics[width=1.0\columnwidth]{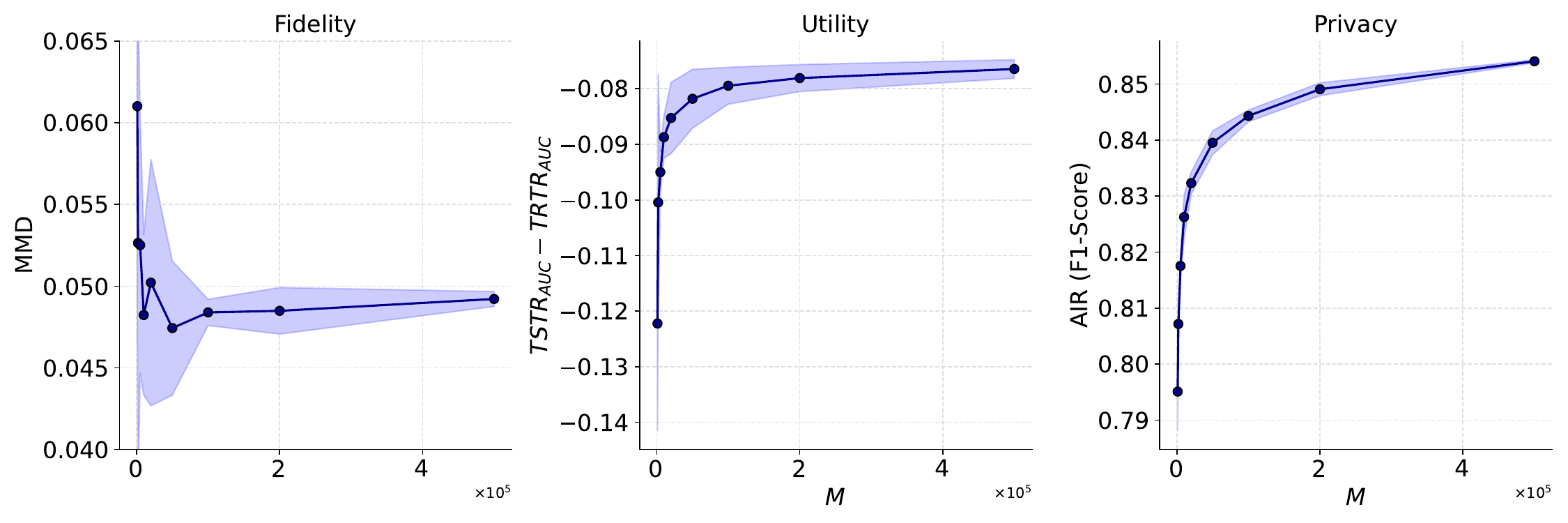}
    \caption{The performance under varying sizes of the synthetic data ($M$) generated by CorGAN trained on full real data. The chosen evaluation metrics for the three perspectives are Maximum Mean Discrepancy (MMD), Train on Synthetic, Test on Real (TSTR) measured by Area Under the Curve (AUC), and F1 score for Attribute Inference Risk (AIR), respectively. The synthetic data generation process is repeated five times, and the mean evaluation metrics are reported. Note that for MMD and F1 score in AIR, lower values indicate better performance.}
    \label{fig:gen_sample_size}
\end{figure}

\begin{figure}[H]
    \centering
    \includegraphics[width=1.0\columnwidth]{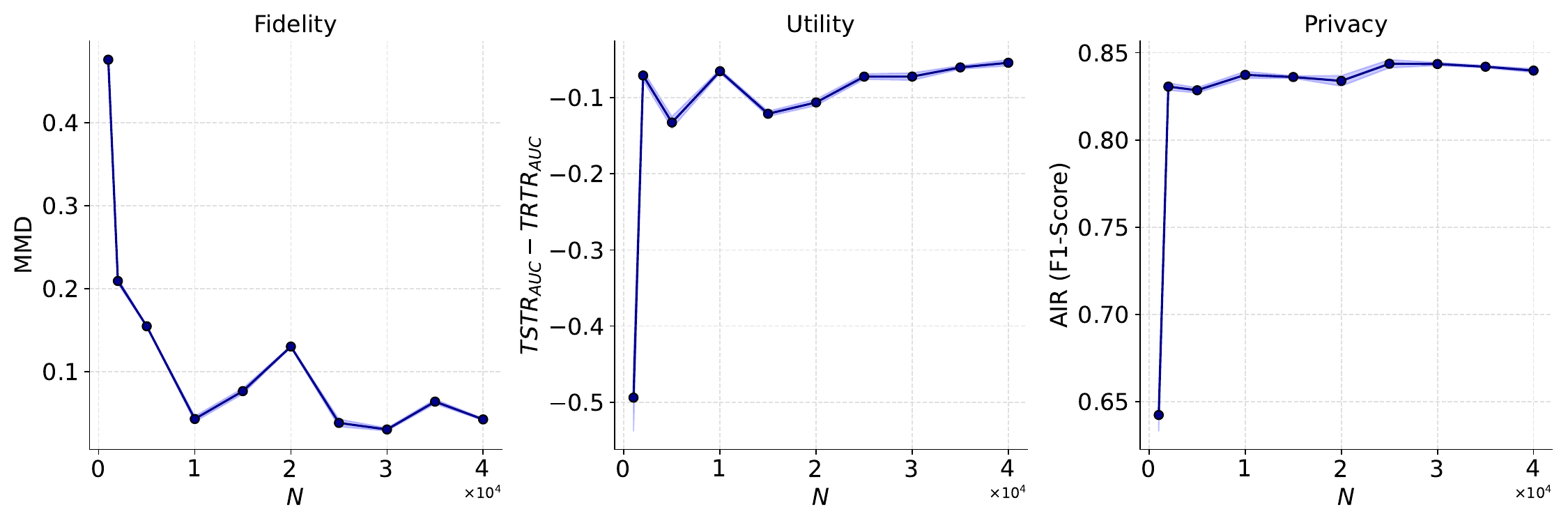}
    \caption{The performance of the synthetic data generated by CorGAN trained under varying sizes of the real data for training ($N$). The chosen evaluation metrics for the three perspectives are Maximum Mean Discrepancy (MMD), Train on Synthetic, Test on Real (TSTR) measured by Area Under the Curve (AUC), and F1 score for Attribute Inference Risk (AIR), respectively. The synthetic data generation process is repeated five times, and the mean evaluation metrics are reported. Note that for MMD and F1 score in AIR, lower values indicate better performance.}
    \label{fig:train_size}
\end{figure}

\begin{figure}[H]
    \centering
    \begin{subfigure}[b]{0.48\textwidth}
        \centering
        \includegraphics[width=\linewidth]{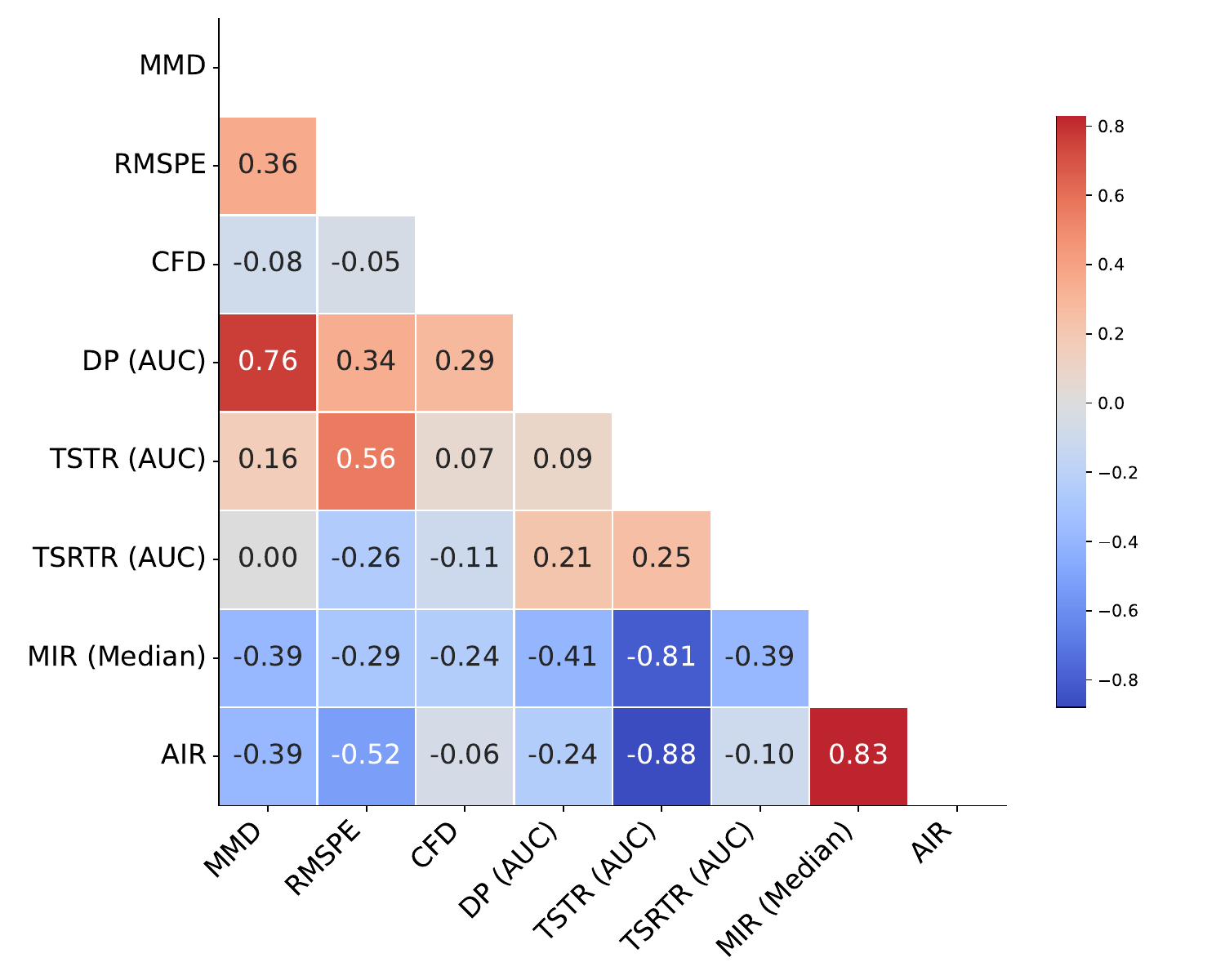}
        \caption{Evaluated on MIMIC-III.}
        \label{fig:correlation-evaluation-mimic3}
    \end{subfigure}
    \hfill 
    \begin{subfigure}[b]{0.48\textwidth}
        \centering
        \includegraphics[width=\linewidth]{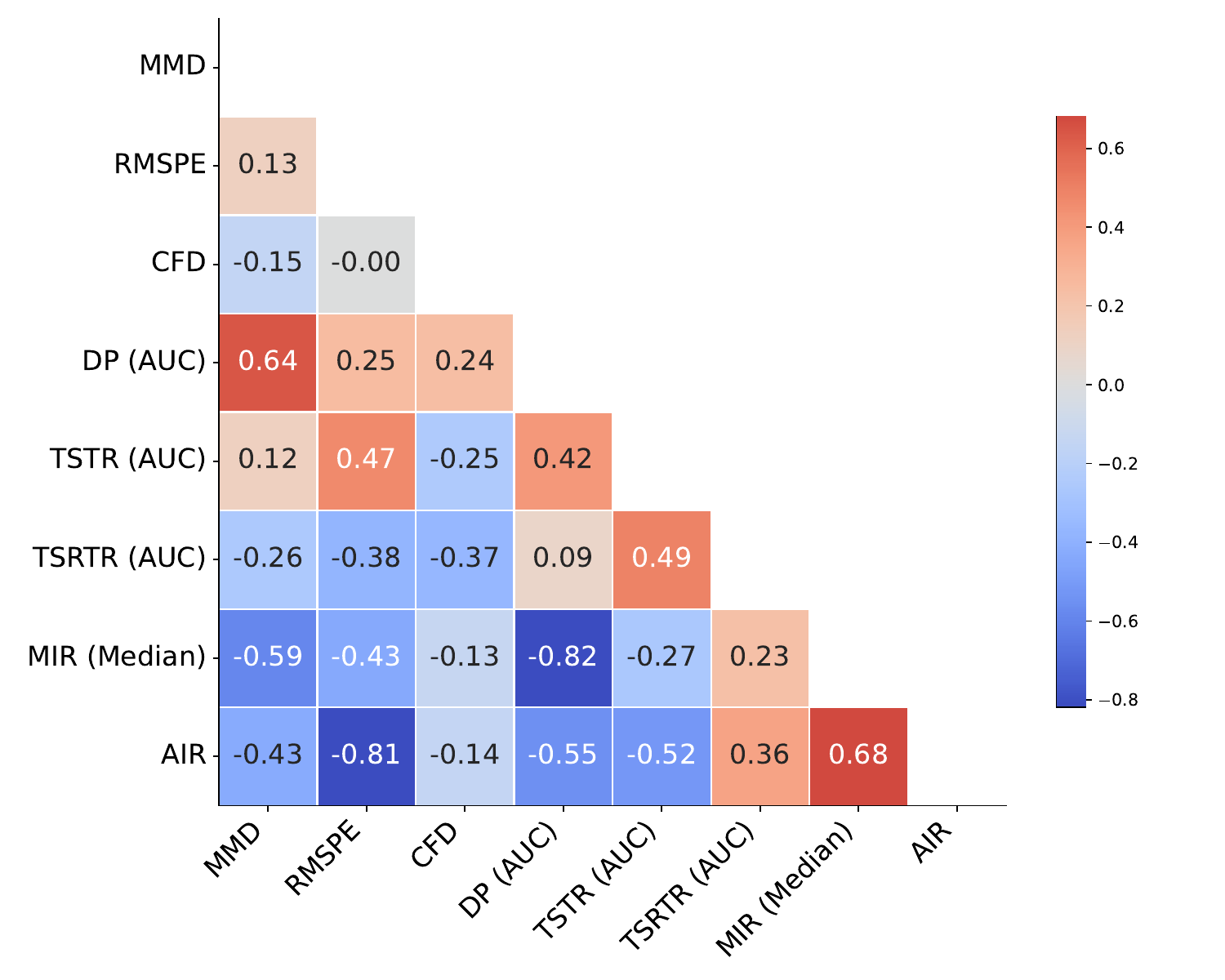}
        \caption{Evaluated on MIMIC-IV.}
        \label{fig:correlation-evaluation-mimic4}
    \end{subfigure}
    \caption{Within-dataset Pearson correlation among different evaluation metrics in Table \ref{tab:quantitative} for (a) MIMIC-III, and (b) MIMIC-IV, respectively. All ``lower-is-better'' evaluation metrics (e.g., Maximum Mean Discrepancy (MMD)) are flipped to their opposite numbers before calculating the Pearson correlation. Positive correlations (grids in red) indicate synergistic relationships between two metrics, and negative correlations (girds in blue) indicate trade-offs. No numbers are shown on the diagonal because here we focus on the relationships between distinct metrics evaluated on the same data; only lower triangular numbers are shown because of symmetry. \textcolor{black}{For fidelity, Maximum Mean Discrepancy (MMD),  Root Mean Squared Percentage Error (RMSPE), Correlation Frobenius Distance (CFD), and the Area Under the Curve (AUC) of discriminative prediction are included. For utility, the performance gap of AUC between Train on Synthetic, Test on Real (TSTR) and Train on Real, Test on Real (TRTR) are included, and similarly for the Train on Synthetic + Real, Test on Real (TSRTR) one. For privacy, the median of minimum Euclidean distance between each real medical record and the synthetic EHR dataset are included for Membership Inference Risk (MIR); the F1 score of predictive performance of 1-Nearest Neighbor (1-NN) are included for Attribute Inference Risk (AIR).}}
    \label{fig:correlation-evaluation}
\end{figure}

\begin{figure}[H]
    \centering
    \includegraphics[width=\linewidth]{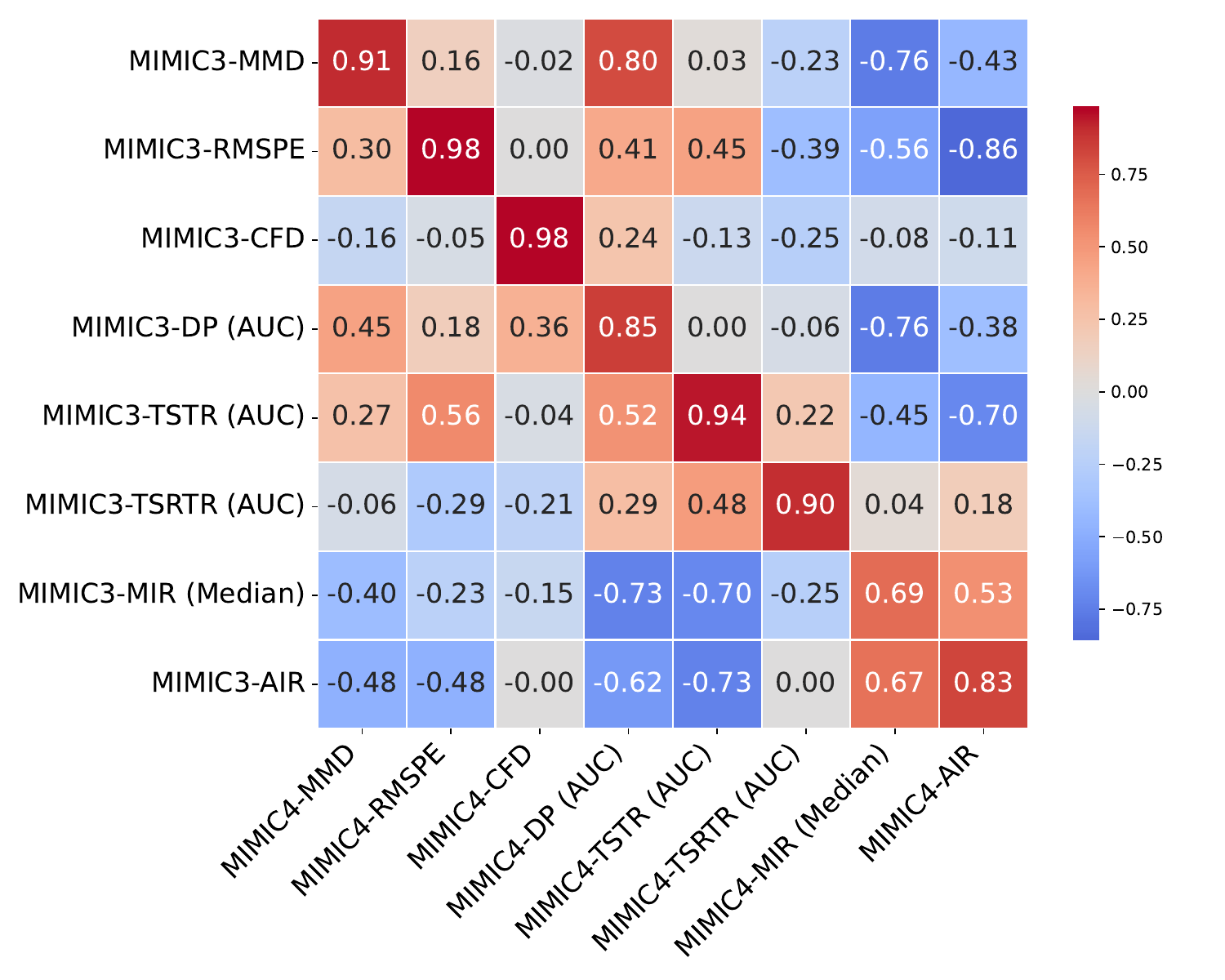}
    \caption{Cross-dataset Pearson correlations of the evaluation metrics (in Table \ref{tab:quantitative}) between MIMIC-III and MIMIC-IV. All ``lower-is-better'' evaluation metrics (e.g., Maximum Mean Discrepancy (MMD)) are flipped to their opposite numbers before calculating the Pearson correlation. Positive correlations (grids in red) indicate synergistic relationships between two metrics, and negative correlations (girds in blue) indicate trade-offs. We need the numbers on the diagonal and the upper triangular matrix because here we focus on the relationships of each metric evaluated on two distinct data sets, MIMIC-III and IV. \textcolor{black}{For fidelity, Maximum Mean Discrepancy (MMD),  Root Mean Squared Percentage Error (RMSPE), Correlation Frobenius Distance (CFD), and the Area Under the Curve (AUC) of discriminative prediction are included. For utility, the performance gap of AUC between Train on Synthetic, Test on Real (TSTR) and Train on Real, Test on Real (TRTR) are included, and similarly for the Train on Synthetic + Real, Test on Real (TSRTR) one. For privacy, the median of minimum Euclidean distance between each real medical record and the synthetic EHR dataset are included for Membership Inference Risk (MIR); the F1 score of predictive performance of 1-Nearest Neighbor (1-NN) are included for Attribute Inference Risk (AIR).}}
    \label{fig:correlation-evaluation-cross}
\end{figure}

\begin{figure}[H]
    \centering
    \includegraphics[width=1.0\linewidth]{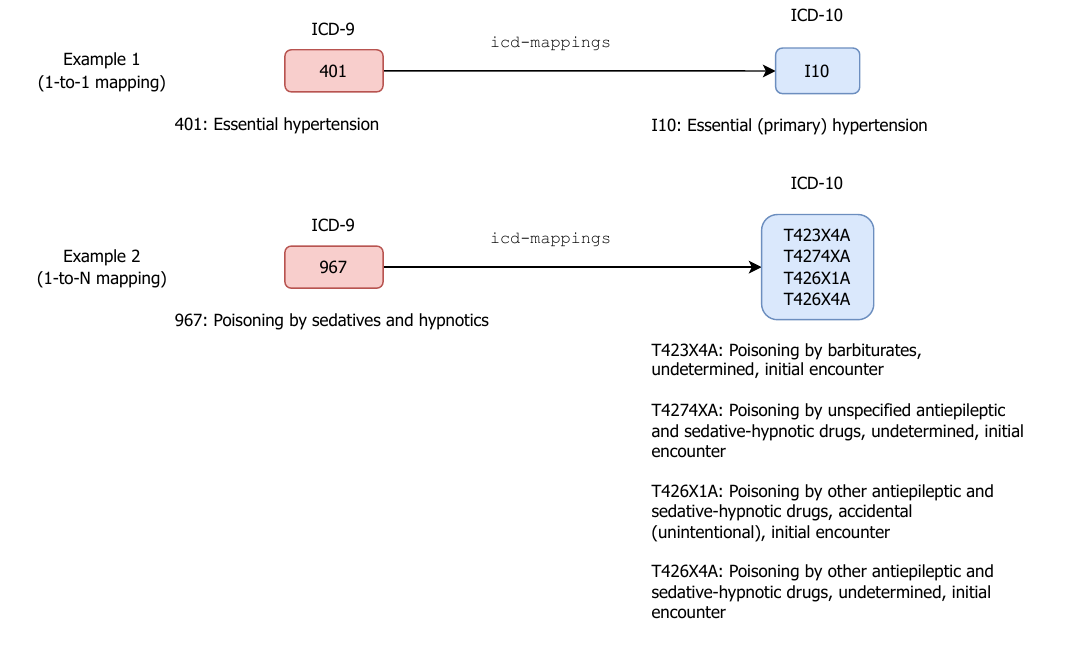}
    \caption{\textcolor{black}{Two examples of the General Equivalence Mappings (GEM) used in the \texttt{icd-mappings} package for mapping from ICD-9 to ICD-10. The two examples, including a 1-to-1 mapping and a 1-to-N mapping, show that the meaning of the codes remains similar and consistent before and after the transformation.}}
    \label{fig:icd9-icd10-example}
\end{figure}

\begin{figure}[H]
    \centering
    \includegraphics[width=0.76\linewidth]{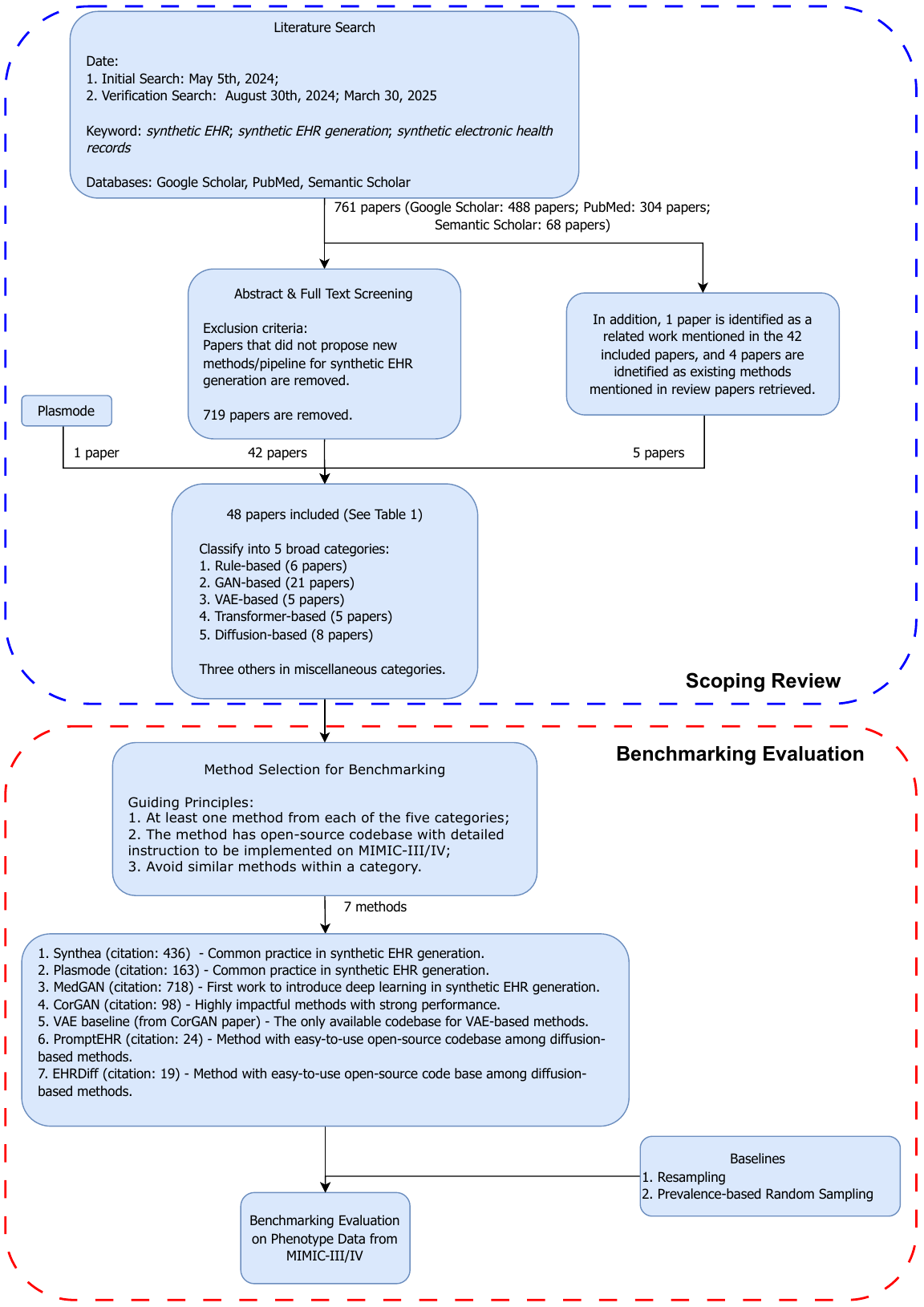}
    \caption{\textcolor{black}{Flow diagram of the scoping review and the benchmarking evaluation process conducted in this study. For the scoping review, 761 papers were retrieved via keyword searches across three databases. Second, screening and citation tracking procedures were performed to identify studies proposing new methodologies for synthetic EHR generation, resulting in 48 papers. The Plasmode framework was additionally included due to its widespread use, despite not appearing in the initial search results. In total, 48 papers were included in the literature review and summarized in Table~\ref{tab:literature}. In terms of the benchmarking evaluation, 7 methods were selected for benchmarking on phenotype data from MIMIC-III/IV based on guiding principles in the box ``Method Selection for Benchmarking''. Two additional baselines were introduced into the benchmarking evaluation.}}   
    \label{fig:review-flowchart}
\end{figure}

\begin{table}[H]
\centering
\caption{\textcolor{black}{A comparison table of the evaluation metrics in the previous benchmarking study\cite{yan2022multifaceted} and our study.}}
\label{tab:metrics-comparison}
\resizebox{0.8\columnwidth}{!}{
\begin{tabular}{l|c|c}
\toprule
\textbf{Evaluation Metrics} & \cite{yan2022multifaceted} & Ours  \\
\midrule
Dimension-wise distributional discrepancy & \cmark & \cmark \\
Bivariate correlation distance & \cmark & \cmark \\
Discriminative prediction & \cmark & \cmark \\
Latent cluster analysis & \cmark &  \\
Clinical knowledge violation & \cmark &  \\
Medical concept abundance & \cmark &  \\
TSTR performance & \cmark & \cmark \\
TSRTR performance & & \cmark \\
TRTS performance & \cmark &  \\
Feature selection & \cmark &  \\
Attribute inference risk & \cmark & \cmark \\
Membership inference risk & \cmark & \cmark \\
Meaningful identity disclosure risk & \cmark &  \\
Nearest neighbor adversarial accuracy risk & \cmark &  \\
\bottomrule
\end{tabular}}
\end{table}

\begin{table}[H]
\centering
\caption{\textcolor{black}{A comparison table of the benchmarked methods included in the previous benchmarking study\cite{yan2022multifaceted} and our study. We do not include DPGAN~\cite{xie2018differentially} in our literature review as it was not a method specifically designed for generating synthetic EHR data.}}
\label{tab:methods-comparison}
\resizebox{0.45\columnwidth}{!}{
\begin{tabular}{l|c|c}
\toprule
\textbf{Methods} & \cite{yan2022multifaceted} & Ours  \\
\midrule
Plasmode~\cite{franklin2014plasmode} & & \cmark \\
MedGAN~\cite{choi2017generating} & \cmark & \cmark \\
medBGAN~\cite{baowaly2019synthesizing} & \cmark & \\
medWGAN~\cite{baowaly2019synthesizing} & \cmark &  \\
EMR-WGAN~\cite{zhang2020ensuring} & \cmark & \\
DPGAN~\cite{xie2018differentially} & \cmark &  \\
Synthea~\cite{walonoski2018synthea} & & \cmark \\
CorGAN~\cite{torfi2020corgan} & & \cmark \\
VAE~\cite{torfi2020corgan} & & \cmark \\
PromptEHR~\cite{wang2022promptehr} & & \cmark \\
EHRDiff~\cite{yuan2023ehrdiff} &  & \cmark \\ \bottomrule
\end{tabular}} 
\end{table}

\begin{table}[H]
\centering
\caption{\textcolor{black}{Spearman correlations between evaluation results computed using the PhecodeX system with only parent phecodes and those based on alternative coding systems of dimension $K$. For fidelity, Maximum Mean Discrepancy (MMD),  Root Mean Squared Percentage Error (RMSPE), Correlation Frobenius Distance (CFD), and the Area Under the Curve (AUC) of discriminative prediction are reported. For utility, the performance gap of AUC between Train on Synthetic, Test on Real (TSTR) and Train on Real, Test on Real (TRTR) are reported, and similarly for the Train on Synthetic + Real, Test on Real (TSRTR) one. For privacy, the median of minimum Euclidean distance between each real medical record and the synthetic EHR dataset are reported for Membership Inference Risk (MIR); the F1 score of predictive performance of 1-Nearest Neighbor (1-NN) are reported for Attribute Inference Risk (AIR). $K$ represents the dimensions of the coding systems.}}
\resizebox{1.0\columnwidth}{!}{
\begin{tabular}{lc|cccc|cc|cc}
\toprule
 & & \multicolumn{4}{c|}{Fidelity} & \multicolumn{2}{c|}{Utility} & \multicolumn{2}{c}{Privacy} \\
Coding System & $K$ & MMD & RMSPE & CFD & AUC & TSTR (AUC) & TSRTR (AUC) & MIR (median) & AIR \\
\midrule
ICD-9     & 1,071 & 0.83 & 0.98 & 0.52 & 0.98 & 0.93 & 0.95 & 0.88 & 0.81 \\
ICD-10    & 5,613 & 0.97 & 0.97 & 0.95 & 0.97 & 0.85 & 0.90 & 0.75 & 0.83 \\
PhecodeX (with child phecodes) & 2,254 & 1.00 & 0.98 & 0.98 & 0.98 & 0.85 & 0.97 & 0.75 & 0.92 \\
\bottomrule
\end{tabular}}
\label{tab:code_system_metrics}
\end{table}

\begin{table}[H]
\centering
\caption{\textcolor{black}{Spearman correlations between the evaluation results computed on the full MIMIC-III cohort and those computed on a subgroup of the cohort. For fidelity, Maximum Mean Discrepancy (MMD),  Root Mean Squared Percentage Error (RMSPE), Correlation Frobenius Distance (CFD), and the Area Under the Curve (AUC) of discriminative prediction are reported. For utility, the performance gap of AUC between Train on Synthetic, Test on Real (TSTR) and Train on Real, Test on Real (TRTR) are reported, and similarly for the Train on Synthetic + Real, Test on Real (TSRTR) one. For privacy, the median of minimum Euclidean distance between each real medical record and the synthetic EHR dataset are reported for Membership Inference Risk (MIR); the F1 score of predictive performance of 1-Nearest Neighbor (1-NN) are reported for Attribute Inference Risk (AIR).}}
\resizebox{1.0\columnwidth}{!}{%
\begin{tabular}{l|cccc|cc|cc}
\toprule
\textbf{Subgroup} & \multicolumn{4}{c|}{Fidelity} & \multicolumn{2}{c|}{Utility} & \multicolumn{2}{c}{Privacy} \\
                  & MMD  & RMSPE & CFD  & AUC  & TSTR (AUC) & TSRTR (AUC) & MIR (median) & AIR \\
\midrule
Older (Age $>$ 50)     & 0.97 & 0.27 & 0.93 & 0.98 & 0.95 & 0.98 & 0.70 & 0.93 \\
Younger (Age $\leq$ 50)     & 0.75 & 0.75 & 1.00 & 0.97 & 0.95 & 0.88 & 0.80 & 0.90 \\ \midrule
Female    & 1.00 & 0.98 & 1.00 & 1.00 & 0.98 & 0.98 & 0.70 & 0.93 \\
Male       & 1.00 & 1.00 & 1.00 & 1.00 & 0.97 & 1.00 & 0.75 & 0.92 \\ \midrule
Non-White & 1.00 & 0.97 & 1.00 & 0.98 & 0.93 & 0.98 & 0.75 & 0.97 \\
White      & 1.00 & 1.00 & 1.00 & 1.00 & 1.00 & 1.00 & 0.70 & 0.93 \\

\bottomrule
\end{tabular}}
\label{tab:subgroup_metrics}
\end{table}

\begin{table}[H]
\centering
\caption{\textcolor{black}{Summary statistics of the number of visits in the synthetic data generated by methods that support longitudinal generation. Synthea exhibits significantly different distribution in the number of visits as it does not rely on training data for generation.}}
\label{tab:longitudinal_generation_statistics}
\resizebox{0.6\columnwidth}{!}{
\begin{tabular}{l|ccc}
\toprule
Method & Mean (Std) & Median & Maximum \\
\midrule
Synthea & 67.6 (118.8) & 32.0 & 1121.0 \\
PromptEHR & 2.5 (1.8) & 2.0 & 20.0 \\ \bottomrule
\end{tabular}} 
\end{table}

\begin{figure}[H]
    \centering
    \includegraphics[width=1.0\columnwidth]{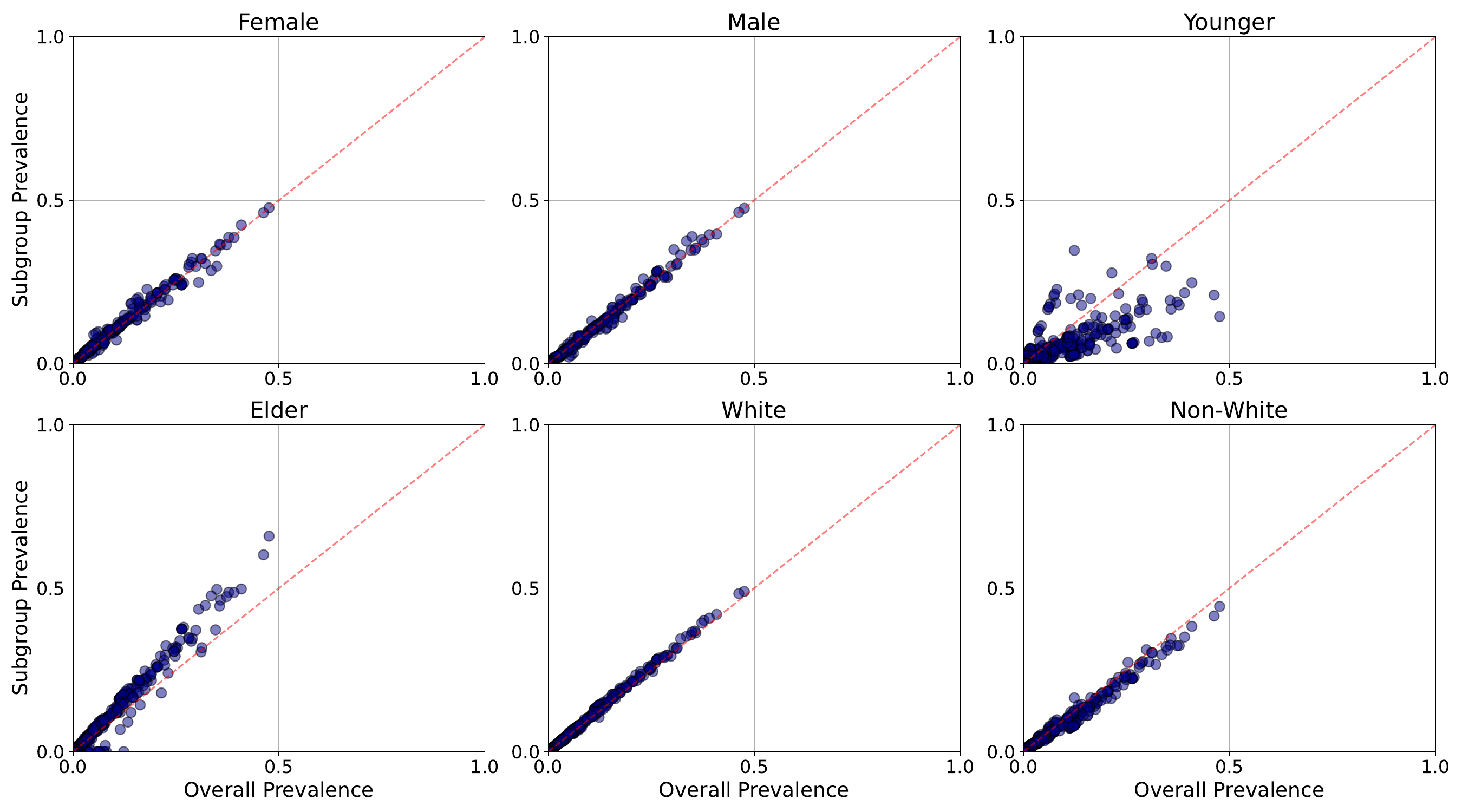}
    \caption{\textcolor{black}{Scatterplots comparing the phecode-wise prevalence between the overall MIMIC-III population and each subpopulation. Red dashed lines indicate equality in phecode prevalence between the subpopulation and the overall population.}}   
    \label{fig:subgroup-prevalence-scatterplots}
\end{figure}

\begin{figure}[H]
    \centering
    \includegraphics[width=1.0\columnwidth]{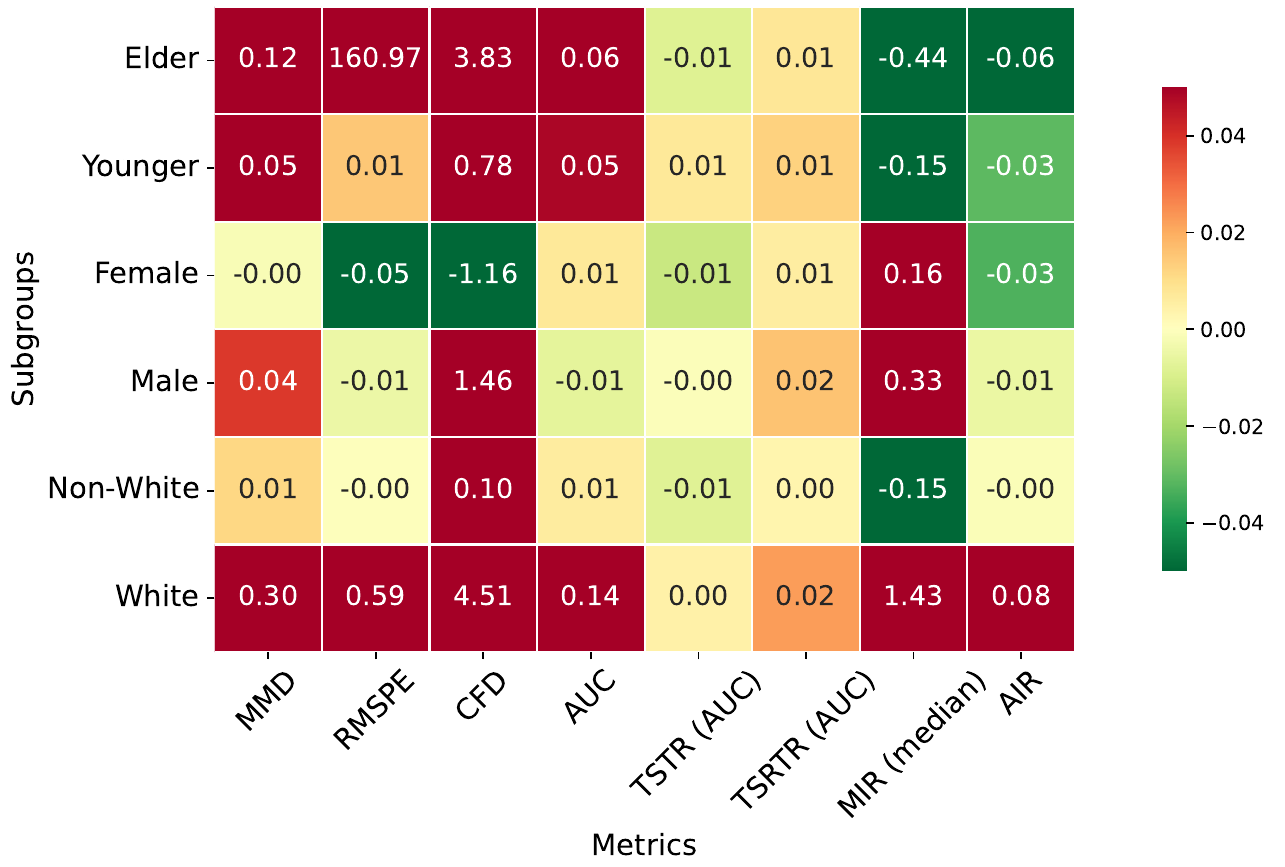}
    \caption{\textcolor{black}{Heatmap of the actual performance differences for CorGAN when switching the focus from the overall population to each subgroup population. The difference of all ``higher-is-better'' evaluation metrics are flipped to their opposite numbers for better presentation. Red cells indicate poorer performance on the subgroup compared to the overall population, green cells indicate better performance, and yellow cells indicate similar performance between the subgroup and overall populations. \textcolor{black}{For fidelity, Maximum Mean Discrepancy (MMD),  Root Mean Squared Percentage Error (RMSPE), Correlation Frobenius Distance (CFD), and the Area Under the Curve (AUC) of discriminative prediction are included. For utility, the performance gap of AUC between Train on Synthetic, Test on Real (TSTR) and Train on Real, Test on Real (TRTR) are included, and similarly for the Train on Synthetic + Real, Test on Real (TSRTR) one. For privacy, the median of minimum Euclidean distance between each real medical record and the synthetic EHR dataset are included for Membership Inference Risk (MIR); the F1 score of predictive performance of 1-Nearest Neighbor (1-NN) are included for Attribute Inference Risk (AIR).}}}   
    \label{fig:subgroup-heatmap}
\end{figure}

\begin{figure}[H]
    \centering
    \includegraphics[width=1.0\columnwidth]{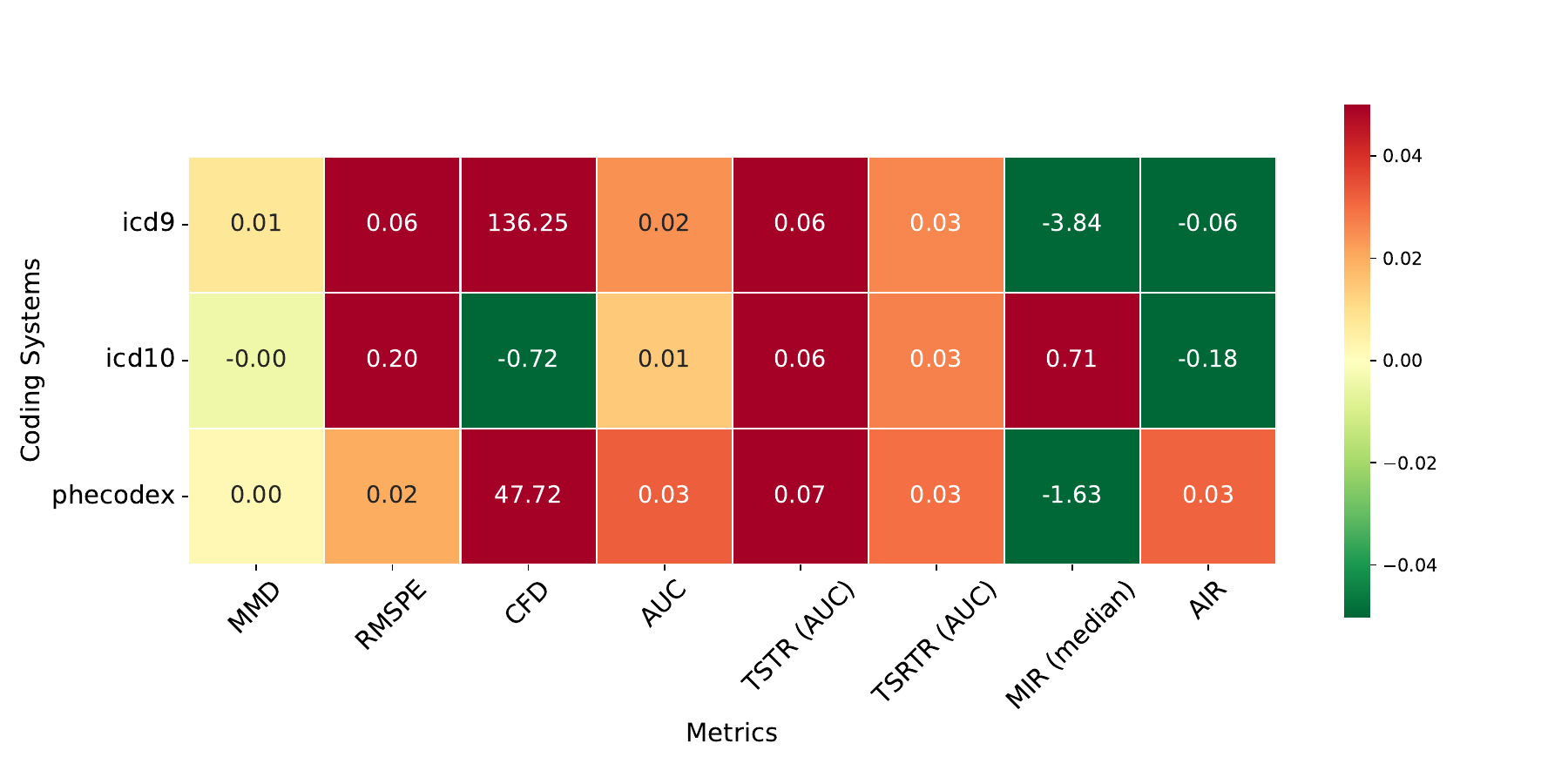}
    \caption{\textcolor{black}{Heatmap of the actual performance differences for CorGAN when switching the coding systems from PhecodeX with parent code only to 1) ICD-9; 2) ICD-10; 3) PhecodeX with child codes. The difference of all ``higher-is-better'' evaluation metrics are flipped to their opposite numbers for better presentation. Red cells indicate poorer performance on the alternative coding system compared to PhecodeX with parent code only, green cells indicate better performance, and yellow cells indicate similar performance between the two coding systems. \textcolor{black}{For fidelity, Maximum Mean Discrepancy (MMD),  Root Mean Squared Percentage Error (RMSPE), Correlation Frobenius Distance (CFD), and the Area Under the Curve (AUC) of discriminative prediction are included. For utility, the performance gap of AUC between Train on Synthetic, Test on Real (TSTR) and Train on Real, Test on Real (TRTR) are included, and similarly for the Train on Synthetic + Real, Test on Real (TSRTR) one. For privacy, the median of minimum Euclidean distance between each real medical record and the synthetic EHR dataset are included for Membership Inference Risk (MIR); the F1 score of predictive performance of 1-Nearest Neighbor (1-NN) are included for Attribute Inference Risk (AIR).}}}   
    \label{fig:coding-systems-heatmap}
\end{figure}

\newpage
\section{Supplementary Methods}
\label{secsupp:method}

\paragraph{\textcolor{black}{Full search strings: }} \textcolor{black}{The full search strings used in the three databases are shown as follows:
\begin{itemize}
    \item Google Scholar: ("synthetic EHR") OR ("synthetic EHR generation") OR ("synthetic electronic health records");
    \item Semantic Scholar: ("synthetic EHR") OR ("synthetic EHR generation") OR ("synthetic electronic health records");
    \item PubMed: ((synthetic EHR) OR (synthetic EHR generation)) OR (synthetic electronic health records)
\end{itemize}}

\paragraph{Details of benchmarked datasets: } In this study, version 1.4 of MIMIC-III and version 2.2 of MIMIC-IV are used for the comparative evaluation and benchmarking. To highlight the difference from ``synthetic data'', we use ``real data'' to refer to either the MIMIC-III or MIMIC-IV data which we will distinguish when necessary; we use ``training data'' to refer to the subset of data for training methods which can be real, synthetic data, or their combination. Note that the scenario where the synthetic data can be treated as training data only appears in the predictive utility evaluation described in Section~\ref{sec:evaluation}. The entire MIMIC-III dataset is used to train generative models if no pre-trained models are available. For a fair evaluation, we adopt PhecodeX 1.0~\cite{10.1093/bioinformatics/btad655} as the standard coding system for evaluation. This is due to the fact that the included methods in the benchmark were originally designed to adopt different \textcolor{black}{coding} systems (e.g., ICD-9, SNOMED-CT). Therefore, it is impossible to conduct a head-to-head comparison among these methods. On the other hand, the coding systems that most existing methods adopt (e.g., ICD-9, ICD-10) sometimes are not practically relevant for downstream analysis\cite{bastarache2021using}. Grouping ICD codes into meaningful phenotypes has become a popular choice for association analysis\cite{alessandrini2010new, rassekh2010reclassification, denny2010phewas}. Therefore, we transform all the \textcolor{black}{coding} systems to PhecodeX 1.0, the latest version of the phecode system that provides more granularity to its predecessor Phecode v1.2, to align with current research practice. However, potential loss may occur during the \textcolor{black}{coding} system transformation process. To mitigate this loss, whenever a one-to-multiple mapping exists we retain all possible transformed codes to ensure that information from the original codes is preserved as best as possible.

Specifically, we map ICD-9 to ICD-10 based on the \texttt{icd-mappings} package in Python\footnote{\href{https://pypi.org/project/icd-mappings/0.1.1/}{https://pypi.org/project/icd-mappings/0.1.1/}}. \texttt{icd-mappings} is a script designed to convert the ICD-9 \textcolor{black}{coding} system to ICD-10 based on the General Equivalence Mappings (GEM)\footnote{\href{cms.gov/Medicare/Coding/ICD10/downloads/ICD10MappingFactSheetIntroduction.pdf}{cms.gov/Medicare/Coding/ICD10/downloads/ICD10MappingFactSheetIntroduction.pdf}}. GEM is a mapping system developed by the Centers for Medicare \& Medicaid Services (CMS) and the Centers for Disease Control and Prevention (CDC) to facilitate conversions between ICD-9-CM and ICD-10-CM/PCS. While previous studies\cite{turer2015icd} have identified subtle inconsistencies between GEM and manual mappings by experts, we believe our evaluation results remain reliable as the mapping was consistently applied across all methods. Two examples in Figure~\ref{fig:icd9-icd10-example} show that the meaning of the ICD-10 codes after transformation remain consistent with their corresponding ICD-9 codes. We map Systematized Medical Nomenclature for Medicine–Clinical Terminology (SNOMED-CT) to ICD-10 via the SNOMED mapping tool\footnote{\href{https://prod-mapping.ihtsdotools.org/\#/project/records}{https://prod-mapping.ihtsdotools.org/\#/project/records}}; we then convert ICD-10 coding system to PhecodeX according to the Phecode Map X (Extended)\footnote{\href{https://phewascatalog.org/phecode\_x}{https://phewascatalog.org/phecode\_x}}. Based on MIMIC-III, we identify $K = 2,254$ distinct phecodes. For visualization, we keep only phecodes with more than 50 patients in MIMIC-III, resulting in $K = 1,773$ phecodes. For quantitative evaluation, on the other hand, all additional granularity (i.e., digits after the decimal point) in phecodes were omitted and merged into their corresponding parent categories, resulting in $K = 595$ parent phecodes. 

\paragraph{Goal of synthetic EHR data generation: } The goal of synthetic EHR generation is to produce a synthetic dataset of size $M$, $\mathcal{D}_{syn} = \{\mathbf{x}_{i, syn}\}^M_{i=1}$, where each sample in $\mathcal{D}_{syn}$ can be viewed as samples drawn from the same distribution as those in the real dataset of size $N$ denoted as $\mathcal{D}_{real} = \{\mathbf{x}_{i, real}\}^N_{i=1}$, while ensuring privacy protection for patients in $\mathcal{D}_{real}$.

\paragraph{Details of synthetic EHR data generation: } For MedGAN, CorGAN, VAE, and EHRDiff, we train the models from scratch using the publicly released code by their authors with the entire MIMIC-III dataset, and then generate the synthetic data. For PromptEHR, we use the pretrained checkpoint released by the authors to generate synthetic data. For Synthea, we generate synthetic data based on the default setting. In the case of Plasmode, which generates only one outcome variable at a time, the synthetic EHR data are generated dimension by dimension. For each dimension, the presence of this disease is treated as the outcome variable, and the time variable is defined as either the time to the first diagnosis of the disease or the time to the patient’s last ICU visit, depending on the presence of disease. The treatment variable is randomly sampled from a Bernoulli distribution with probability 0.5. Key socio-demographic characteristics, including age, gender, and ethnicity, are treated as covariates to fit the Cox Proportional Hazards model. For this study, we fix $M = 50,000$ except that we vary the size of generated samples to investigate their impacts on the quality of synthetic data in the numerical studies described in the following part.

\paragraph{Supplementary details of fidelity metrics: } Except for quantitative evaluation, we present visualizations to compare the distribution characteristics between the real and the synthetic EHR data. We use a boxplot to show the dimension-wise prevalence using each method, and another boxplot to show the distribution of the number of unique phecodes per patient in the data generated by each method. A scatterplot is included to compare the dimension-wise prevalence between real and synthetic EHR data, with each dot representing a phecode. For pair-wise correlation, we visualize the element-wise difference of correlation between the real and the synthetic data via boxplots; 100,000 pairs of entries from the upper triangle of the two correlation matrices are sampled and shown in the boxplots.

In addition, considering that each disease dimension consists of binary data indicating occurrence, we present another metric based on  the Frobenius distance between co-occurrence matrices for the real and the synthetic datasets. The \textbf{co-occurrence matrices Frobenius distance} (COFD) is calculated as follows:

$$\mathbf{B}_{real} = \mathbf{X}^\top_{real}\mathbf{X}_{real}, \quad \mathbf{B}_{syn} = \mathbf{X}^\top_{syn}\mathbf{X}_{syn},$$
$$\texttt{COFD} = \left \lVert \mathbf{B}_{real} - \mathbf{B}_{syn} \right \rVert_F = \sqrt{\sum^K_{k=1} \sum^K_{k'=1} (b^{real}_{kk'} - b^{syn}_{kk'})^2},$$

where $b^{real}_{kk'}$  indicates the number of patients having both the $k$-th and $k'$-th phenotypes in the real data; similarly, $b^{syn}_{kk'}$ for the synthetic data. All the COFD values are divided by a factor of 1000 for ease of presentation.

\textcolor{black}{Note that existing work \cite{zhang2022keeping} has shown that training the discriminators by fine-tuning the backbone components of the pre-trained models can result in better performance compared to training the discriminator from scratch. However, as our study includes methods with diverse architectures and modeling strategies, adopting the fine-tuning approach would compromise the fairness of comparisons. Further, most of the methods included in our benchmark do not have a discriminator by nature, unlike \cite{zhang2022keeping}, which adopts a GAN-based architecture and therefore inherently contains a discriminator. Additionally, rule-based methods lack trainable parameters, making fine-tuning infeasible. As such, applying fine-tuning consistently across all models would require designing and implementing custom discriminative components for each method. This would not only be beyond the scope of this benchmarking study, but also significantly compromise the generalizability of the benchmarking with future methods. Therefore, we apply the basic logistic regression models as the backbone of all the discriminators and train from scratch to ensure a fair and comparable evaluation across all methods. Fine-tuning will likely change the comparison of methods on both the absolute and relative scale and merits in-depth future exploration.}

\paragraph{Supplementary details of utility metrics: } For predictive utility, we present the details of three scenarios as follows:

\begin{enumerate}
    \item \textbf{``Train on Synthetic, Test on Real'' (TSTR).} To mimic common scenarios in practice where synthetic data are used, we train an ML model using the synthetic data and test the method on the real data. This setting evaluates whether the synthetic data can substitute the real data in downstream predictive tasks; 
    \item \textbf{``Train on Synthetic + Real, Test on Real'' (TSRTR).} This setting investigates if the synthetic data can be considered as a data augmentation of the real data in downstream applications which have been shown to be capable of boosting performance by preventing overfitting issues when only a limited amount of real data is available~\cite{zheng2023toward}. We stack the training split in the real data and the synthetic data together to train the ML model and test it on the test set of the real data. In practice, this is only feasible when the investigator has access to the real data. Nonetheless, it is included here for showing whether synthetic data has deleterious or beneficial effects when added to the real data;
    \item \textbf{``Train on Real and Test on Real''(TRTR).} This scenario does not use synthetic data. The ML model is trained on the training data split, and finds the test error on the testing data split. This provides the baseline for predictive utility evaluation.
\end{enumerate}

In this study, analytical utility evaluations are conducted under two tasks. In the first task, whether a patient gets any type of cancer (phecodeX starting with ``\texttt{CA}'') is treated as the outcome, and Obesity (\texttt{EM\_236}) is treated as the predictor; for the second task, Diabetes (\texttt{EM\_202}) is treated as the outcome, and Hypertension (\texttt{CV\_401}) is treated as the predictor. See Table \ref{tab:summary} for prevalence of any type of cancer, obesity, hypertension, and diabetes. For predictive utility evaluation, we split the real dataset into training and testing subsets with a 4:1 ratio. Hypertension (\texttt{CV\_401}) is treated as the outcome, because it is the most prevalent phenotype in both MIMIC-III and MIMIC-IV cohorts; the remaining phenotypes are treated as predictors. The logistic regression model is treated as our ML models, with results of more advanced ML methods, including Random Forest~\cite{ho1995random}, Gradient Boosting~\cite{friedman2001greedy}, XGBoost~\cite{chen2016xgboost}, LightGBM~\cite{ke2017lightgbm}, and K-Nearest Neighbors (K-NN), reported in Figure~\ref{fig:advanced-ml-tstr} and Figure~\ref{fig:advanced-ml-tsrtr}.

\textcolor{black}{To investigate whether the prevalence of phecode impacts the TSTR performance, we randomly selected 20 phecodes with prevalence greater than 0.1 to repeat the TSTR utility evaluation. The evaluation metrics are then paired with the corresponding phecode prevalence to depict a scatterplot and compute the Spearman correlation for each benchmarked method. Additionally, we present the TSTR performance with additional concrete examples, by changing the outcome variables to Diabetes, Anxiety, Depression, and Obesity, respectively. These examples have prevalence between 5.4\% and 25.4\% in MIMIC-III, and between 12.6\% and 21.7\% in MIMIC-IV.}

\paragraph{Supplementary details of privacy metrics: } The reason that the membership inference risk is important is because that a subject's membership to the training dataset may be sensitive. Specifically, the patient may wish to keep this information private from potential attackers. For example, the training data may be based on all the HIV-positive patients treated in a regional facility. Therefore, it is important to evaluate the risk of membership inference attacks. The latest work of membership inference attack using synthetic data proposed to conduct a privacy attack via an attacker model that can detect local overfitting which occurs if the synthetic data density evaluated at the real data record has a high value; this typically occurs when the real data record is close to the bulk of synthetic data \cite{pmlr-v206-breugel23a}. Following their work, we define the individual risk of membership inference attack as their distance to the closest synthetic data record in the public synthetic dataset. ~\cite{yan2022multifaceted, yuan2023ehrdiff} are existing works that also apply similar metrics in measuring the membership inference risk. However, unlike their approaches, we do not further specify a threshold to classify into successful or failed attacks and calculate the corresponding F1 score to evaluate the membership inference risk of the synthetic data, because the evaluation results may be sensitive to the subjective threshold value.

Note that the scenario differs from the discriminative prediction in the fidelity evaluation in that 1) unlike the use of both the real and the synthetic data, membership inference uses only the synthetic data and a real medical record of the subject being attacked; 2) unlike the binary label of real or synthetic data, membership inference concerns the binary label of whether the subject being attacked is a member in the real data curated by the facility; and 3) unlike the AUC and ACC metric for discriminative prediction, membership inference risk here is evaluated by how close is the real medical record of the attacked individual to the synthetic data.

When $\mathbf{x}_{i, real}$ belongs to MIMIC-III, $d_i$ measures the risk of identifying the membership of subject $i$ in MIMIC-III based on the synthetic data.  When $\mathbf{x}_{i, real}$ belongs to MIMIC-IV, $d_i$  measures the risk of identifying whether subject $i$ is a member of MIMIC-III which remains meaningful because MIMIC-III and MIMIC-IV datasets overlapped in the periods of data collection (2008-2012), hence containing overlapping subjects. The $\texttt{MIR}_{mean}$ and $\texttt{MIR}_{median}$ then provide an average characterization of this distance assuming all the  patients in MIMIC-IV may be subject to membership inference attacks. 

\textcolor{black}{In addition to the $\texttt{MIR}_{mean}$ and $\texttt{MIR}_{median}$, we provide a histogram array to visualize the distribution of $d_i$ across different methods. We also provide Cumulative Distribution Function (CDF) plots for $d_i$ to make a direct comparison among methods. Furthermore, we compute the proportion of real samples that are perfectly matched with synthetic samples in each method to offer a more nuanced numerical evaluation of the membership inference analysis. See Section~\ref{secsupp:results} for the description of the evaluation results.}

On the other hand, to evaluate AIR, we select the top 10 balanced (prevalence closest to 0.5) and top 10 imbalanced diseases as unknown attributes simultaneously to be inferred. The remaining attributes, on the other hand, will be assumed as known features by the attackers.

\paragraph{Supplementary details of computational cost evaluation: } Beyond assessing the quality of the synthetic data using fidelity, downstream utility, and privacy evaluation metrics, the computational cost of generating synthetic EHR datasets is also an important factor during methods selection. This is especially true if a large amount of synthetic data is needed for evaluating the frequentist performance of analytic and prediction methods. To this end, we evaluate the time needed for all the selected methods in synthesizing 100 EHR samples. The evaluations are conducted on a multi-node cluster equipped with Intel(R) Xeon(R) Gold 6148 CPUs with a total of 40 physical cores. A 16GB Tesla V100 GPU is used for running deep generative models when necessary.

\paragraph{Supplementary details of numerical studies: } We explore the effect by varying the size of training ($N$) and synthetic ($M$) data  on the quality of synthetic EHR data. We address two questions. First, given a fixed amount of training data, how many synthetic samples can a generative model produce before the  benefits of adding one additional synthetic sample diminish? Although deep generative models such as GANs, VAEs, and diffusion models can theoretically generate infinite samples, the amount of novel information does not increase proportionally with the size of the synthetic dataset, as it is constrained by the finite training data. Second, how does the size of the training data impact the quality of the generated synthetic data, given a fixed number of synthetic samples? 

To explore these questions, we select the overall best-performing methods from our benchmarking study. For the first question, we train the selected method and generate $M=1K, 2K, 5K, 10K, 20K, 50K, 100K, 200K, 500K$ synthetic samples. We evaluate the generated data using three metrics: MMD for fidelity, TSTR for utility (AUC is reported), and AIR for privacy. For the second question, we train the selected method with $N =1K, 2K, 5K, 10K, 15K, 20K, 25K, 30K, 35K, 40K$ real samples and generate $50K$ synthetic samples; we evaluate them with the same metrics. The results are visualized to illustrate how these metrics evolve under different values of $M$ and $N$.

\section{Supplementary Results}
\label{secsupp:results}

\paragraph{Key functionalities of synthetic EHR generation methods: } Based on this literature, we identify three key functionalities that are critical for the design of synthetic data generation algorithms:

\begin{itemize}
    \item \textbf{Generation conditional on specific variables.}  Supporting synthetic conditional EHR generation, which produces data based on specific contexts, is critical for downstream applications that require subgroup analyses or the inclusion of socio-demographic factors. Our literature review reveals that \textcolor{black}{30} of the \textcolor{black}{48} studies cover conditional generation. For example, ~\cite{esteban2017real} generates synthetic data based on specific labels to augment downstream predictive applications; ~\cite{zhang2020ensuring} generates data conditioning on specific socio-demographic concepts to overcome small data size challenges in training; ~\cite{biswal2021eva} generates synthetic data given specific medical conditions of interest for further scientific investigation.
    \item \textbf{Generation of repeated visits over time.}  As EHR datasets include patients with multiple visits in nature, it is important to capture the longitudinal patterns to generate realistic synthetic timed-stamped data. Of the \textcolor{black}{48} reviewed studies, \textcolor{black}{28} incorporate temporal information for longitudinal generation. For example, Synthea~\cite{walonoski2018synthea} generates synthetic EHR longitudinally by generating synthetic patients and simulating their encounters across their lifespans; TimeDiff~\cite{sun2021generating} and PromptEHR~\cite{wang2022promptehr}, on the other hand, applied neural networks with architectures designed to model sequential data to simulate longitudinal EHR.
    \item \textbf{Generation of multiple modalities of data.} EHR datasets are inherently multimodal, containing various types of medical encounters such as diagnoses, medications, procedures, chart events, clinical notes, and laboratory results. Understanding which modalities each method can generate is crucial for effective data generation. In this study, we focus on events that could be characterized by coding systems (i.e., diagnoses, medications, procedures) and summarize if existing methods could model those events. As shown in Table~\ref{tab:literature}, \textcolor{black}{18} of the \textcolor{black}{23} existing studies that evaluate their methods on MIMIC-III primarily focus on diagnostic events; \cite{baowaly2019synthesizing, torfi2020corgan, sun2024collaborative, chen2024guided, vardhan2024large} additionally include medication or procedures in their modeling; \cite{choi2017generating, wang2022promptehr, theodorou2023synthesize, zhong2024synthesizing} consider all the three types of events in their modeling. 
\end{itemize}

\paragraph{Reason of the inclusion of \textit{Resample}: } While the \textit{Resample} baseline is impractical for synthetic data generation when privacy concern exists, it serves as a valuable reference by providing an upper bound of the fidelity and utility capability of the existing synthetic EHR generation methods, as it closely resembles the real data.

\textcolor{black}{\paragraph{Summary statistics of the longitudinal synthetic data:} Table~\ref{tab:longitudinal_generation_statistics} shows the summary statistics corresponding to the number of visits in the synthetic data generated by PromptEHR and Synthea. The average number of visits is 2.5 for PromptEHR and 67.6 for Synthea. The large distributional discrepancy in the number of visits between the two methods is due to the fact that the synthetic data generated from Synthea does not rely on the real training data in the MIMIC-III dataset. In other words, it is not reflective of an ICU-based EHR dataset. As shown in Table~\ref{tab:quantitative} and the decision tree in Figure~\ref{fig:decision-tree}, both PromptEHR and Synthea underperform compared to cross-sectional generation methods such as CorGAN and MedGAN. Between the two longitudinal approaches, PromptEHR demonstrates superior performance in most fidelity and utility metrics, while Synthea achieves relatively better results in privacy preservation.}

\paragraph{Implementation details of the baselines: } To account for possible loss of information during code transformation (Section \ref{sec:evaluation}), we compute phenotype prevalence based on the MIMIC-III data in ICD-9 system for \textit{PBR} to generate synthetic data. Similarly, we bootstrap real data under the same coding system for \textit{Resample}.

\paragraph{Summary statistics of the benchmarking datasets: } Table~\ref{tab:summary} presents the summary statistics for the MIMIC-III and MIMIC-IV datasets. Overall, the socio-demographic distributions are largely similar between MIMIC-III and MIMIC-IV, except that MIMIC-III additionally includes 7,874 neonates in the dataset. More than half of the patients in MIMIC-III have been diagnosed with cancer (52.9\%), higher than the prevalence in MIMIC-IV (40.2\%). The prevalence of mental health conditions, such as anxiety and depression, is higher in MIMIC-IV (13.0\% and 20.8\%, respectively) compared to MIMIC-III (11.0\% and 16.0\%, respectively). Whereas diabetes and hypertension are more common in the MIMIC-III cohort (25.4\% and 47.6\%, respectively) compared to the MIMIC-IV cohort (21.7\% and 46.0\%, respectively). The prevalence of obesity, on the other hand, is higher in the MIMIC-IV cohort (12.6\%) than in the MIMIC-III cohort (5.4\%). Regarding EHR characteristics, both MIMIC-III and MIMIC-IV exhibit right-skewed distributions for the number of encounters per person, unique phecodes per person, and length of follow-up. The median numbers of encounters per person for both datasets are 1.0, indicating that over half of patients in both data visited ICU or ED (in MIMIC-IV) only once during the data collection period. The mean number of encounters is higher in MIMIC-IV (2.4 visits) than that in MIMIC-III (1.3 visits). In terms of the length of follow-up, significant zero-inflation issues are observed for both datasets, with a median of 0 year of follow-up. This is consistent with our previous findings that half of patients have only one hospital admission during data collection period. The average length of follow-up, on the other hand, is higher in MIMIC-IV (1.2 years) than that in MIMIC-III (0.3 years). The average number of unique phecodes per patient is greater in MIMIC-III (31.3 phecodes) than in MIMIC-IV (26.0 phecodes), although MIMIC-IV has a higher maximum number of unique phecodes for a single patient (260 phecodes) than MIMIC-III (203 phecodes). The summary statistics results reveal a significant distributional discrepancy between MIMIC-III and MIMIC-IV, making them well-suited for evaluating the transportability of the synthetic EHR methods.
\paragraph{Supplementary evaluation results of fidelity: } Figure~\ref{fig:real-versus-synthetic-prevalence} compares the phecode-wise prevalence between the synthetic data and the real data (MIMIC-III). \textit{PBR} achieves an almost perfect match with the real data but tends to overestimate the prevalence for several non-rare phecodes, which may be due to code transformation. Among the selected methods, MedGAN and CorGAN exhibit less divergence from the real data, while Plasmode, PromptEHR, VAE, and EHRDiff show greater discrepancies. Notably, Synthea generates non-zero prevalence for only 160 phecodes, likely due to its limited support for only around 90 disease modules, and potential information losses during multi-step code transformations from the SNOMED-CT~\cite{donnelly2006snomed} to the PhecodeX coding system.
\textcolor{black}{\paragraph{Supplementary evaluation results of utility: } Based on Figure~\ref{fig:tstr-code-frequency-a} (MIMIC-III), we observe that the Spearman correlation between AUC$_{TSTR}$ and phecode prevalence for top 2 non-baseline methods (in terms of predictive utility measured by AUC$_{TSTR}$, Table 2), MedGAN and CorGAN, are -0.38 and -0.3, respectively. The values are comparable to a Spearman correlation of -0.34 between AUC$_{TSTR}$ and prevalence calculated based on the real data. The two worst methods, Plasmode and EHRDiff, have Spearman correlations (with phecode prevalence) of -0.11 and -0.33, respectively. On MIMIC-IV (Figure~\ref{fig:tstr-code-frequency-b}), we observe that the Spearman correlation between AUC$_{TSTR}$ and phecode prevalence for top 2 non-baseline methods, MedGAN and PromptEHR, are -0.18 and -0.05, respectively.  In contrast, Plasmode and EHRDiff have Spearman correlations -0.11 and 0.03, respectively. Furthermore, it is observed that in both datasets, most benchmarked methods exhibit similar bivariate patterns between TSTR performance and phecode frequency when compared to the bivariate pattern between TRTR performance and phecode frequency obtained from the real data, with the exceptions of VAE, Synthea, and Plasmode. Notably, the outlier phecodes identified in the TRTR evaluation persist in the TSTR evaluation across both datasets. Similar findings in the relationship between TSTR/TRTR performance and code frequency are also observed in Table~\ref{tab:disease_comparison}.}

\textcolor{black}{\paragraph{Supplementary evaluation results of privacy: } Figure~\ref{fig:mir-hist} shows that most of the distributions of $d_i$ exhibit right-skewness with few large values. Compared to MIMIC-IV, the minimum distances from the MIMIC-III dataset tend to cluster more closely around the sample median, exhibiting reduced variability. Figure~\ref{fig:mir-cdf} reveals that the \textit{Resample} baseline, CorGAN, and MedGAN consistently exhibit higher cumulative probability values at a given distance cutoff compared to other methods, indicating the higher membership inference risk within the synthetic data generated by these methods. On the other hand, the rule-based methods achieve the lowest probability values. Both findings are consistent with those in the quantitative evaluation. Regarding the proportion of exact matches, Table~\ref{tab:exact_match} shows that MedGAN and CorGAN result in the highest proportion of exact matches compared to other benchmarked methods. On the other hand, methods that do not rely on training data (i.e., the prevalence-based random baseline, Synthea) result in the lowest proportion of exact matches. In addition, it can be observed that most of the methods show a decreased proportion of exact matches when evaluated on MIMIC-IV compared to MIMIC-III.}

\paragraph{Supplementary evaluation results of computational cost: } The results of computational costs are presented in Table~\ref{tab:computational-cost}. Overall, rule-based methods have higher computational costs compared to most of the deep generative methods in other categories. Specifically, Synthea and Plasmode cost 24 and 12,292 seconds to generate 100 samples, respectively\footnote{Although Plasmode is extremely expensive in computation under this setting, the cost increases only sub-linearly with sample sizes. For example, generating 50,000 synthetic samples with Plasmode costs 17,225 seconds.}. On the other hand, among deep generative methods, GAN-based and VAE-based methods are significantly faster than Transformer-based and Diffusion-based approaches. Specifically, CorGAN and MedGAN generate 100 samples in 0.26 and 0.08 seconds, respectively, while VAE requires 0.16 seconds. In contrast, EHRDiff takes 0.65 seconds to generate the same number of samples, and PromptEHR achieves the highest computational cost among deep generative methods at 77 seconds per 100 samples, which is primarily because of the longitudinal nature of its EHR generation process.

\paragraph{Supplementary results of the numerical studies: } Based on the evaluation results shown in Figure~\ref{fig:prevalence-boxplot},~\ref{fig:num_of_phecode-boxplot},~\ref{fig:mimic3-corr-diff},~\ref{fig:mimic4-corr-diff} and Table~\ref{tab:quantitative} and~\ref{tab:analytical-utility}, CorGAN demonstrates outstanding performance across all three key perspectives of synthetic data quality. Hence, CorGAN is selected for the numerical studies to explore the effect of varying sizes of $M$ and $N$ on the quality of the synthetic data.

\begin{itemize}
    \item \textbf{Effect of varying synthetic data size $M$: } Figure~\ref{fig:gen_sample_size} presents the results of varying the synthetic sample sizes ($M$) on the performance of synthetic EHR data. Fidelity improves (the metric MMD declines) consistently as the number of generated samples increases, which is expected since sufficiently large sample sizes are required to approximate the real data distribution. When the number of synthetic samples exceeds 10,000, the MMD score begins to fluctuate and stabilizes between 0.048-0.049. In terms of utility, increasing the synthetic sample size narrows the performance gap between synthetic and real data, although the rate of improvement is non-linear. For privacy, larger synthetic sample sizes increase the risk of the attribute inference attack, indicating that generating more synthetic EHR data increases the risk of information leakage from the training dataset.
    \item \textbf{Effect of varying training data size $N$: } Figure~\ref{fig:train_size} illustrates the results of varying training data sizes on synthetic EHR data performance. As the training data size increases, both the fidelity and utility of the generated synthetic data improve. However, this improvement comes with an increased risk of privacy exposure, indicating a trade-off between synthetic data quality and privacy protection.
\end{itemize}

\textcolor{black}{\paragraph{Sensitivity analysis of evaluation on different coding systems: } To further explore how relative performance and ranking of the methods vary with different coding systems when compared to PhecodeX parent codes, we have repeated the benchmarking evaluation on MIMIC-III using the same metrics in Table~\ref{tab:quantitative}, but using the following coding systems: 1) the PhecodeX coding system with child codes; 2) the ICD-10 coding system with only the parent codes; 3) the ICD-9 coding system with only the parent codes. We compute Spearman correlations between the ranks of methods (for a given evaluation metric)  obtained from PhecodeX and each of the three alternative coding systems. Note that in the predictive utility evaluation, hypertension is retained as the outcome variable. Specifically, the target outcome is defined using the following codes: \texttt{401} in ICD-9, \texttt{I10} in ICD-10, and \texttt{CV\_401.1} in PhecodeX (with child phecodes). In addition, we compute the actual performance differences for CorGAN, the best-performing method in terms of the fidelity and analytical utility, when we switch the coding system from the three coding systems above to PhecodeX (with parent code only). We depict a heatmap to visually display performance across different coding systems.}

\textcolor{black}{The results from Table~\ref{tab:code_system_metrics} indicate strong correlations across all coding systems for most evaluation metrics, which indicates that the choice of the coding system is likely to lead to minor changes in the evaluation rankings. Note that moderate correlations are observed between the ICD-9 coding system and the PhecodeX (with only parent phecodes) in terms of the performance in the Correlation Frobenius Distance (CFD). We hypothesize that this may be due to the change of granularity levels during the transfer mapping between coding systems.}

\textcolor{black}{In terms of the actual performance differences, we picked one method CorGAN which performs well in our initial simulation and studied its performance across different coding systems. From Figure~\ref{fig:coding-systems-heatmap}, all alternative coding systems result in a worse fidelity and utility performance, except CFD when  ICD-10 is used. Alternative coding systems exhibit a better performance in terms of the privacy metrics, except MIR (median) from ICD-10 and AIR from PhecodeX with child code. However, as the substantive meaning of the evaluation metrics changed, we are unable to conclude if PhecodeX (with only parent code) coding system facilitates better quality of the synthetic data.}

\paragraph{\textcolor{black}{Sub-experiment results of evaluation in different subpopulations: }}\textcolor{black}{To explore the performance of the proposed methods on subpopulations, we conduct additional sub-experiments on evaluating the generated synthetic data across six subgroups from the MIMIC-III dataset: 1) younger population (age $\leq$ 50); 2) older population (age $>$ 50); 3) male patients; 4) female patients; 5) white patients; 6) non-white patients. Similar to what we have done for the coding systems, we compute Spearman correlations between the evaluation results for these subgroups with those reported in Table~\ref{tab:quantitative} Furthermore, we compute the actual performance differences for CorGAN, the best-performing method in terms of the fidelity and analytical utility, when evaluated on each subgroup population versus the overall population and create a heatmap to visually display performance across subgroups. The results are presented in Table~\ref{tab:subgroup_metrics} and Figure~\ref{fig:subgroup-heatmap}, respectively. We also depict scatterplots of the phecode-wise prevalence between the overall MIMIC-III population and each subpopulation in Figure~\ref{fig:subgroup-prevalence-scatterplots}.}

\textcolor{black}{The results from Table~\ref{tab:subgroup_metrics} exhibit strong correlations across all subgroups for most evaluation metrics. However, the correlation between the overall population and the older subgroup on the RMSPE metric is 0.27. According to the definition of the RMSPE metric, RMSPE places greater weight on rare phecodes. Furthermore, Figure~\ref{fig:subgroup-prevalence-scatterplots} shows that several phecodes have near-zero prevalence in the older subpopulation but are more prevalent in the overall population. This suggests that variations in disease prevalence across subpopulations may impact the quality of synthetic data in accurately representing those subgroups. This experiment suggests that the relative performance among the competing methods trained and evaluated on the overall population may not always carry over to some subpopulations. For example, the RMSPEs of all the methods do not correlate well when switching from the overall to the older subpopulation (Spearman correlation of 0.27). In our evaluation, this is caused by some codes being much rarer in the older subpopulation (e.g., \texttt{NB\_875} has prevalence $<0.1\%$ in older population) versus those in the overall population (\texttt{NB\_875} has prevalence 12.3\% in overall population), resulting in less reliable RMSPE values that do not preserve the ranking of the methods. Regarding the actual performance differences between each subgroup and the overall population in the MIMIC-III cohort, Figure~\ref{fig:subgroup-heatmap} illustrates that CorGAN generally exhibits lower fidelity performance across most subgroups relative to the overall population. In contrast, CorGAN demonstrates better privacy performance in both the younger and older populations, suggesting a trade-off between fidelity and privacy. Notably, predictive utility performance remains relatively stable across all subgroups when compared to the overall population. However, in the white subgroup, CorGAN shows decreased performance in both fidelity and privacy metrics.}

\section{\textcolor{black}{Exploration of Trade-Offs in Evaluation Metrics and Transportability across Datasets}}
\label{secsupp:tradeoff}

The evaluation results reveal a trade-off between fidelity, utility, and privacy protection in synthetic EHR data generation. To further explore these relationships, we present a heatmap of the correlations among the evaluation metrics for both datasets in Figure~\ref{fig:correlation-evaluation}. For consistency, all ``lower-is-better" metrics are inverted so that positive correlations indicate synergistic relationships and negative correlations indicate trade-offs. 

\textcolor{black}{As one may expect, in MIMIC-III, we observe negative associations in all the utility-privacy metric pairs, as well as the fidelity-privacy pairs, suggesting that improved utility and fidelity may come at the expense of reduced privacy.} When evaluated on MIMIC-IV, negative associations are observed between the privacy and almost all the non-privacy evaluation metrics, which are consistent with the findings on MIMIC-III. These findings suggest that the trade-offs observed above are common despite distributional differences between the MIMIC-III and the MIMIC-IV data.

In addition, we also explore the relationships between the evaluation metrics across the two datasets. As shown in Figure~\ref{fig:correlation-evaluation-cross}, for each metric, there are strong positive correlations between MIMIC-III and MIMIC-IV, suggesting that the transportability of the existing methods is closely related to their performance in the \textcolor{black}{data} they are trained on. In addition, negative associations are observed between privacy and all the remaining non-privacy metrics, showing that privacy-related trade-offs remain across datasets.

\section{\textcolor{black}{Supplementary Discussion}}
\label{secsupp:discussion}

\textcolor{black}{We discuss several other open opportunities in the area. First, for downstream analytic tasks on subpopulations defined by specific values of socio-demographic variables, it is unclear whether training the synthetic EHR generation methods on the entire population is the most effective strategy and needs further studies. For example, MIMIC-III contains a significant number of neonatal patients with distinct distributional characteristics than the adult patients. Scientific studies often either excluded adult\cite{huang2021nomogram, shi2022evaluation} or neonatal and pediatric patients\cite{tsiklidis2022predicting,hou2020predicting,dai2020analysis} to conduct analysis. However, synthetic data are often generated by methods trained on all patients combined. The consequence of this mismatch between the population used to generate synthetic data and the analytic subpopulation is less well understood and needs further research. Second, as the privacy concerns may vary significantly depending on the context of use,  it will be useful to develop a framework and tools to fix privacy exposure risk at a pre-specified level and then optimize fidelity or utility.}

\textcolor{black}{We now discuss the remaining limitations of this study.}

\textcolor{black}{First}, with the rapid expansion of this area, this study does not cover all the available methods for benchmarking. However, our publicly available evaluation package facilitates the addition of new methods to the benchmark. 

\textcolor{black}{Second}, when evaluating synthetic EHR generation on MIMIC-IV, we excluded diagnostic events not present in MIMIC-III to ensure consistent evaluation dimensions, which may limit insights into domain adaptation. Future research should develop better evaluation metrics for domain adaptation scenarios. 

\textcolor{black}{Third}, while we selected key evaluation metrics, the study does not cover all the metrics ever used in previous works. However, we believe that our selected metrics address the most critical concerns. Our benchmarking package also enables the inclusion of additional metrics as needed. 

\textcolor{black}{Fourth}, the benchmarking in this study did not create a composite weighted evaluation metric to compare the methods across different scenarios. Assigning weights to different metrics can be application-dependent. We leave the development of such a metric as a potential future direction. 

\textcolor{black}{Fifth}, as the decision tree is obtained based on our evaluation results of the seven selected methods and baselines, it is not necessarily always applicable when we face different EHR datasets or populations with different distributions from the MIMIC-III/IV cohorts. We suggest similar evaluation pipelines in these settings, e.g., by adapting the SynthEHRella package which is released for public use. 

\textcolor{black}{Sixth}, the analytical utility evaluation in this work, conducted via bivariate logistic regressions, overlaps conceptually with the evaluation of correlation difference in fidelity evaluation. This overlap is primarily because the lack of socio-demographic characteristics in the generated EHR data. However, analytical utility evaluation provides further comparisons of the confidence intervals on the association of popular diseases. Furthermore, in the future one could add confounders from methods capable of generating data with socio-demographic information to evaluate their ability in preserving adjusted associations. 

\textcolor{black}{Seventh}, this study focuses only on the transportability of models trained on MIMIC-III to MIMIC-IV, without considering the possibility of training models directly on MIMIC-IV. While extending existing methods to train on a subset of MIMIC-IV data is possible, we leave this for future work. 

Lastly, in this work we assumed that no missing data in the real EHR data ~\cite{yoon2023ehr, li2023generating}. Future benchmarking work is needed to evaluate the implication of missing data in the quality of synthetic EHR generation. 

\section{\textcolor{black}{PRISMA-ScR Checklist}}

\textcolor{black}{For ensuring clarity, transparency and reproducibility, in this section, we include a flow diagram of our literature review and benchmarking procedure in Figure~\ref{fig:review-flowchart}, as well as a PRISMA-ScR checklist\cite{tricco2018prisma} (attached at the end of the Supplmentary Materials) corresponding to our literature search. Note that because this paper is a methodological scoping review rather than a scoping review that involves synthesizing numerical evidence about specific parameters of scientific interest, some cells in the PRISMA-ScR checklist are not applicable to our context (e.g., Item 10-12,15-17).} \textcolor{black}{The scoping review presented in this paper, primarily focuses on surveying the existing computational methodologies for synthetic EHR generation and defines the scope for future work for new methods development.}

\clearpage
\includepdf[pages=-,pagecommand=\thispagestyle{plain}]{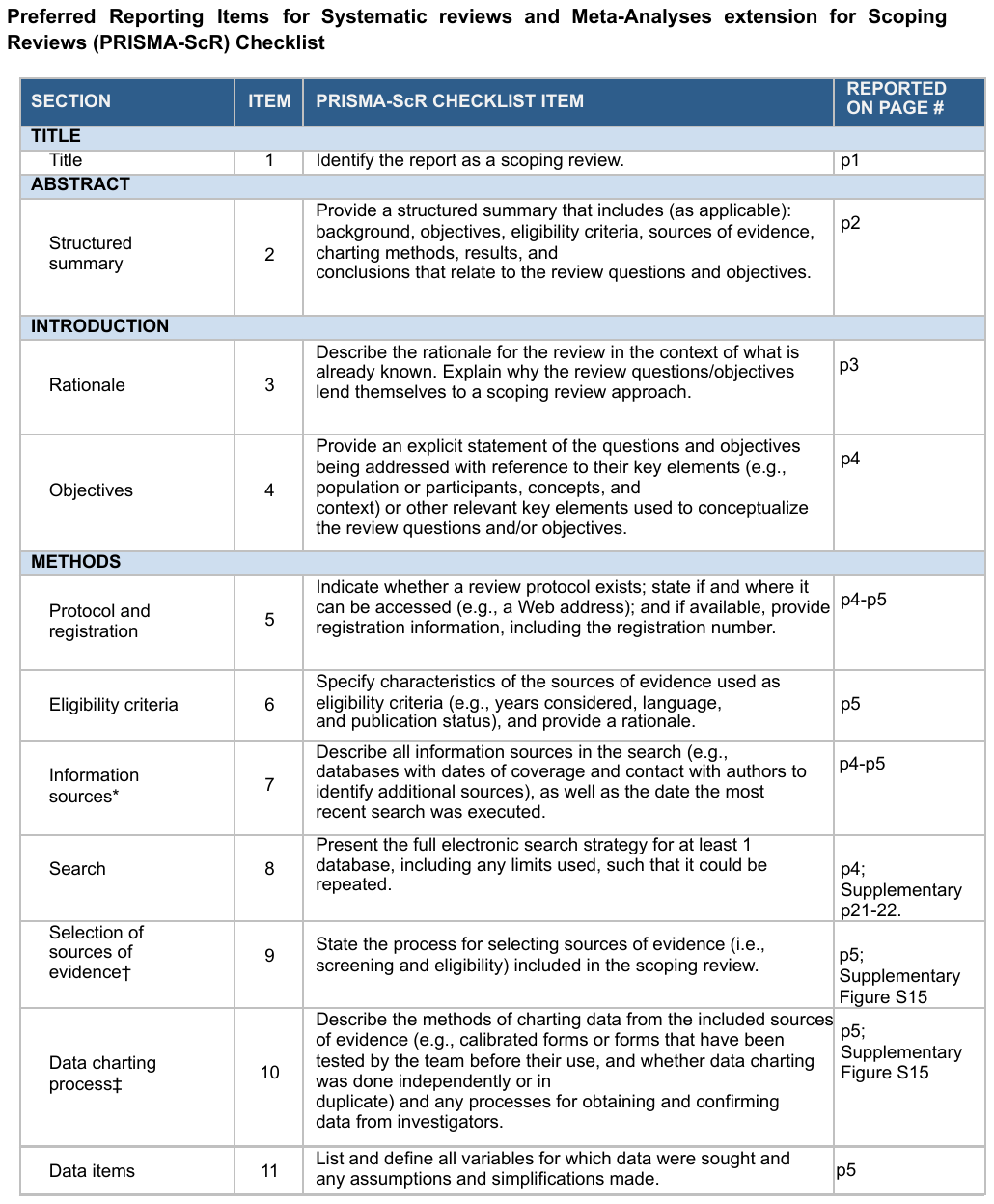}

\end{document}